%% file: main.tex
\documentclass[10pt,twocolumn,letterpaper]{article}

\usepackage[pagenumbers]{iccv} 
\usepackage[linesnumbered, boxed, ruled]{algorithm2e}
\usepackage{tikz}
\usepackage{adjustbox}
\usepackage{colortbl}
\usepackage[accsupp]{axessibility}  
\usepackage{multirow}     
\usepackage{graphicx}     
\usepackage{hhline}

\newcommand{\rulesep}{\unskip\ \vrule\ }
\definecolor{iccvblue}{rgb}{0.21,0.49,0.74}
\definecolor{lightcyan}{rgb}{0,1,1}
\usepackage[pagebackref,breaklinks,colorlinks,allcolors=iccvblue]{hyperref}

\def\paperID{2610} 
\def\confName{ICCV}
\def\confYear{2025}

\title{Streamlining Image Editing with Layered Diffusion Brushes}

\author{
Peyman Gholami \qquad Robert Xiao\\
University of British Columbia\\
{\tt\small \{peymang, brx\}@cs.ubc.ca}\\[0.5em] 
{\tt \normalsize \href{https://layered-diffusion-brushes.github.io/}{https://layered-diffusion-brushes.github.io/}}
}

\begin{document}
\twocolumn[{%
\renewcommand\twocolumn[1][]{#1}%
\maketitle
\vspace{-1em}
\centering
\begin{minipage}[t]{0.08\textwidth}
    \small{\textbf{Edit prompt\\(type)}}
  \label{fig:mon3}
  \end{minipage}
    \begingroup
    \fontsize{8pt}{10pt}\selectfont
\begin{minipage}[t]{0.13\textwidth}
\centering
  sunglasses\\
  (add objects)
\end{minipage}
\begin{minipage}[t]{0.13\textwidth}
\centering
  hat, Takashi \\
  Murakami style\\
  (change style)
\end{minipage}
\begin{minipage}[t]{0.13\textwidth}
\centering
  starry night, van gogh style
  (fix problems, maintaining style)
\end{minipage}
\begin{minipage}[t]{0.13\textwidth}
\centering
  sky, Leonid \\
  Afremov style\\
  (mixing styles)
\end{minipage}
\begin{minipage}[t]{0.13\textwidth}
\centering
  ornate frame\\
  (change attribute / texture)
\end{minipage}
\endgroup
\\
\begin{minipage}[t]{0.08\textwidth}
\vspace{-2.0cm}
  \small\textbf{Layer Mask}
\end{minipage}
\begin{minipage}[b]{0.13\textwidth}
  \includegraphics[width=\textwidth]{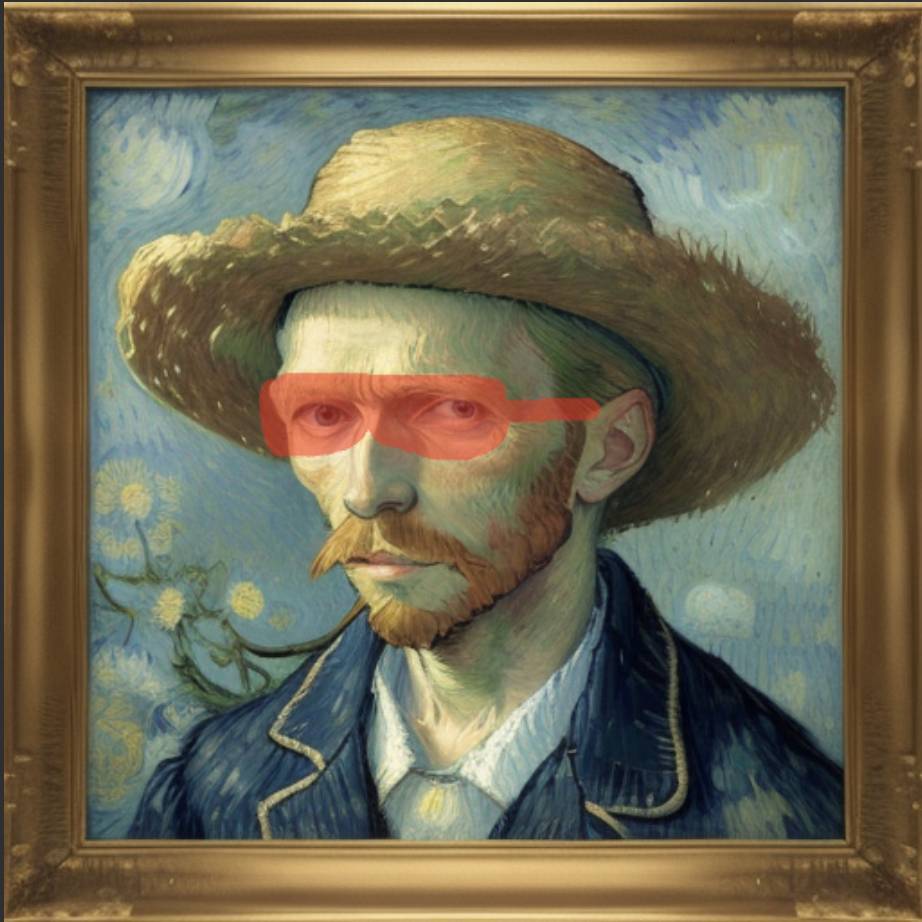}
\end{minipage}
\begin{minipage}[b]{0.13\textwidth}
  \includegraphics[width=\textwidth]{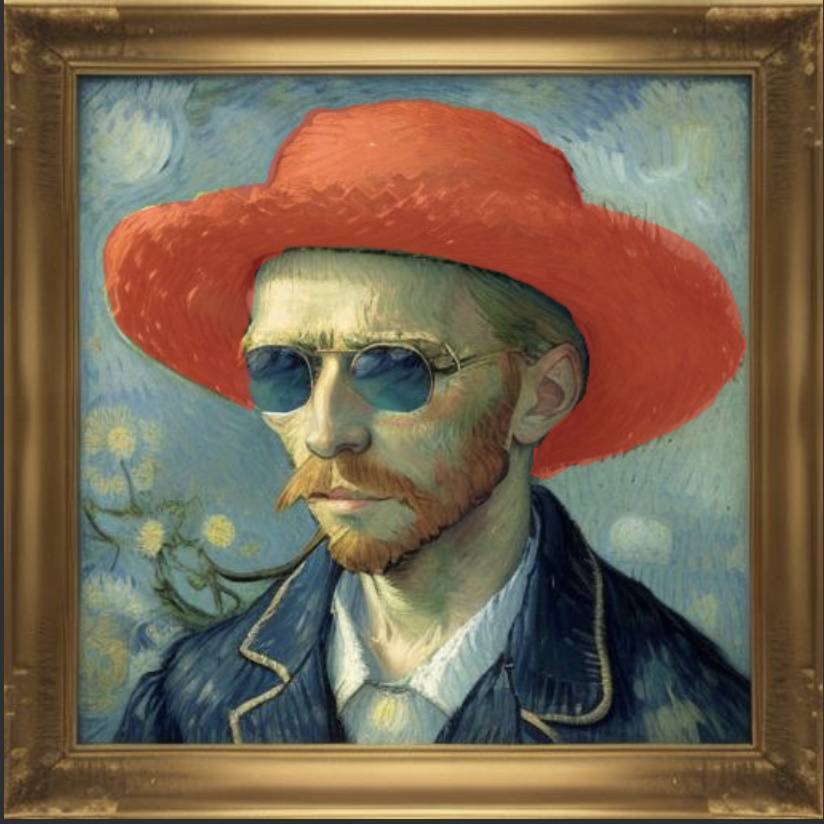}
\end{minipage}
\begin{minipage}[b]{0.13\textwidth}
  \includegraphics[width=\textwidth]{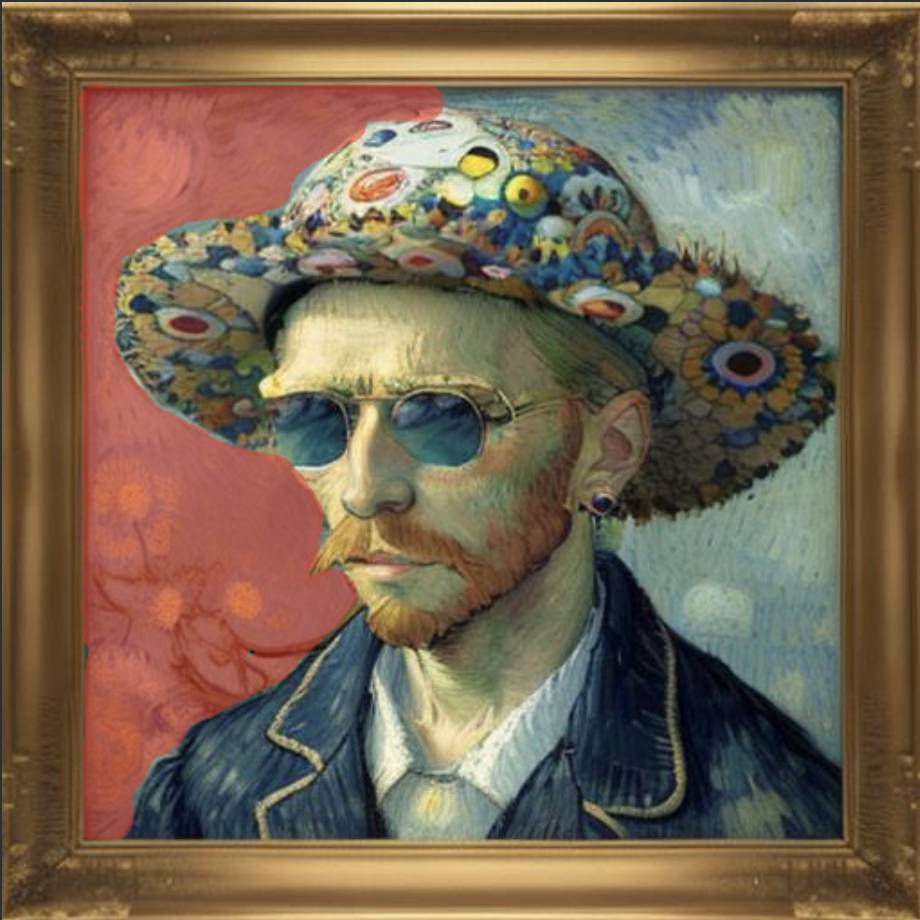}
\end{minipage}
\begin{minipage}[b]{0.13\textwidth}
  \includegraphics[width=\textwidth]{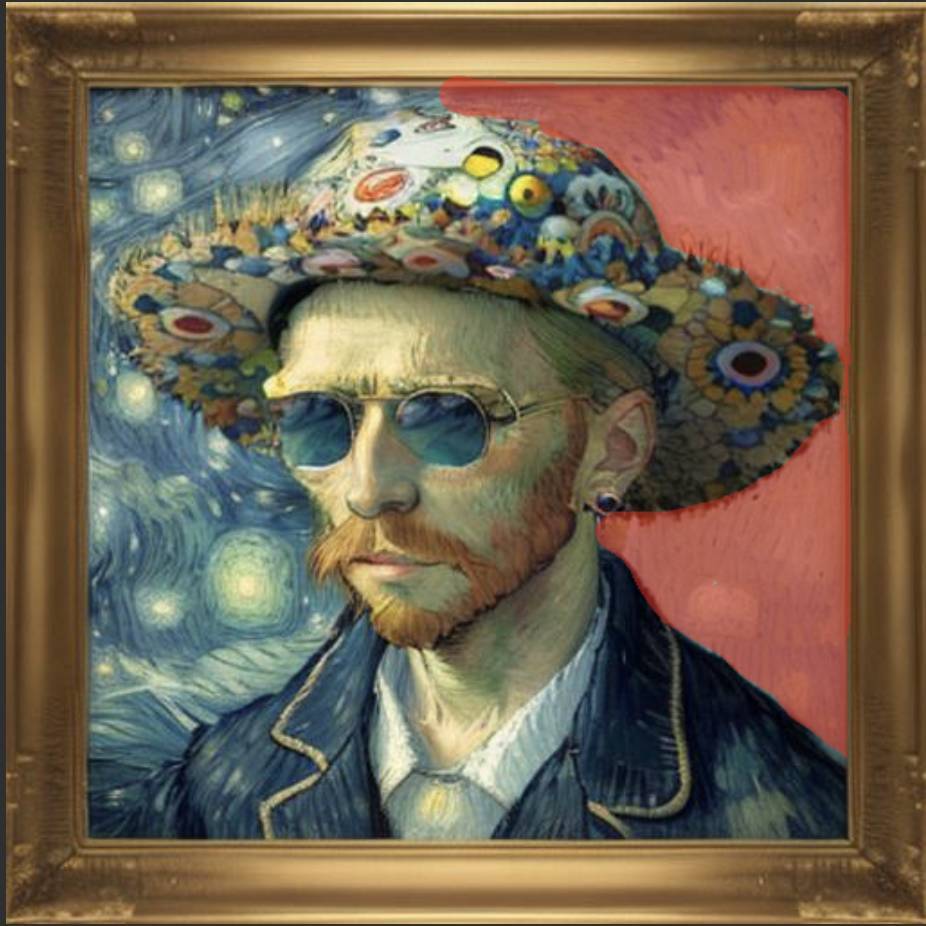}
\end{minipage}
\begin{minipage}[b]{0.13\textwidth}
  \includegraphics[width=\textwidth]{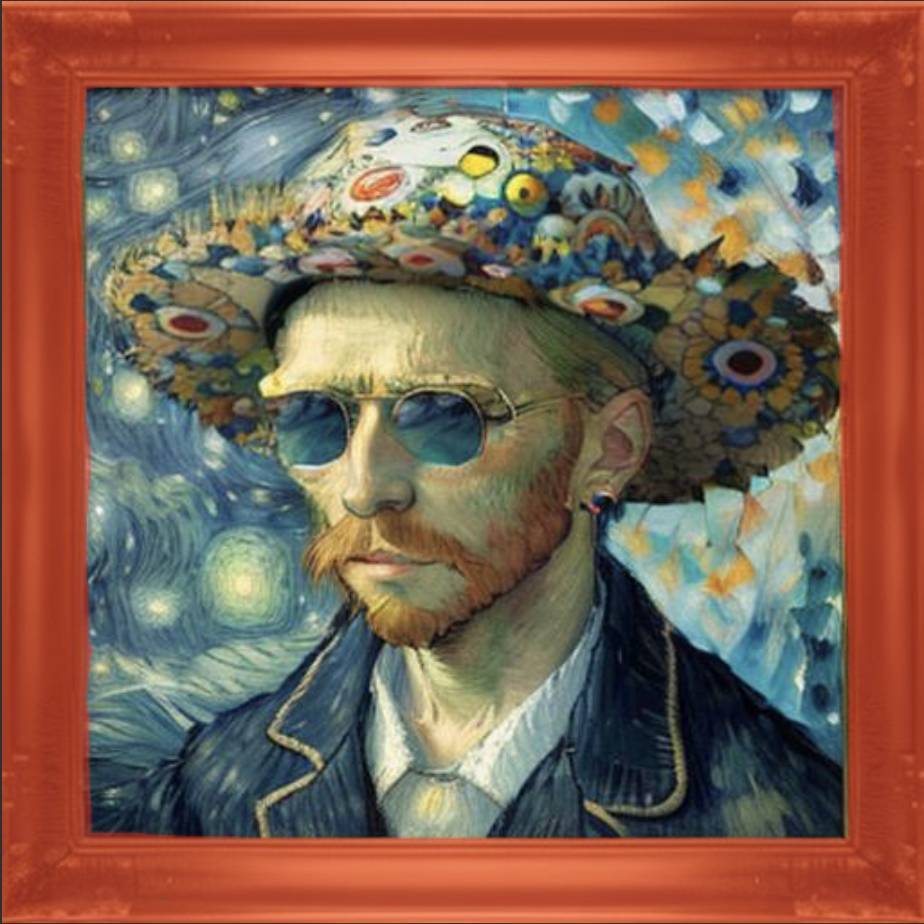}
\end{minipage}
\\
\begin{minipage}[t]{0.08\textwidth}
\vspace{-2.3cm}
    \small\textbf{Edited\\ image}
\end{minipage}
\begin{minipage}[b]{0.13\textwidth}
  \includegraphics[width=\textwidth]{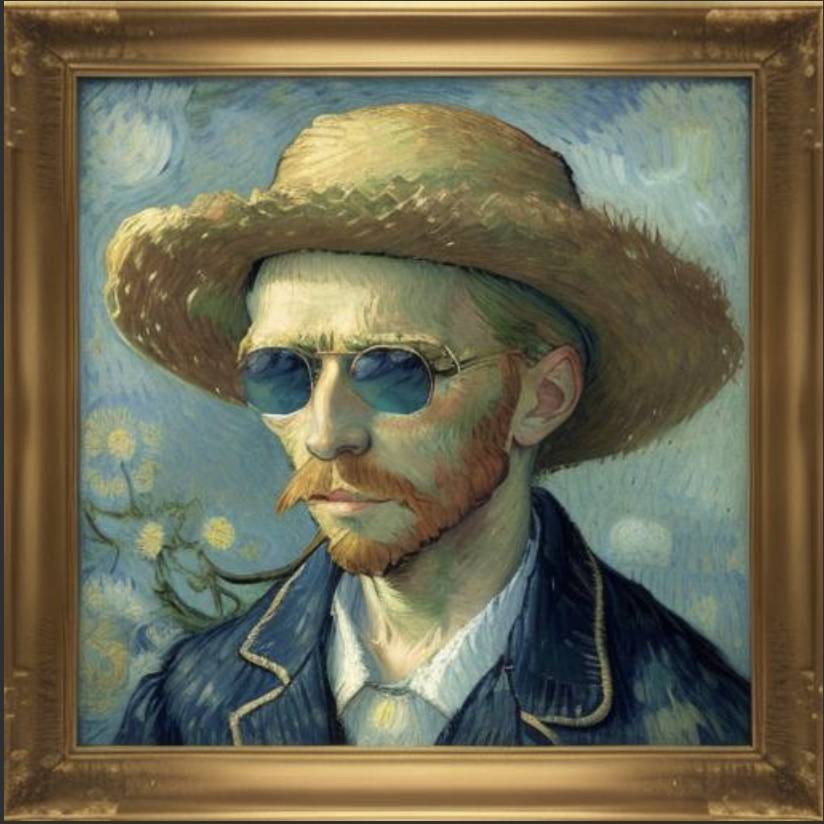}
    \captionof*{minipage}{Layer 1}
\end{minipage}
\begin{minipage}[b]{0.13\textwidth}
  \includegraphics[width=\textwidth]{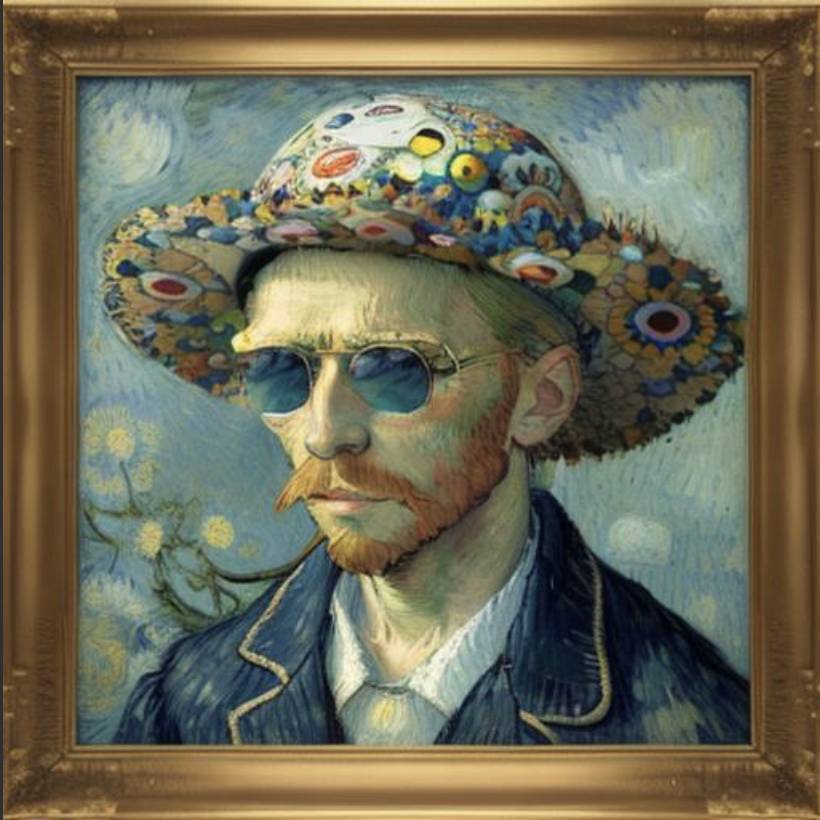}
    \captionof*{minipage}{Layer 2}
\end{minipage}
\begin{minipage}[b]{0.13\textwidth}
  \includegraphics[width=\textwidth]{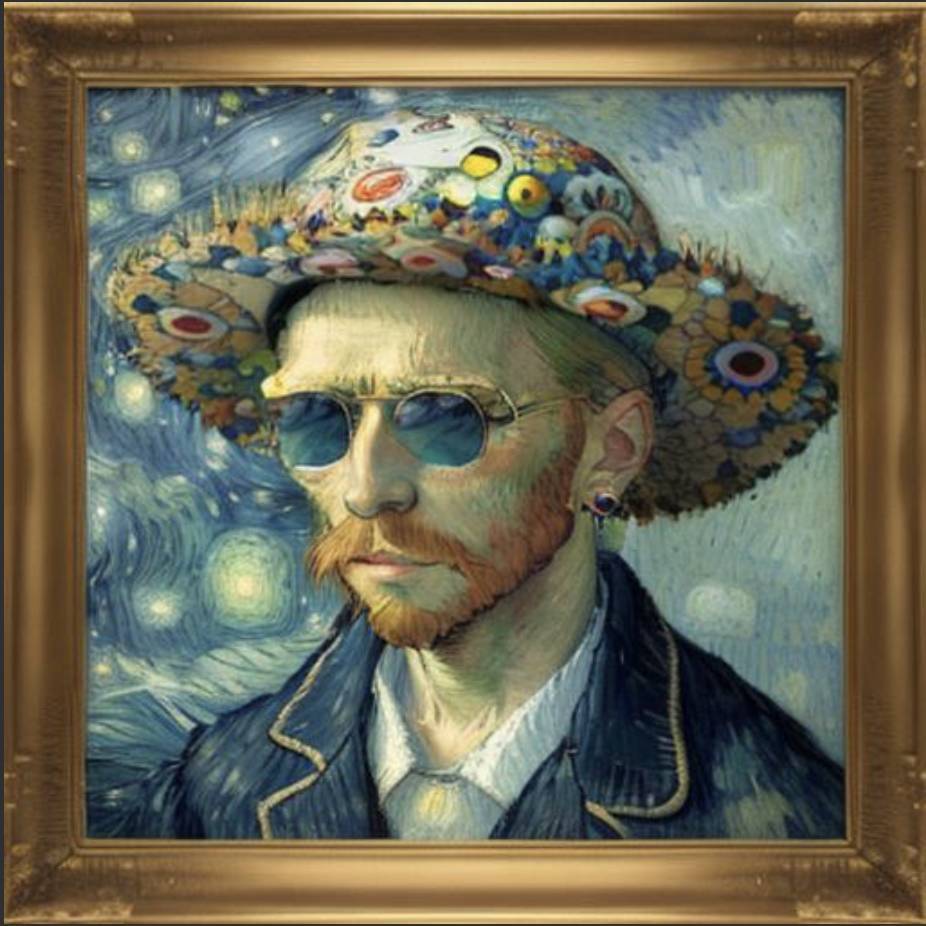}
  \captionof*{minipage}{Layer 3}
\end{minipage}
\begin{minipage}[b]{0.13\textwidth}
  \includegraphics[width=\textwidth]{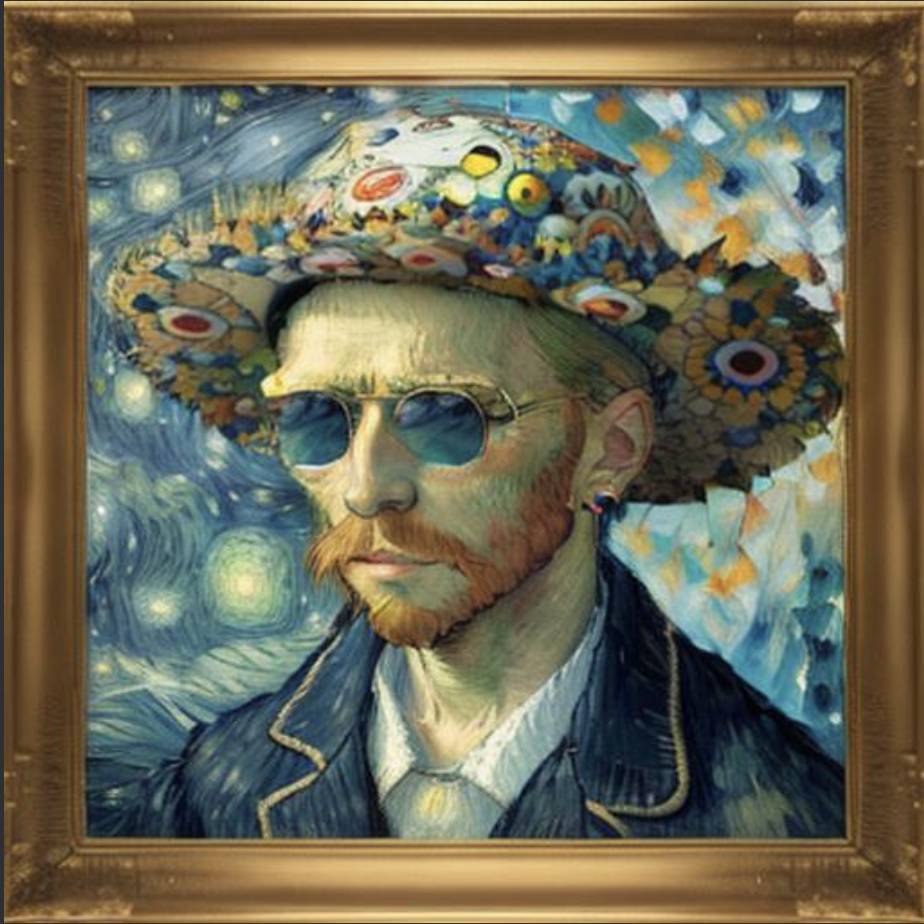}
  \captionof*{minipage}{Layer 4}
\end{minipage}
\begin{minipage}[b]{0.13\textwidth}
  \includegraphics[width=\textwidth]{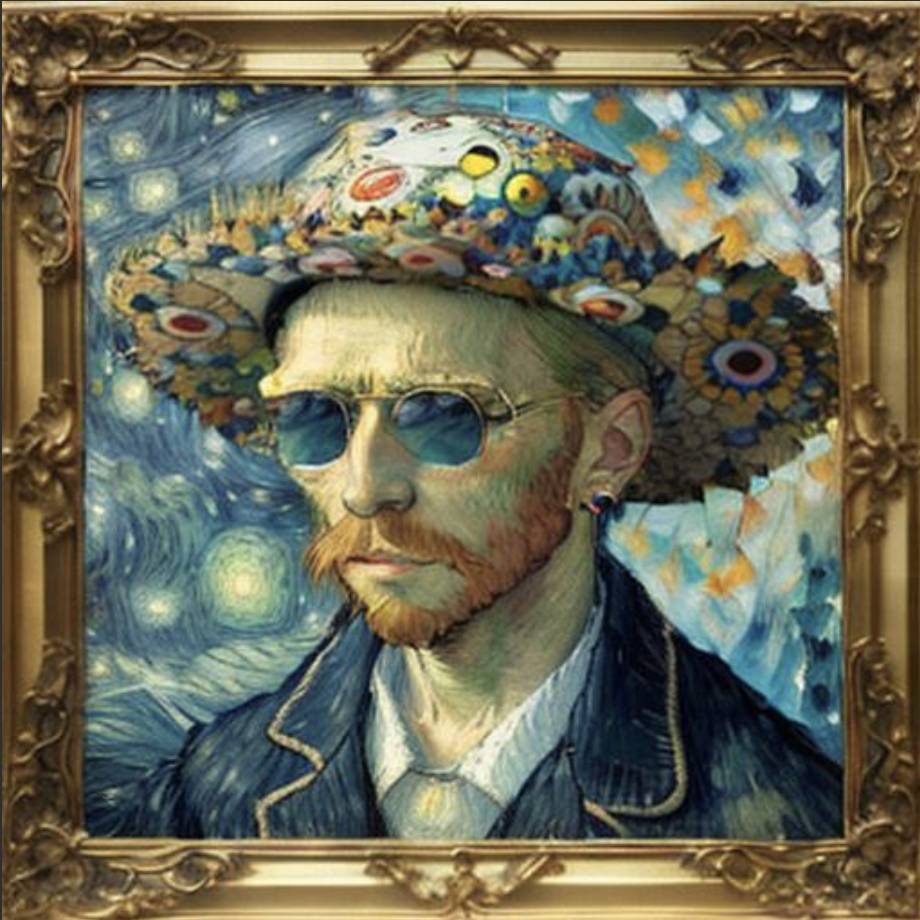}
  \captionof*{minipage}{Layer 5}
\end{minipage}
\captionof{figure}{Hierarchical image editing with Layered Diffusion Brushes: LDB is capable of creating and stacking a wide range of independent edits, including object addition, removal, or replacement, colour and style changes/combining, and object attribute modification. Each edit is performed independently, and users are able to switch between the edits seamlessly.
\vspace{2em}
}
\label{fig:teaser}
}]

\input{sec/0_abstract}    
\input{sec/1_intro}
\input{sec/2_related_works}

\input{sec/3_method}

\input{sec/4_experiments}

\input{sec/5_discussion}
\section*{Acknowledgment}
This work was supported by the Natural Sciences and Engineering Research Council of Canada (NSERC). The authors would also like to thank Professors Leonid Sigal and Kwang Moo Yi for their guidance and support throughout this project.

{
    \small
    \bibliographystyle{ieeenat_fullname}
    \bibliography{main}
}

\input{sec/X_suppl}

\end{document}

%% file: sec/0_abstract.tex
\begin{abstract}
Denoising diffusion models have emerged as powerful tools for image manipulation, yet interactive, localized editing workflows remain underdeveloped. We introduce Layered Diffusion Brushes (LDB), a novel training-free framework that enables interactive, layer-based editing using standard diffusion models. LDB defines each ``layer'' as a self‑contained set of parameters guiding the generative process, enabling independent, non-destructive, and fine-grained prompt-guided edits, even in overlapping regions. LDB leverages a unique intermediate latent caching approach to reduce each edit to only a few denoising steps, achieving 140 ms per edit on consumer GPUs. An editor implementing LDB, incorporating familiar layer concepts, was evaluated via user study and quantitative metrics. Results demonstrate LDB's superior speed alongside comparable or improved image quality, background preservation, and edit fidelity relative to state-of-the-art methods across various sequential image manipulation tasks. The findings highlight LDB's ability to significantly enhance creative workflows by providing an intuitive and efficient approach to diffusion-based image editing and its potential for expansion into related subdomains, such as video editing.
\end{abstract}

%% file: sec/1_intro.tex
\section{Introduction}
\label{sec:intro}

Image editing has undergone transformative advancements with the rise of text-to-image (T2I) generative models, enabling unprecedented creative expression through textual guidance. These models, including Generative Adversarial Networks (GANs) \cite{goodfellow2014generative}, Variational Autoencoders (VAEs), and Denoising Diffusion Models (DMs) \cite{ho2020denoising}, have redefined image synthesis and manipulation. Among these, DMs \cite{song2020score} have emerged as the state of the art due to their training stability, high-fidelity outputs, and versatility across tasks like inpainting \cite{lugmayr2022repaint}, super-resolution \cite{saharia2022image}, and style transfer \cite{hertz2022prompt}. However, despite their capabilities, a critical gap remains: enabling \textbf{real-time, localized, and iterative edits} that align with professional workflows, where artists demand precise control over specific regions without disrupting the global composition.

Existing DM-based editing methods face several core challenges. First, their stochastic nature often necessitates numerous generations to achieve desired results \cite{avrahami2023blended}. Second, they lack intuitive mechanisms for layered, non-destructive editing—a cornerstone of tools like Adobe Photoshop \cite{adobephotoshop}—where edits can be independently adjusted, stacked, or removed. Third, while mask-guided approaches  enable regional control, they struggle with seamless blending, artifact-free transitions, and real-time feedback. These limitations restrict their adoption in creative pipelines, where rapid iteration and granular control are critical.

To address these challenges, we propose \textit{Layered Diffusion Brushes (LDB)}, a novel framework based on Latent Diffusion Models (LDM)  \cite{rombach2022high} that integrates mask-guided diffusion with a non-destructive layered editing paradigm. 

At its core, LDB introduces new noise patterns into the image latents during diffusion process, guided by both the user-specified mask and the edit prompt. This preserves the original context while seamlessly integrating localized edits. We implement an intuitive user interface (UI) with a layering system to support consecutive edits (\cref{fig:teaser}). Specifically, as key contributions, LDB introduces: 
\begin{itemize}
\item \textbf{Latent Caching for Real-Time Edits:} By reusing intermediate denoising states from initial generation, edits bypass redundant computations and achieve as low as 140 ms per edit on 512×512 images  (53× faster than BrushNet \cite{ju2024brushnet} using the same consumer GPUs).

\item \textbf{Non-destructive Layered Editing:} LDB introduces an order-agnostic layering mechanism by defining the concept of layers for DMs, enabling:
\begin{itemize}[leftmargin=*]
\item Region-targeted adjustments with background preservation, using mask-prompt pairs,
\item  Stacking, toggling, or deleting layers without cross-interference—even in overlapping regions, 
\item Post-hoc revision of edits while preserving underlying content.
\end{itemize}

\item \textbf{Seed-Driven Exploration:}
Our UI provides familiar ``brush'' and ``scroll'' gestures to enable instant exploration of variations by modulating noise seeds, bridging stochastic generation with deterministic refinement and instant feedback.
\end{itemize}

We validate LDB through extensive experiments and a user study with graphic designers. Quantitatively, LDB outperforms state-of-the-art methods in terms of speed and image quality and is comparable in terms of edit fidelity. The user study revealed superior usability and creativity support in iterative design.
Additionally, LDB is a plug-and-play, training-free system adaptable to existing models and applications, and we demonstrate this by applying LDB to the task of video editing.

%% file: sec/2_related_works.tex


\section{Related Work}
\label{sec:related}

\subsection{DM-based Image Editing}
Image editing is the task of modifying existing images in terms of appearance, structure, or composition, ranging from subtle adjustments to major transformations.
Unlike GAN-based approaches \cite{abdal2021styleflow, lang2021explaining, pan2023drag}, which are prone to limitations in inversion stability \cite{richardson2021encoding} and localized control \cite{bar2022text2live}, diffusion-based methods harness the power of controllable, high-quality DMs in various image-editing tasks, including text and image-driven image manipulation studies \cite{kim2022diffusionclip,hu2022global,couairon2022diffedit,lugmayr2022repaint}.

Instruction-based text editing methods \cite{brooks2022instructpix2pix, guo2023focus, geng2023instructdiffusion, geng2024instructdiffusion, zeng2025promptfix} typically train DMs on instruction-image pairs. For example, InstructPix2Pix \cite{brooks2022instructpix2pix} is trained using synthetic pairs from Stable Diffusion \cite{rombach2022high} and Prompt-to-Prompt \cite{hertz2022prompt}. However, expressing nuanced edits solely through text instructions remains challenging, particularly for object-specific style or color changes.

Mask-based methods \cite{avrahami2022blended, yu2023inpaint, avrahami2023blended, couairon2022diffedit} sample within specified regions. While effective for localized edits, they can introduce unintended global changes, especially problematic in sequential editing, and may struggle with complex edits. For instance, Blended Latent Diffusion's  lossy VAE latent space hinders perfect reconstruction even before noise addition \cite{avrahami2023blended}. Though a background reconstruction strategy is included, it increases computation and may still yield incoherent results for complex edits. Conversely, our method directly modifies the original latent space, enhancing context preservation and natural blending.

Attention-based editing manipulates cross-attention maps to guide the image generation process toward the desired modifications \cite{hertz2022prompt, zeroshot}. These methods generally face challenges in achieving fine-grained edits without unwanted global modifications. Yang et al. \cite{yang2023dynamic} attribute unintended changes to inaccurate attention maps and propose attention focusing. Inversion-based methods like ILVR \cite{choi2021ilvr}, Textual Inversion \cite{gal2022image}, and DreamBooth \cite{ruiz2023dreambooth} focus on context modification while preserving subjects. DDIM inversion converts images to noisy latents, and sampling generates edited results based on prompts. We employ Direct Inversion \cite{ju2023direct} for efficient real image latent inversion.

Image inpainting involves replacing or restoring the missing regions while maintaining global coherency \cite{xu2023review}. 
Many inpainting works \cite{rombach2022high, manukyan2023hd, zhuang2024task, yang2023uni} require using a fine-tuned DM specifically designed for inpainting tasks, limiting their applicability. Some, including SmartBrush, which uses object-mask prediction guidance \cite{xie2023smartbrush}, offer more flexibility. PowerPaint \cite{zhuang2024task} introduces learnable task embeddings for improved control. While these models effectively generate new content, they are generally unsuitable for making small, targeted adjustments \cite{avrahami2022blended, lugmayr2022repaint, saharia2022palette}. Inspired by ControlNet \cite{zhang2023adding}, BrushNet \cite{ju2024brushnet} builds a decomposed plug-and-play dual-branch DM, but struggles with real-time interaction due to its computational overhead. In \cref{experiments} we compare LDB with several inpainting techniques.

\subsection{Layered and Sequential Image editing}
Layer-based image editing is fundamental in computer graphics \cite{porter1984compositing}, and recent works integrate this concept into AI methodologies \cite{bar2022text2live, Sarukkai_2024_WACV}. Layered representations enable dynamic manipulation of image components, transforming single images into multi-layered structures.

LayeringDiff \cite{kang2025layeringdiff} decomposes images into foreground and background. ParallelEdits \cite{huang2024paralleledits} uses attention for efficient multi-aspect text edits. MAG-Edit \cite{mao2024mag} employs a two-layer process with attention injection to a single edit from background.
Joseph et al. \cite{Joseph_2024_WACV} highlight error accumulation in sequential editing, where artifacts compound across edits.
Collage Diffusion \cite{Sarukkai_2024_WACV}, built on modified Blended Latent Diffusion \cite{avrahami2023blended}, employs alpha masks to guide cross-attention and generate harmonized images while respecting scene composition. However, it assumes pre-layered inputs and synthesizes scenes from scratch. In contrast, LDB is training-free, operates directly on existing images, and supports fully independent layers—unlike methods such as \cite{bar2022text2live} that require per-image training.

\subsection{Accelerated Generation using Caching}
Caching and reusing intermediate features has proven effective for accelerating DM inference through reducing redundant computations.  Several works have utilized caching in diffusion transformers (DiTs) for video generation. DeepCache \cite{ma2024deepcache} reuses high-level U-Net features in video generation, while AdaCache \cite{kahatapitiya2024adaptive} dynamically adjusts cached residuals based on temporal content. Cache Me If You Can \cite{wimbauer2024cache} employs block caching by reusing outputs from layer blocks of previous steps
 during inference.  For image generation, Approximate Caching \cite{agarwal2024approximate} reuses intermediate latents created during prior image generation processes for similar prompts. 
We employ a similar strategy through caching key latent representations and adapt it specifically for interactive image editing, enabling the real-time feedback that is crucial for creative workflows.

%% file: sec/3_method.tex
\section{Method}

We use an LDM-based variant of image generative models and make intermediate adjustments to the latent space, similar to \cite{lugmayr2022repaint, avrahami2023blended}. Therefore, LDB requires no additional training or fine-tuning of the underlying LDM; all modifications are applied during the reverse diffusion process.

We adopt the standard LDM formulation, where image generation begins with a sample from a Gaussian distribution,  $Z_0 \sim \mathcal{N}(0, \sigma^2_{max}I)$ and is iteratively denoised through a sequence of steps $N$, resulting in a series of latents $Z_i$ corresponding to decreasing noise levels $\sigma_i$, where $\sigma_0 = \sigma_{max} > \sigma_1 > \cdots > \sigma_N \approx 0$. 

\begin{figure*}[!ht]
    \centering
    \includegraphics[width=0.79\textwidth]{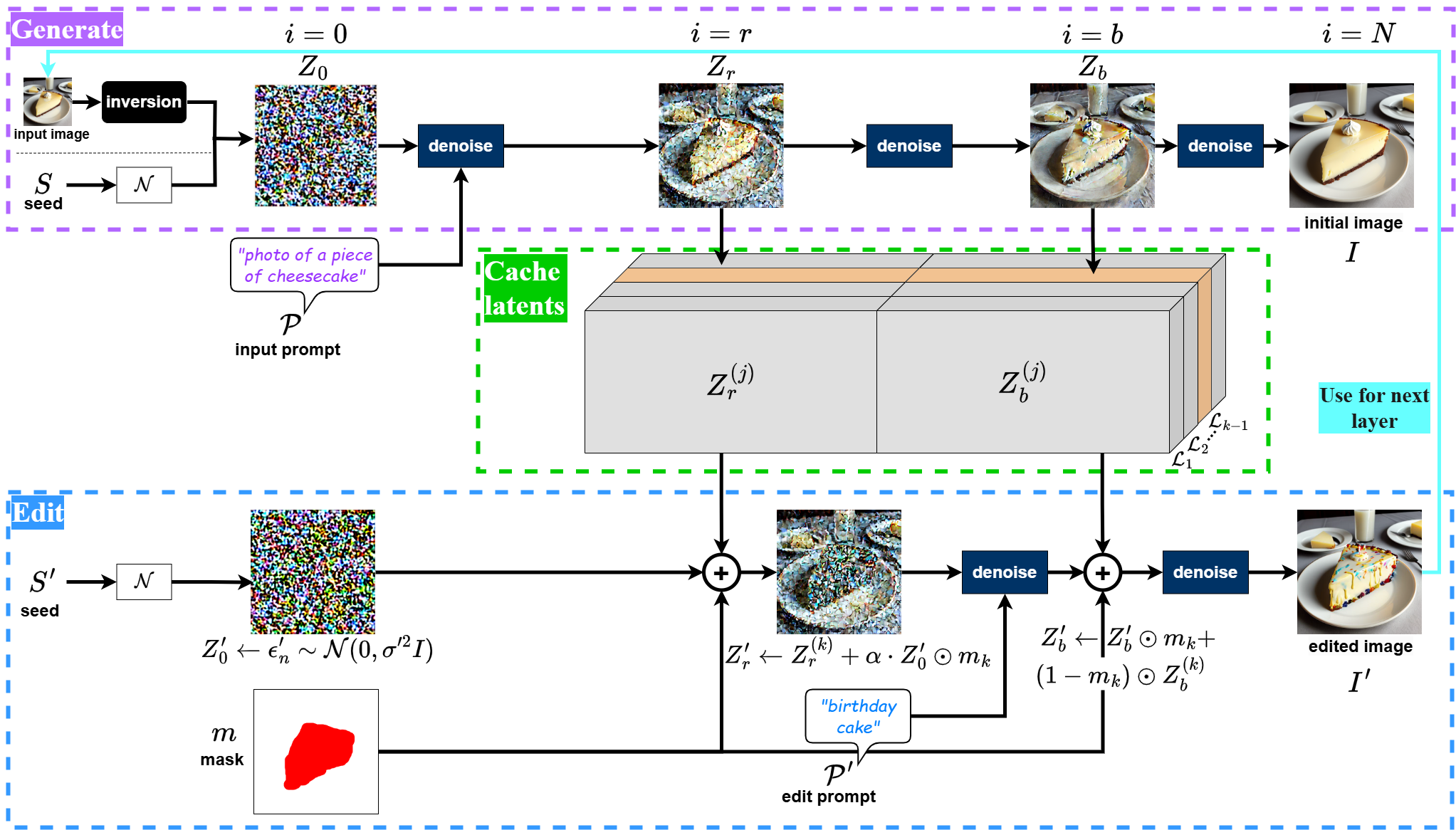}
    \caption{Overview of the Proposed Method: The top box shows standard DM-based image generation from noisy latent $Z_0$ and prompt $\mathcal{P}$. The middle section depicts the latent caching module, storing and retrieving intermediate latents for different layers. The bottom box illustrates the editing process: a new noise sample $S'$ merges with the original latent at step $r$ using mask $m$ and strength control $\alpha$. Diffusion continues until step $b$, where modified and cached latents blend to generate the final edited image.}
    \label{fig:overview}
\end{figure*}

As demonstrated in \cref{fig:overview}, the overall LDB pipeline comprises three key stages: \textbf{\textcolor{Orchid}{initial image generation (or inversion)}, \textcolor{ForestGreen}{latent caching}, }and\textbf{ \textcolor{Cerulean}{iterative layered editing.} }

For DM-generated images, we first initialize the sample $Z_0 = \epsilon_0$ and noise level $\sigma_0$ ($i = 0$). For real images, the initial noise latent is obtained using inversion. We use Direct Inversion \cite{ju2023direct} for its high speed and comparable performance to other inversion methods, including Null-Text Inversion \cite{mokady2023null} and Negative-Prompt Inversion \cite{miyake2023negative}. The noisy sample then undergoes the diffusion process, caching certain intermediate latents to facilitate editing.

\subsection{Latent Caching}
\label{caching}
To enable rapid, interactive editing with instant exploration and feedback, we employ latent caching to reuse intermediate representations in subsequent steps, minimizing redundant computations.
We store two key intermediate latents:
\begin{itemize}
    \item \textbf{Regeneration Latent $\mathbf{Z_r}$:}  At diffusion step $r=N-n$, where $N$ is the total number of diffusion steps for initial image generation and $n$ is the number of editing steps, we cache the latent $Z_r$, which serves as the starting point for all subsequent edits. By reusing $Z_r$, we avoid recomputing the initial denoising steps for each new edit, significantly speeding up the editing process (from $N$ denoising steps to $n$). Effectively, $Z_r$ represents a partially denoised latent state that retains the global image structure but is still malleable enough to accommodate localized edits.  

    \item \textbf{Blending Latent $\mathbf{Z_b}$:} We cache the latent at diffusion step $b$ which is specifically used for the layer merging process (\cref{alg:regen}, line 7). We set $b=N-2$ for maximum background preservation (as discussed in \cref{abl}). $Z_{b}$ represents a more denoised latent compared to $Z_r$, capturing more refined image details while still allowing for seamless blending of new edits into the existing image context.  Utilizing this cached blending latent ensures smoother integration of edits and reduces visual artifacts at layer boundaries during the merging process.
\end{itemize}

\subsection{Layered Diffusion Brushes Editing}

To initiate an edit, the algorithm begins by generating a new noise pattern $Z^{\prime}_0 = \epsilon^{\prime}_k$, sampled from $\mathcal{N}(0, \sigma^{2}I)$ using a different seed $S'$, and scaling it to match the variance of the cached latent $Z_{r}$. This ensures that the additive noise stays in a reasonable range from the latent for editing, preventing visual artifacts. $Z^{\prime}_0$ is then added to the regeneration latent $Z_r$, controlled by the mask $m$ and strength $\alpha$. 

In the editing stage, at step $b$, a new noisy sample is merged with the cached blending latent using the strength control and the mask, resulting in $Z^\prime_{b}$. Subsequently, the new latent is progressively denoised from steps $b$ through $N$ and processed through the VAE to output edited image $I^\prime$. \cref{alg:regen} presents the pseudocode for the editing process for a single layer (for simplicity):




\begin{algorithm}
\caption{Single-Layer LDB Editing}
\DontPrintSemicolon
\SetAlgoLined
\SetKwInOut{Input}{Input}
\SetKwInOut{Output}{Output}
\setcounter{algocf}{0}
\Input{
    Edit prompt $\mathcal{P}'$, 
    Mask $m \in [0,1]^{H\times W}$, 
    Random seed $S'$, 
    Strength $\alpha$, 
    Number of edit steps $n$, 
    Regeneration latent $Z_r$, 
    Blending latent $Z_{b}$
}
\Output{Edited latent $Z^{\prime}_N$}
     $Z^{\prime}_0  \gets \epsilon^{\prime}_{n_k} \sim \mathcal{N}(0, \sigma^{2}I)$ \tcp*[r]{\scriptsize{sampled using seed $S'$}} 
     $Z^{\prime}_0  \gets \sqrt{Var( Z_{r})} \cdot Z^{\prime}_0 $ \tcp*[r]{\scriptsize{scale new sample}}
     $Z^{\prime}_0 \gets Z_r + \alpha \cdot (Z'_0 \odot m)$\tcp*[r]{\scriptsize{noise injection}}
 \For{$i=0,1,\ldots,n$}{
    $Z^{\prime}_{i+1}  \gets DM(Z^{\prime}_i , \mathcal{P'}, i, S^{\prime})$ \;
  \uIf{$i==b$}{
   $Z^{\prime}_{b} \gets Z^{\prime}_{b} \odot m + Z_{b} \odot (1 - m) $ \tcp*[r]{\scriptsize{blending}}
  }
 }
 \textbf{Return} $Z^{\prime}_N$
\caption{LDB editing process (single layer)}
\label{alg:regen}
\end{algorithm}

\subsection{Layer Formulation}
Unlike prior works that rely on transparent decomposable layers \cite{zhang2024transparent} or explicit object segmentation \cite{Sarukkai_2024_WACV}, we redefine a layer as a self-contained set of reproducible parameters that govern localized edits. For layer \(\mathcal{L}_k\), we formalize this as a generalized version of parameters in \cref{alg:regen}:
\begin{equation}
\resizebox{0.43\textwidth}{!}{$\mathcal{L}^{(k)} =\left(\mathbf{S^\prime}^{(k)}, \mathbf{m}^{(k)}, \mathbf{v}^{(k)}, \mathbf{Z}_r^{(k)}, \mathbf{Z}_b^{(k)}, \alpha^{(k)}, n^{(k)}, \mathcal{P^\prime}^{(k)}, j\right)$}
\end{equation}

\begin{itemize}
    \item \(\mathbf{S^\prime}^{(k)} \in \mathbb{Z}^+\): Seed space for stochastic variations
    \item \(\mathbf{m}^{(k)} \in [0,1]^{H\times W}\): Edit mask
    \item \(\mathbf{v}^{(k)} \in \{0,1\}\): Visibility state
    \item \(\mathbf{Z}_r^{(k)}, \mathbf{Z}_b^{(k)} \in \mathbb{R}^{C\times H\times W}\): Regeneration/blending latents
    \item \(\mathbf{\alpha}^{(k)} \in [0, 1]  \): Layer strength value \
    \item \(\mathbf{n}^{(k)} \in [0, N]\) Number of denoising steps
     \item \(\mathcal{P^\prime}^{(k)}\): Edit prompt
     \item \(j \in \mathbb{Z}^+\): Index of last layer index.
\end{itemize}

Notably, within a given layer \(\mathcal{L}_k\) with previous layer \(\mathcal{L}_j\), the cached latents \(\mathbf{Z}_r^{(j)}\) and  \(\mathbf{Z}_b^{(j)}\) inherently incorporate the cumulative edits from all preceding layers. This is because edits to layer \(\mathcal{L}_k\), are applied to the already edited output of layer \(\mathcal{L}_{j}\) which serves as the input to the diffusion process and the algorithm always keeps the last layer updated. Therefore, any modification in a previous layer automatically propagates through the subsequent layers.
By defining $\Phi$ as a single-layer latent generation and caching step as:
\begin{equation}
    ({Z}_r^{(k)}, {Z}_b^{(k)}) = \Phi(\mathcal{L}^{(k)}, \mathcal{L}^{(j)})
\end{equation}

\noindent in essence, if a given layer \(\mathcal{L}^{(i)}\) (where \(i < k\)) is removed or its visibility \(\mathbf{v}^{(i)}\)  is toggled, the operator \(\Phi\) will be recursively invoked to recreate all latents for layers from \(\mathcal{L}^{(i)}\) to \(\mathcal{L}^{(k)}\).
This recomputation, accelerated by latent caching,  is automatically triggered and typically completes within milliseconds to a few seconds, depending on the number of layers.
This design allows edits to remain independent yet seamlessly integrated into the final composition.



\subsubsection{Overlapping Regions}
A key advantage of layered editing in LDB is the ability to create overlapping edits, where one layer can  partially or fully modify  areas affected by earlier layers. This requires careful handling of each layer's regeneration latent, $Z_r$, to ensure that changes in visibility or content from higher layers are accurately reflected in subsequent layers, even in overlapping regions.

\begin{figure*}[!ht]
\centering
\begin{minipage}{0.62\textwidth}
    \includegraphics[width=\textwidth]{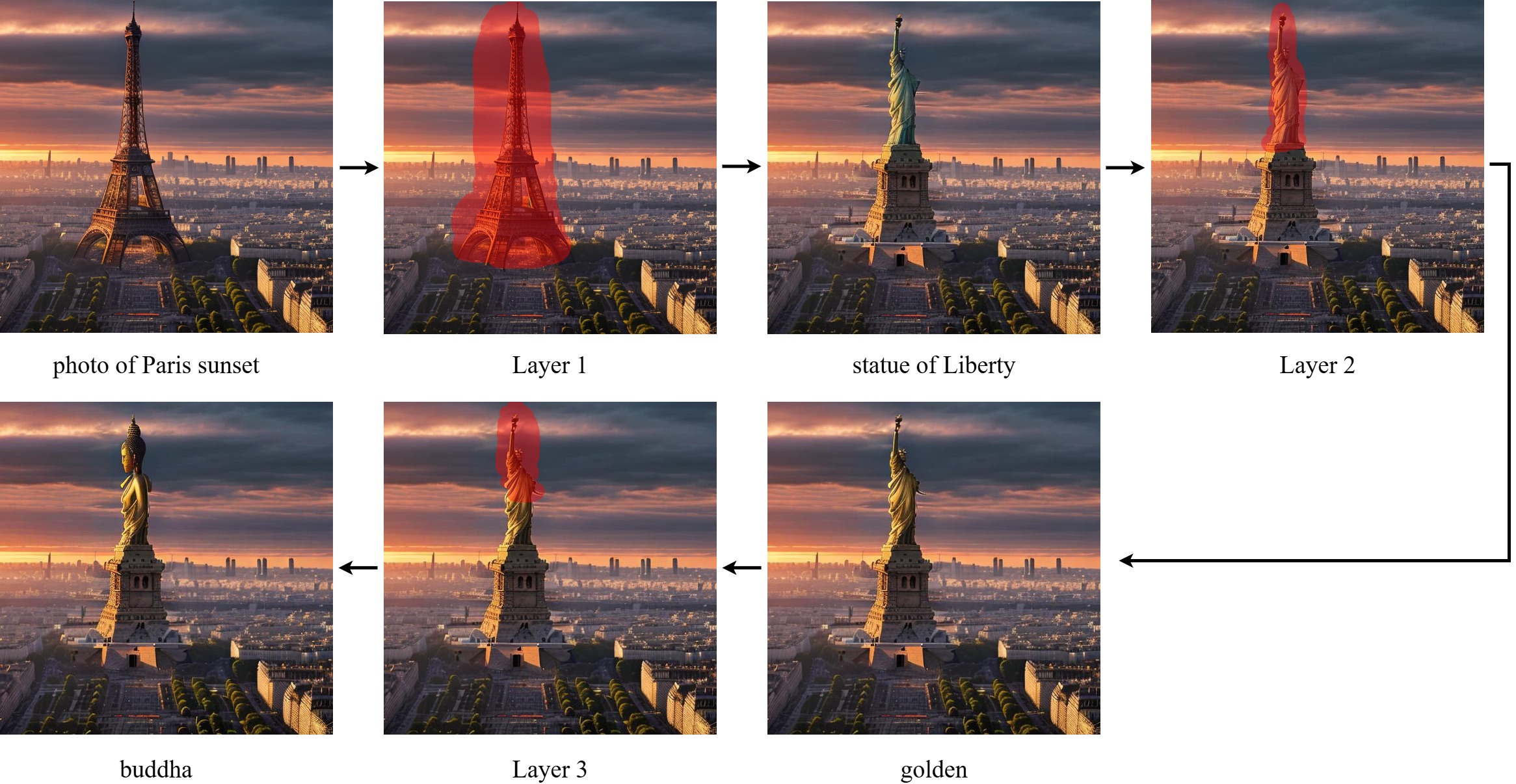}
\caption{Overlapping edit regions in LDB: overlapping edits enable complex, interacting modifications. For example, one layer can adjust color while another changes shape, with the final result combining both.}
    \label{fig:overlap}
\end{minipage}%
\hfill
\begin{minipage}{0.3\textwidth}
    
    \begin{subfigure}[t]{0.49\textwidth}
        \includegraphics[width=\textwidth, trim=0 0 0 5, clip]{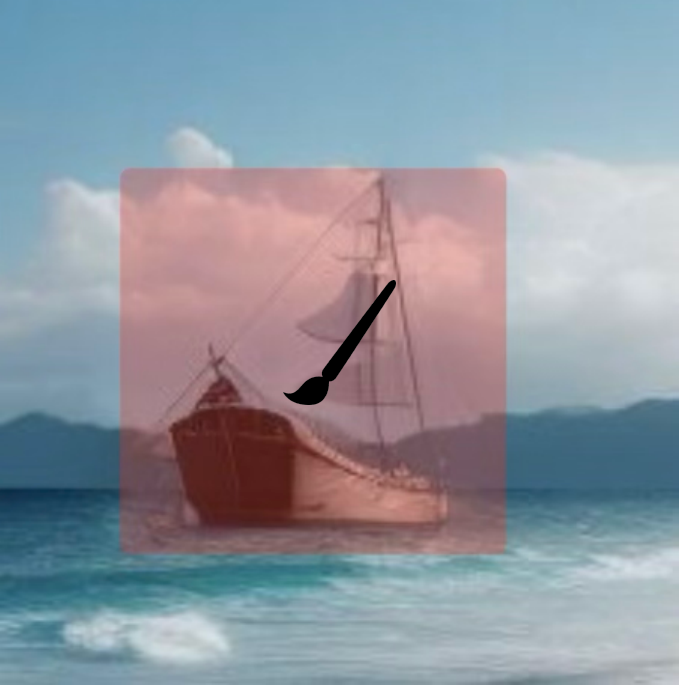}         \caption{Box option with moving cursor}
    \end{subfigure}%
    \hfill
    \begin{subfigure}[t]{0.49\textwidth}
        \includegraphics[width=\textwidth, trim=0 0 0 0, clip]{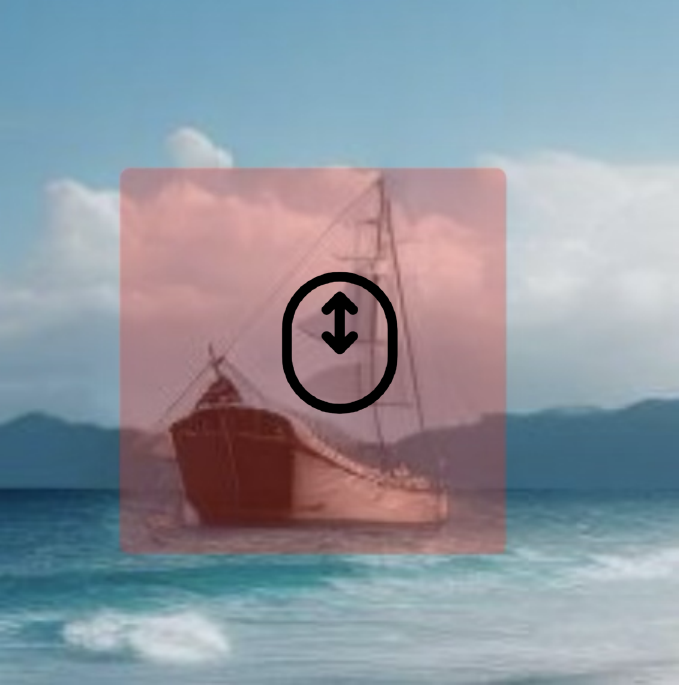}
        \caption{Custom mask option with mouse scroll}
    \end{subfigure}

    \caption{Box and Custom Mask Options: In box mode, users click the target region's center to generate edits within the specified area and can drag the box to explore variations instantly. In custom mask mode, users draw a mask over the desired region and adjust the seed using the mouse wheel or scrolling gestures to generate new variations.}
    \label{fig:brush}
\end{minipage}
\end{figure*}

By default, all layers use the initial image's latent ($Z_{r}$) as their regeneration latent. However, this approach fails to account for overlapping edits from preceding layers. To address this, when processing a layer $k$, we compute its regeneration latent by inverting the output image of the previous layer ($I^{\prime(k)}$) as shown using the \textcolor{lightcyan}{feedback arrow} on \cref{fig:overview}. This inversion yields $Z_0^{(k)}$, which is then sent through the generation stage in LDB. Both $Z_{r}^{(k)}$ and $Z_{b}^{(k)}$ are cached for efficient processing ( as shown in \cref{fig:overlap}).

This mechanism enables precise control and seamless integration of edits across overlapping regions. Changes to any layer propagate correctly without introducing artifacts, offering flexibility and fine-grained control.


\subsection{User-Interface and Interaction Design}

To develop a practical tool for artists and designers, we designed an custom UI that balances control and simplicity.
The UI allows users to generate, upload, and edit images, manage layers, and adjust parameters seamlessly.
Two interaction modes streamline edits (\cref{fig:brush}):

\textbf{Box Mode:}
Users can click or drag on the image to move a resizable square mask around. This option enables a quick and interactive exploration of how various parts of the image will change in response to a given set of editing settings (prompt and strength), simply by moving the cursor.

\textbf{Custom Mask Mode:}
Users can draw free-form masks over the desired around and navigate between new generation samples by scrolling the mouse up or down while hovering over the image, allowing them to rapidly explore variations on their edit.

We propose a workflow where users first position edits spatially using Box Mode, then refine mask geometry and appearance details via Custom Mask Mode. 

Layering capabilities include stacking, visibility toggling, and deletion. Each layer is independently modifiable. Detailed information on the UI design user interactions and a demo video can be found in supplementary material.

%% file: sec/4_experiments.tex
\section{Experiments}
\subsection{User Study}
\label{experiments}
We conducted a user study in order to evaluate the effectiveness of LDB for providing targeted image fine-tuning, using two other well-known existing image editing tools, InstructPix2Pix (IP2P) \cite{brooks2022instructpix2pix} and Stable Diffusion Inpainting (SDI) \cite{rombach2022high} as baselines for comparison.

We recruited a cohort of seven expert participants with extensive experience in using image editing software. As part of our selection criteria, we ensured that all had at least a basic level of familiarity with AI image generation techniques \cite{midjourney, ideogram} and were regular users of editing software, such as Adobe Photoshop \cite{adobephotoshop} for creating visual art.

\subsubsection{Study Procedure and Task Description}
\label{study_proc}
Each user engaged in two types of tasks: free-form tasks where users generated an image for editing using a fixed prompt and seed (type 1), and pre-determined tasks where the user worked with existing real images from the MagicBrush dataset \cite{zhang2024magicbrush} (type 2).

For the type 1 tasks, we selected specific types of edits that showcase various functionalities and capabilities of the system, including:
\begin{enumerate}
    \item Stack layers and create sequential edits (draw with LDB)

    \item Modify attributes and features of objects
    \item Correct image imperfections and errors
    \item Enhance discernibility of similar objects
    \item Target specific regions for style transfer, refine aesthetics
\end{enumerate}

\noindent Type 2 tasks were more structured, with the mask, edit prompt, and input images provided by the dataset. The dataset provides manually annotated masks and instructions for each edit. We selected a subset of 35 input images, each containing up to three layers of edits. Users refined masks/parameters if necessary and completed editing tasks.

\begin{figure*}[!ht]
  \centering

        \begin{minipage}[c]{0.99\textwidth}
      \centering
    \footnotesize \textcolor{red}{Layer 1}: \textit{``wax statue''}
  \end{minipage}
        \begin{minipage}[c]{0.10\textwidth}
        \small{PIE-Bench}
  \end{minipage} 
      \begin{minipage}[c]{0.10\textwidth}
    \includegraphics[width=\textwidth]{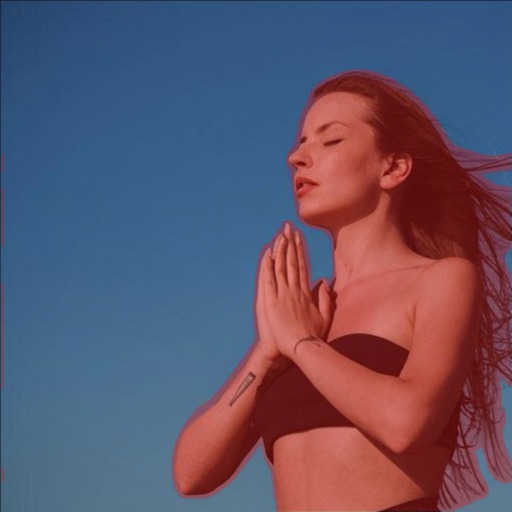}

  \end{minipage} \rulesep
          \begin{minipage}[c]{0.10\textwidth}
    \includegraphics[width=\textwidth]{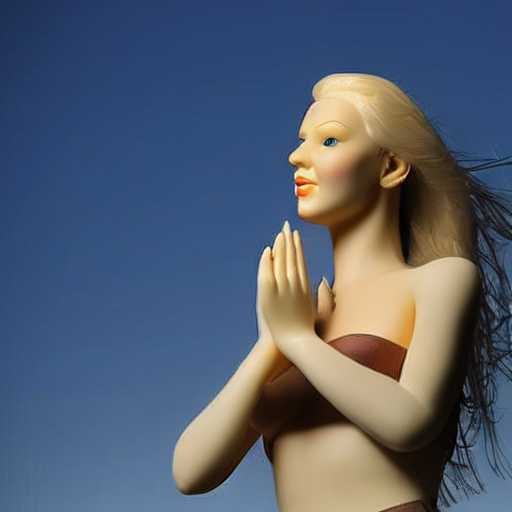}
  \end{minipage} 
    \begin{minipage}[c]{0.10\textwidth}
    \includegraphics[width=\textwidth]{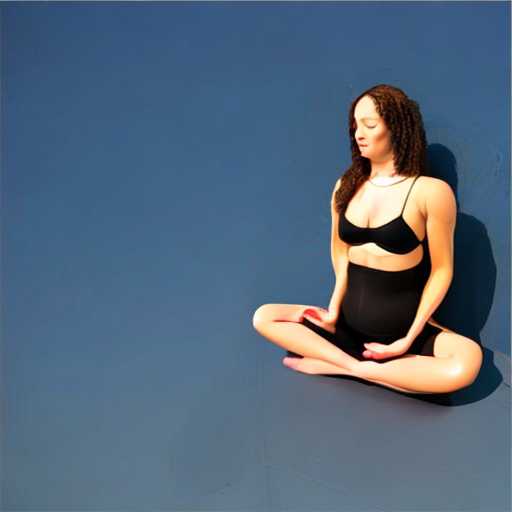}

  \end{minipage}
      \begin{minipage}[c]{0.10\textwidth}
    \includegraphics[width=\textwidth]{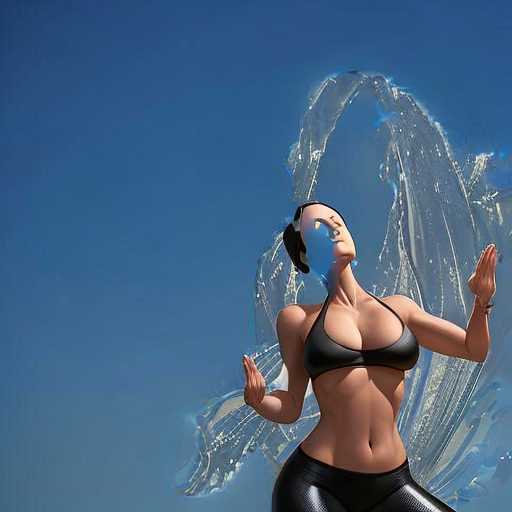}

  \end{minipage}
        \begin{minipage}[c]{0.10\textwidth}
    \includegraphics[width=\textwidth]{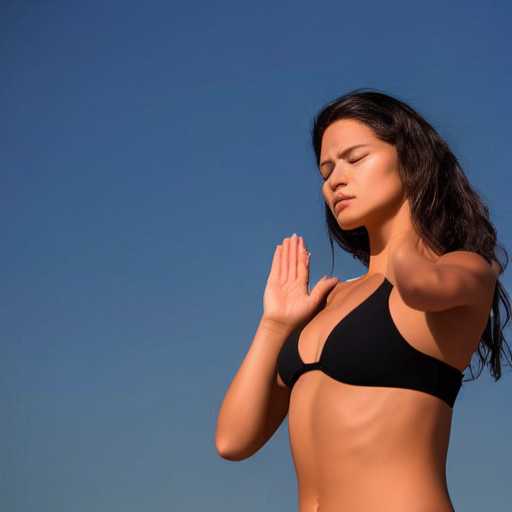}

  \end{minipage}
          \begin{minipage}[c]{0.10\textwidth}
    \includegraphics[width=\textwidth]{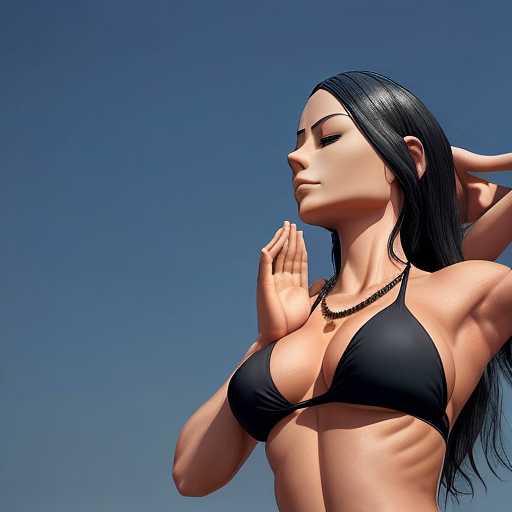}

  \end{minipage}
       \begin{minipage}[c]{0.10\textwidth}
      
    \includegraphics[width=\textwidth]{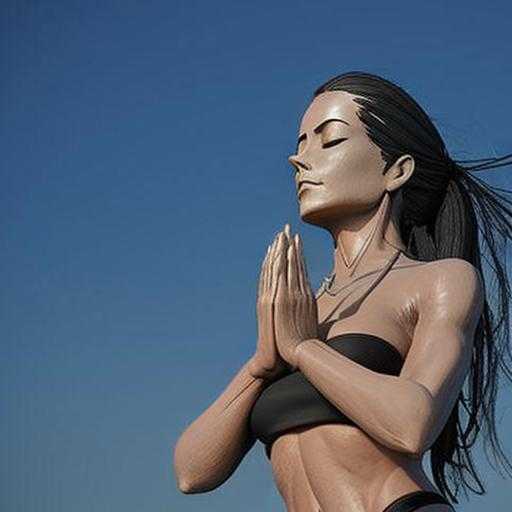}

  \end{minipage}
      \begin{minipage}[c]{0.10\textwidth}
    \centering N/A
  \end{minipage}
          \begin{minipage}[b]{0.99\textwidth}
      \centering
    \footnotesize \textcolor{red}{\textbf{}Layer 1}: \textit{boat} \qquad\textcolor{SeaGreen}{\textbf{Layer 2}}: \textit{``turtle''}
  \end{minipage}
          \begin{minipage}[c]{0.10\textwidth}
        \small{Magicbrush}

  \end{minipage} 
      \begin{minipage}[c]{0.10\textwidth}
    \includegraphics[width=\textwidth]{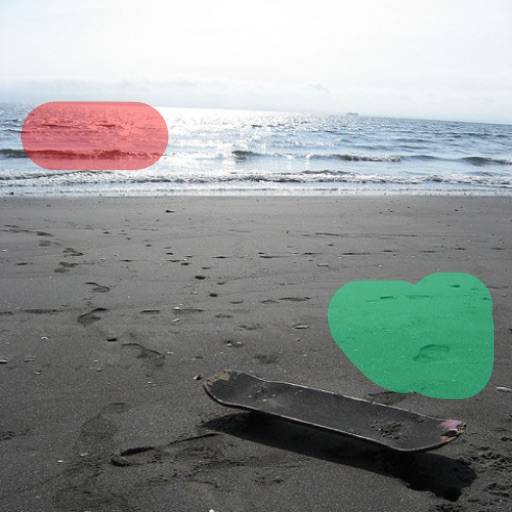}
     \caption*{\small Input image}
  \end{minipage} \rulesep
          \begin{minipage}[c]{0.10\textwidth}
    \includegraphics[width=\textwidth]{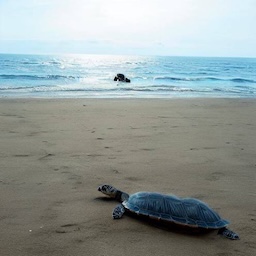}
    \caption*{IP2P}
  \end{minipage}
    \begin{minipage}[c]{0.10\textwidth}
    \includegraphics[width=\textwidth]{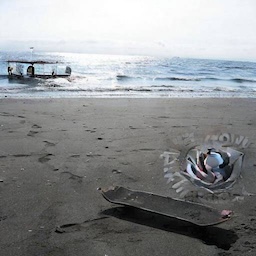}
        \caption*{SDI}
  \end{minipage}
    \begin{minipage}[c]{0.10\textwidth}
    \includegraphics[width=\textwidth]{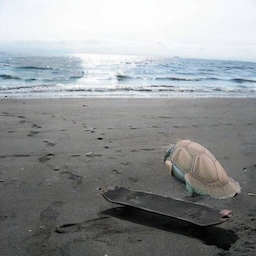}
        \caption*{BLD}
  \end{minipage}
      \begin{minipage}[c]{0.10\textwidth}
    \includegraphics[width=\textwidth]{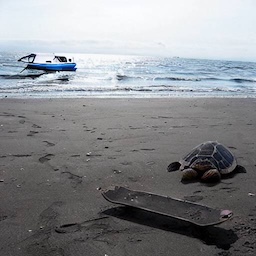}
        \caption*{HDP}
  \end{minipage}
      \begin{minipage}[c]{0.10\textwidth}
    \includegraphics[width=\textwidth]{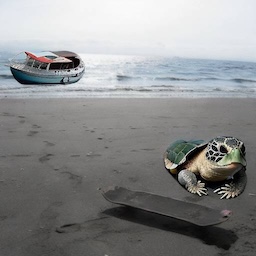}
        \caption*{BN}
  \end{minipage}
         \begin{minipage}[c]{0.10\textwidth}
    \includegraphics[width=\textwidth]{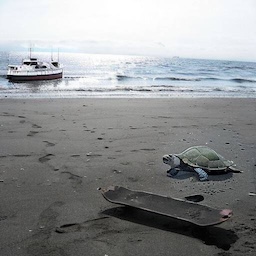}
        \caption*{  \textbf{LDB (ours)}}
  \end{minipage}
          \begin{minipage}[c]{0.10\textwidth}
    \includegraphics[width=\textwidth]{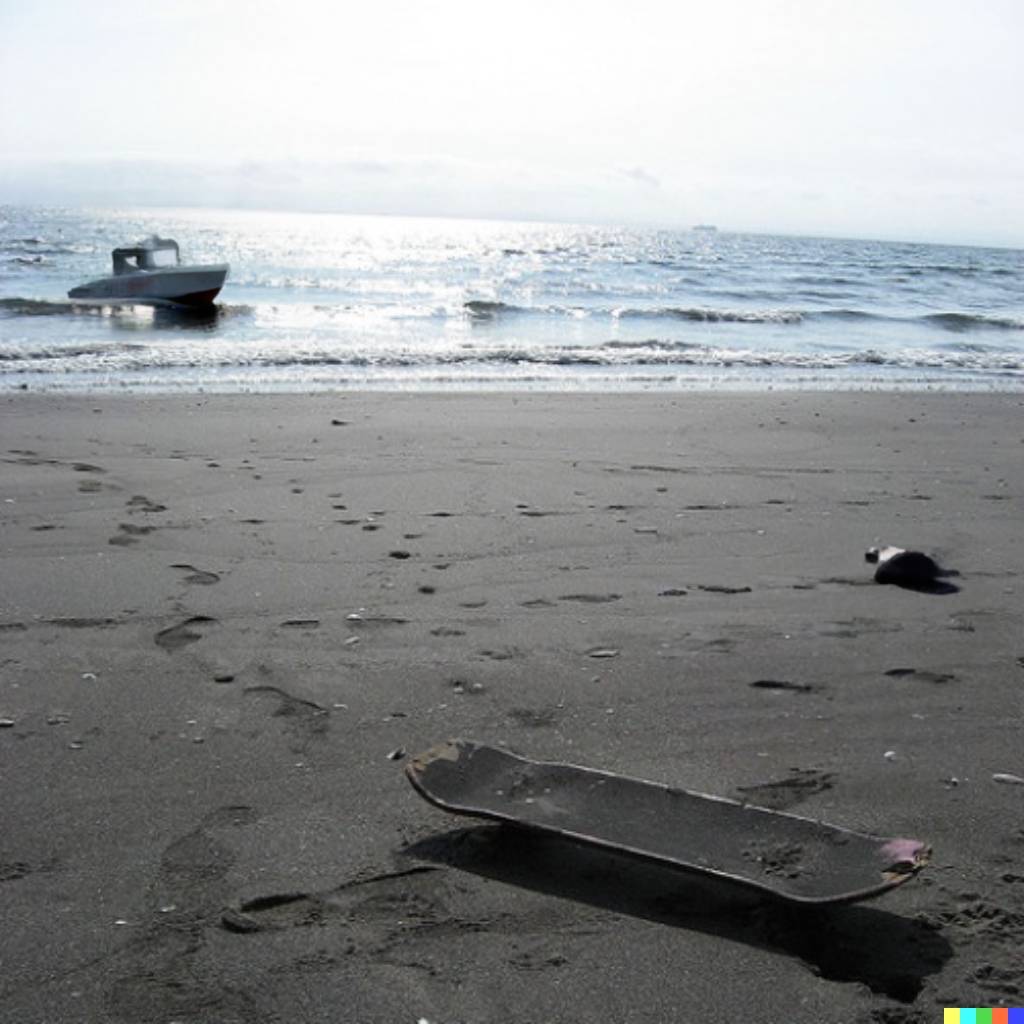}
        \caption*{GT}
  \end{minipage}

\caption{Qualitative editing results on PIE-Bench (top) and MagicBrush (bottom) benchmarks using different methods. Edit prompts are presented on top of each row. More examples available in supplementary material.}
\label{fig:user-study-images}
\end{figure*}

Figure \ref{fig:user-study-images}, second row, shows example edits generated by the participants. Additional examples are provided in the supplementary material. As shown, LDB produces targeted edits that integrate seamlessly with the images.

\subsubsection{Evaluation Survey Results}
The participants completed a three stage evaluation survey following the image editing tasks. The first part included a System Usability Scale (\textbf{SUS}) form to rate the usability, ease of use, design, and performance of each method. SUS is a standard usability evaluation survey widely used in user-experience literature \cite{brooke_susa_1996}.
Overall, participants indicated that they are more likely to use LDB compared to IP2P and SDI, and that they find it the easiest tool to use.
\textbf{LDB} obtained a SUS score of \textbf{80.35\%}, while \textbf{IP2P} and \textbf{SDI }achieved a SUS of \textbf{38.21\%} and \textbf{37.5\% }respectively. 

The SUS survey was followed by a Creativity Support Index \cite{cherry2014quantifying} survey to evaluate the system's degree of creative work support. Participants expressed positivity towards LDB, indicating that it enhanced their enjoyment, exploration, expressiveness, and immersion, while also deeming the results worth their effort. Lastly, the survey was followed by a semi-structured interview where participants appreciated the intuitiveness, ease of use and versatility of LDB. Further details about the study, interview, results, and discussion can be found in supplemental material.

    

\subsection{Quantitative Analysis}
To quantitatively evaluate the performance of LDB, we employed a comprehensive suite of metrics, aligning with established practices in image editing evaluation.

\begin{table*}[h]
\centering
\footnotesize
\setlength{\tabcolsep}{3pt}
\renewcommand{\arraystretch}{1.2}
\begin{adjustbox}{max width=\textwidth}
\begin{tabular}{l|c|ccc|cc|ccc|c}
\toprule
\multirow{2}{*}{\textbf{Benchmark}} & \multirow{2}{*}{\textbf{Method}} 
& \multicolumn{3}{c|}{\textbf{Image Quality}} 
& \multicolumn{2}{c|}{\textbf{Masked Region Preservation}} 
& \multicolumn{3}{c|}{\textbf{Text Alignment}} 
&  \textbf{Time (s)} \\
\hhline{|~|~|-|-|-|-|-|-|-|-|~|}
& & \textbf{IR} $_{\times 10} \uparrow$ & \textbf{HPS} $_{\times 10^2} \uparrow$ & \textbf{AS} $\uparrow$ 
  & \textbf{PSNR} $\uparrow$ & \textbf{LPIPS} $_{\times 10^2}  \downarrow$ 
  & \textbf{CS} $\uparrow$ & \textbf{CS-L} $\uparrow$ &\textbf{CS-D} $_{\times 10^2} \uparrow$ 
  & (per edit) $\downarrow$ \\
\midrule
\multirow{7}{*}{MagicBrush}
& IP2P & -62.83 & 21.16 & 5.29 & 7.28 & 15.07 & 29.39 & 22.01 & 6.64 & 1.72 \\
& SDI & -39.21 & 20.88 & 5.48 & 12.20 & 8.70 & 30.08 & 22.15 & 4.11 & 1.84 \\
& HDP & -20.69 & \textcolor{ForestGreen}{\textbf{23.27}} & 5.44 & 12.05 & \textcolor{ForestGreen}{\textbf{6.13}} & 31.01 & 22.06 & 9.89 & 12.85 \\
& BN & \textcolor{orange!80!black}{\textbf{-0.04}} & 22.57 & \textcolor{orange!80!black}{\textbf{5.73}} & 11.55 & 8.75 & \textcolor{ForestGreen}{\textbf{31.16}} & \textcolor{ForestGreen}{\textbf{22.17}} & \textcolor{ForestGreen}{\textbf{12.92}} & 7.49 \\
& BLD & -24.10 & \textcolor{orange!80!black}{\textbf{22.80}} & 5.48 & \textcolor{orange!80!black}{\textbf{12.64}} & \textcolor{orange!80!black}{\textbf{6.94}} & 30.64 & 21.99 & \textcolor{orange!80!black}{\textbf{10.05}} & \textcolor{orange!80!black}{\textbf{1.41}} \\
& GT & -1.93 & 22.62 & 5.36 & 17.64 & 2.30 & 30.75 & 22.14 & 9.78 & NA \\
& Ours & \textcolor{ForestGreen}{\textbf{7.74}} & 22.65 & \textcolor{ForestGreen}{\textbf{5.74}} & \textcolor{ForestGreen}{\textbf{12.85}} & 7.05 & \textcolor{orange!80!black}{\textbf{31.04}} & \textcolor{orange!80!black}{\textbf{22.07}} & 9.54 & \textcolor{ForestGreen}{\textbf{0.26}} \\

\midrule
\multirow{6}{*}{PIE-Bench}
& IP2P & -40.73 & 23.12 & 5.76 & 172.18 & 15.47 & 30.00 & \textcolor{orange!80!black}{\textbf{22.79}} & 14.27 & 1.83 \\
& SDI & 43.46 & 25.77 & 6.00 & \textcolor{orange!80!black}{\textbf{181.58}} & \textcolor{orange!80!black}{\textbf{3.89}} & 31.24 & 22.71 & 14.83 & 3.36 \\
& HDP & 39.02 & 25.92 & 6.01 & 178.84 & 4.62 & 31.08 & 22.73 & 16.20 & 13.44 \\
& BN & 72.77 & \textcolor{ForestGreen}{\textbf{26.66}} & \textcolor{orange!80!black}{\textbf{6.17}} & 177.07 & 8.67 & 31.50 & \textcolor{ForestGreen}{\textbf{22.80}} & \textcolor{orange!80!black}{\textbf{16.88}} & 7.51 \\
& BLD & \textcolor{orange!80!black}{\textbf{50.68}} & 26.36 & 6.11 & 180.85 & 4.19 & \textcolor{orange!80!black}{\textbf{31.35}} & 22.78 & \textcolor{ForestGreen}{\textbf{17.22}} & \textcolor{orange!80!black}{\textbf{1.47}} \\
& Ours & \textcolor{ForestGreen}{\textbf{86.02}} & \textcolor{orange!80!black}{\textbf{26.60}} & \textcolor{ForestGreen}{\textbf{6.51}} & \textcolor{ForestGreen}{\textbf{184.57}} & \textcolor{ForestGreen}{\textbf{1.91}} & \textcolor{ForestGreen}{\textbf{31.66}} & 22.76 & 16.74 & \textcolor{ForestGreen}{\textbf{0.25}} \\
\bottomrule
\end{tabular}
\end{adjustbox}
\vspace{-0.2cm}
\caption{
Quantitative results on MagicBrush and PIE-Bench.
Metrics are grouped into Image Quality, Masked Region Preservation, and Text Alignment.
$\uparrow$ indicates higher is better; $\downarrow$ indicates lower is better. 
The \textbf{\textcolor{ForestGreen}{best}} and \textbf{\textcolor{orange!80!black}{second-best}} scores are highlighted.
}
\label{tab:comparison}
\end{table*}

Specifically, for text-image alignment, we used CLIP Score (CS) \cite{radford2021learning} for global alignment, CS-L for masked-region alignment, and CS-D \cite{gal2022stylegan} for consistency between image and caption changes in CLIP space.

We adopted Learned Perceptual Image Patch Similarity (\textbf{LPIPS}) \cite{zhang2018unreasonable} and Peak Signal-to-Noise Ratio (\textbf{PSNR}) \cite{hore2010image} for evaluating content preservation and pixel-level fidelity in unmasked regions.  
Furthermore, to gauge overall image quality and aesthetic appeal, we incorporated Aesthetic Score (\textbf{AS}) \cite{schuhmann2022laion}, Image Reward (\textbf{IR}), and Human Preference Score V2 (\textbf{HPS}) \cite{wu2023human}, the latter two reflecting human-aligned preferences. 

We compared \textbf{LDB} against a diverse set of state-of-the-art editing and inpainting methods, including InstructPix2Pix (\textbf{IP2P}) \cite{brooks2022instructpix2pix}, Stable Diffusion Inpainting (\textbf{SDI}) \cite{rombach2022high}, HD-Painter (\textbf{HDP}) \cite{manukyan2023hd}, BrushNet (\textbf{BN}) \cite{ju2024brushnet}, and Blended Latent Diffusion (\textbf{BLD}) \cite{avrahami2023blended}, on two benchmarks: MagicBrush \cite{zhang2024magicbrush} and PIE-Bench \cite{ju2023direct}. For MagicBrush, we also report results on the provided ground truth (\textbf{GT}) images.

Quantitative results are summarized in \cref{tab:comparison}. All methods were evaluated using their default editing settings, except for LDB, IP2P, and SDI on the MagicBrush benchmark, where we used user-edited images from our user study for consecutive edits. Inference times denote average per-edit durations, measured on a single NVIDIA RTX 4090 GPU with $N=25$ diffusion steps for baseline methods and $n=8$ for LDB.

\subsection{Ablation Study}
\label{abl}
We perform three ablation studies for two main components of the LDB caching mechanism, \ie the caching timesteps for the regeneration latent ($r$), and the blending latent ($b$). We also ablate and discuss the effect of strength control $\alpha$ and its relationship with $n$ in supplementary material.

\subsubsection{Ablation on Regeneration Latent Step}
\begin{figure*}
    \centering
    \includegraphics[width=0.9\linewidth]{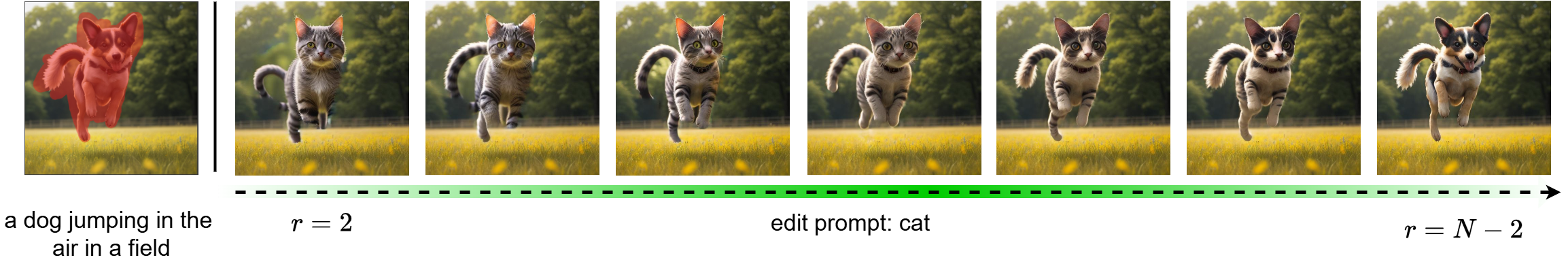}
\caption{Ablation study on regeneration latent step $r$ (increasing left to right). Small $r$ results in strong prompt adherence (``cat'') but introduces artifacts. Large $r$ (near $N$) leads to insufficient modification, retaining the original ``dog''. An intermediate $r$ achieves the best balance of edit fidelity and background preservation.}
\label{fig:abl}
\end{figure*}

The timestep $r$ for caching the regeneration latent is critical, as it dictates the extent of possible modifications during the regeneration process. We performed an ablation study by varying $r$ while holding the total diffusion steps $N$ constant. This variation in $r$ implicitly changes the number of regeneration steps ($n$) and necessitates adjustments to the strength parameter accordingly. Qualitatively, as shown in \cref{fig:abl}, excessively small $r$ values lead to incoherent edits and noticeable artifacts due to insufficient blending with the original image. Conversely, large $r$ values limit the model's ability to modify the masked region, resulting in minimal changes and preserving the original content. 

Quantitatively, we observe that smaller $r$ steps (\eg $r=2$) yield higher LPIPS (0.04) and low PSNR (27.03), indicating poor image quality and fidelity.  Edit fidelity scores such as CS-L also confirm that larger $r$ steps result in lower scores (22.98), suggesting ineffective edits within the masked region. The HPS index demonstrates a higher score for mid-range steps (0.33, $r=12$) compared to both ends of the spectrum (0.29, $r=23$), highlighting a performance sweet spot for intermediate $r$ values. Detailed metric graphs are available in the supplemental material.

\begin{figure*}
    \centering
    \includegraphics[width=0.85\linewidth]{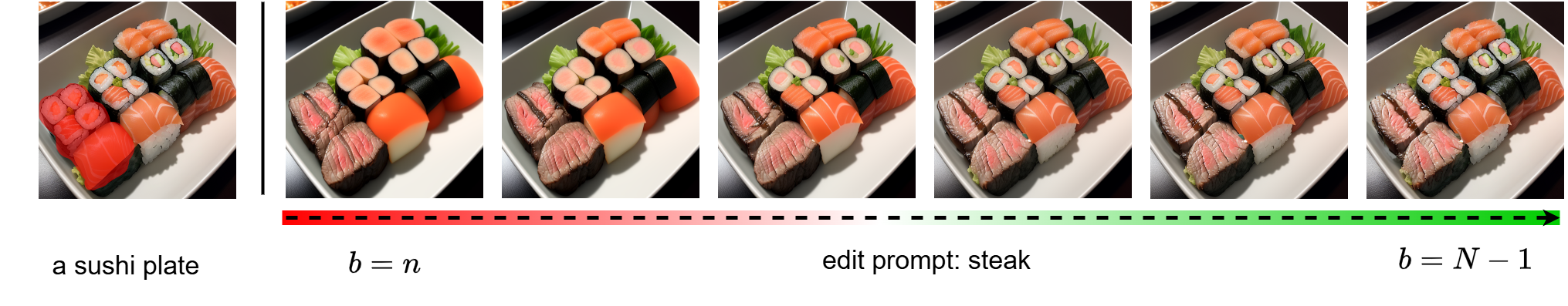}
\caption{Ablation study on blending latent step $b$ (increasing left to right). The prompt ``steak'' is applied to an image of ``sushi plate'' while increasing $b$ from left to right. At $b=n$ (left), the edit disrupts the original structure, affecting unmasked regions. As $b$ approaches $N$ (right), background preservation improves, and edits blend more seamlessly.}
    \label{fig:abl_b}
\end{figure*}

\subsubsection{Ablation on Blending Latent Step}
The blending latent step, controlled by the parameter $b$, determines when the cached regeneration latent is blended back into the diffusion process and is crucial for seamless integration of the edited region with the original image and preserving background. We conduct an ablation study by varying $b$ while keeping $r$ and $N$ fixed.  \cref{fig:abl_b}  qualitatively demonstrates the effect of different $b$ values.

When $b$ is small, the blending process starts prematurely, causing the edit to bleed into the background and distorting the original image context.  Conversely, larger $b$ values, representing late blending, effectively preserve the background integrity while still allowing for meaningful edits within the masked region. 

Quantitatively, smaller $b$ values ($b=n$) lead to higher LPIPS (0.17) and lower PSNR (11.61), indicating worse background preservation. Edit fidelity scores (CS-L) within the masked remained stable across the spectrum while CS-D improves at larger $b$ (0.32 at $b = N{-}1$), reflecting better edit alignment. These findings indicate that later blending is preferable, leading us to select $b=N-2$ in the LDB algorithm to prioritize background preservation while maintaining effective localized editing. Further details and metric plots are available in the supplementary material.

%% file: sec/5_discussion.tex
\section{Discussion}
Our experiments demonstrate that LDB establishes new benchmarks for speed and workflow adaptability in diffusion-based image editing. Key findings include:

\textbf{Enhanced Control via Layering}: LDB's layered design enables creating non-destructive refinements as well as iterative complex compositions. Participants highlighted how this mirrors professional editing tools like Photoshop \cite{adobephotoshop}.

\textbf{Speed and Efficiency}: LDB achieves remarkable speed, 53$\times$ faster than BrushNet (evaluated on the same hardware), crucial for interactive editing. We observe that reducing diffusion steps to as few as $n=4$ maintains reasonable quality (HPS: 0.34, CS-D: 0.35), yielding a latency of \textbf{140ms} per edit. User studies confirm \textit{instant feedback} as a key advantage, enabling rapid iteration (tens of variations per minute \vs 1-2 for baselines). This speed results from efficient latent caching (\cref{caching}), minimizing computation and memory overhead ($\sim$1.25 MB for 10 layers).

\textbf{Quantitative Performance}: LDB demonstrates a superior combination of speed, image quality, and edit fidelity across both benchmarks. On the PIE-Bench dataset, LDB achieves the best performance in six key metrics, excelling in human preference (HPS = 86.02), background preservation (LPIPS = 1.91), and text alignment (CS = 31.66), while also being the fastest method by a significant margin. This highlights its ability to generalize across a diverse set of editing tasks while maintaining high speed. Similarly, on the MagicBrush benchmark, LDB delivers strong performance with highest score in crucial metrics such as IR, AS, and PSNR. While BrushNet shows a slight advantage in some text alignment metrics, its practical usability is hindered by substantially slower runtime. 

\subsection{Limitations and Future Work}
Brush strength ($\alpha$) and diffusion step count ($n$) coupling (\cref{fig:alpha}) still requires minor user tuning across scenarios. Although preset profiles partially address this, future work could explore adaptive parameter tuning mechanisms to further improve usability.
Moreover, semantically implausible edits (\eg placing a boat in the sky) remain challenging due to inherent biases within diffusion models. Integrating techniques like semantic guidance could expand plausible edit ranges.
Finally, responsible deployment necessitates robust watermarking~\cite{fernandez2023stable} and provenance tracking to mitigate misuse and ensure transparency.

\subsection{Broader Applications}
LDB's training-free design only requires a standard iterative denoising process, which allows seamless integration into diverse diffusion models and applications requiring rapid editing.
We validated this by integrating LDB to other commonly used methods, including DiT-based text-to-image (\eg, PixArt‑$\alpha$ \cite{chen2023pixart}) and video generation models ~\cite{blattmann2023stablevideo} without any model-specific tuning.

Traditional diffusion-based video editing typically propagates edits from the first frame using additional supervision (\eg optical flow~\cite{liang2024flowvid}), risking temporal inconsistencies. LDB's high fidelity background preservation and efficiency naturally address these issues.

We demonstrate preliminary success integrating LDB with Stable Video Diffusion (SVD)~\cite{blattmann2023stablevideo}, editing the first frame and applying LDB's latent caching across frames for fast consecutive edits (see supplementary material, Fig. 15). This approach opens avenues for accelerated video manipulation, 3D asset editing, and collaborative design platforms.

\subsection{Conclusion}
LDB reimagines diffusion-based editing through latent caching and non-destructive layering, achieving unmatched  speed and control. Quantitative results and user study show superior performance in image preference, edit fidelity, time, and usability. By bridging interactive editing with high-fidelity generative models, LDB can empower artists to iterate fluidly while maintaining artistic intent.

%% file: sec/X_suppl.tex
\clearpage
\setcounter{page}{1}
\maketitlesupplementary
\section{UI and interaction design}
\cref{ui} provides an overview of the user interface. As demonstrated, the UI comprises the following primary sections (each section highlighted with the corresponding number on the image):
\begin{figure*}
  \centering
  
        \includegraphics[width=0.99\textwidth]{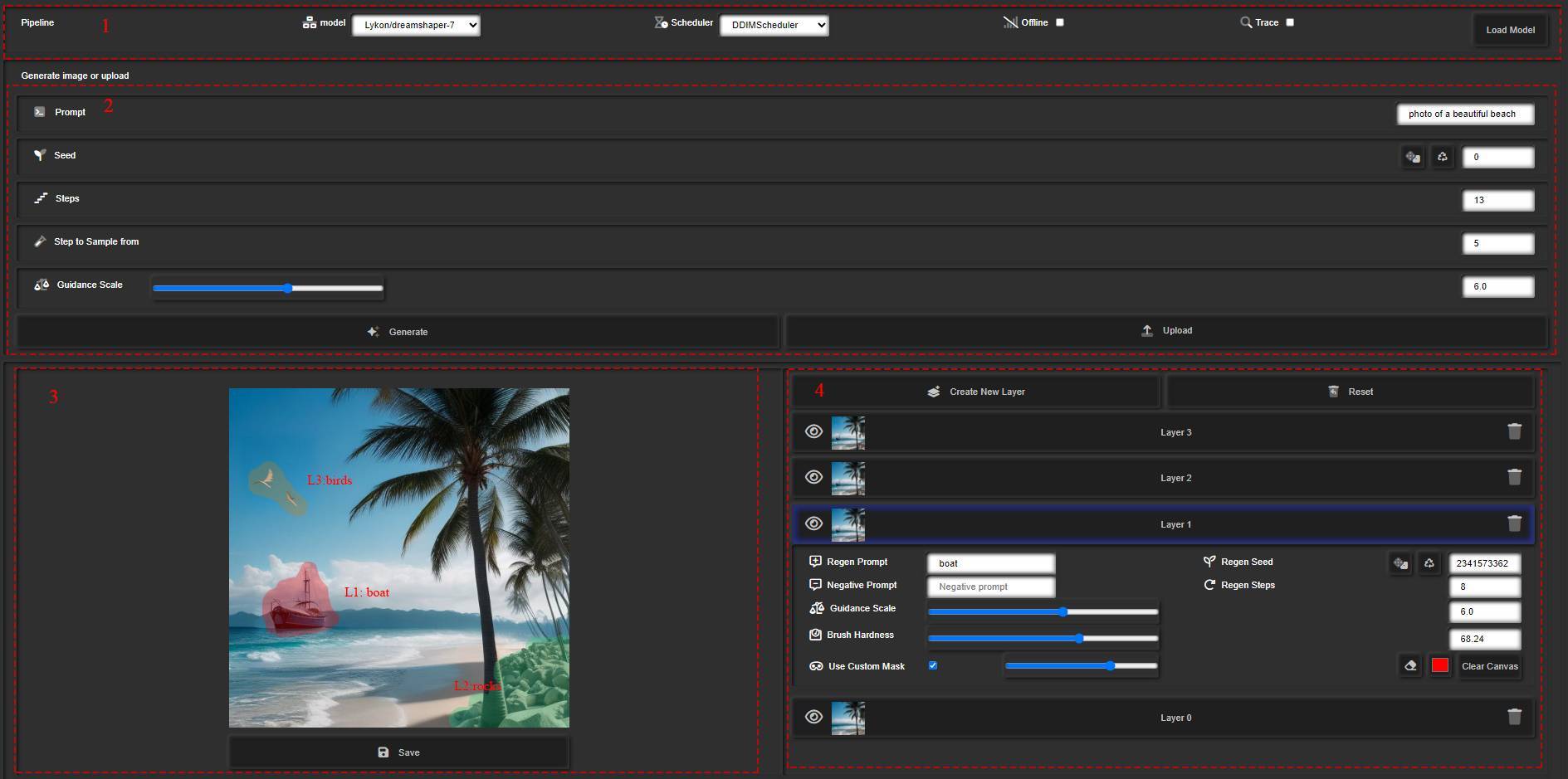}
\caption{
Design of the LDB's UI:
The name and functionality of each section are described in the text. In this example, the user has created three layers, visualized on the image canvas, along with the mask and edit prompt. The selected layer in this picture is Layer 1.}

\label{ui}
\end{figure*}

\begin{enumerate}
    \item Model Section
    \begin{itemize}
        \item This section in the UI enables users to load various model combinations, including pre-trained models, schedulers, and LoRA \cite{hu2021lora}.
    \end{itemize}
    \item Generation section
    \begin{itemize}
        \item This section allows users to either generate a new image using a seed and prompt combination, or upload a real image that will be inverted.
    \end{itemize}
    \item Image Canvas
    \begin{itemize}
        \item This canvas serves as the workspace where users interact with and make edits to images.
    \end{itemize}
    \item Editing Section
    \begin{itemize}
        \item This section provides controls to create and modify different layers to make the desired edits.
    \end{itemize}
\end{enumerate}

We provide the ability to stack and hide/unhide layers, similar to traditional image-editing tools.

When editing a layer, we provide the choice of box mode or brush mode. In box mode, the mask is a square shape controlled by the ``brush size'' parameter. As the box is dragged around the image, the seed value will automatically increment, providing a continuous stream of new edits. The user may stop dragging when a suitable edit is seen.

In brush mode, the mask is an arbitrary shape that can be added to or subtracted from using a circular brush tool. The size of the brush is controlled by the ``brush size'' parameter. In this mode, the user can scroll a mouse wheel or use a scrolling gesture to increment or decrement the seed, allowing them to rapidly explore the space of potential edits and return to any edit that appears suitable.


\subsection{Hyperparameters}

To provide a balance between usability and complexity, we provide control over a number of hyperparameters: number of regeneration steps, ``brush strength'', brush size and seed number. Each hyperparameter is designed to be largely orthogonal to the other parameters, enabling them to independently affect the appearance of the edit without the need to simultaneously adjust multiple inputs.

\begin{itemize}

\item \textbf{Number of regeneration steps ($n$)}: An integer value that specifies the number of steps LDB will run to make the edit. Changing $n$ effectively changes the strength of the modification as well as the processing time.
\item \textbf{Brush Strength ($\alpha$)}: A number that indirectly controls the $\alpha$ value in (\cref{alpha}) which controls how strong the initial noise pattern should be. The user-specified alpha, $\alpha^*$, has a value between 0 and 100, which will be scaled using the following equation:
\begin{equation}
    \alpha = \frac{\sqrt{\left| \frac{\alpha^*}{100} \cdot \left( \sigma - 2 \cdot \frac{\text{Cov}(Z_{r}^{(k)}, Z^{\prime}_0)}{\text{Var}(Z_{r}^{(k)})} \right) \right|}}{\sqrt{\frac{\sum_{i=1}^{W} \sum_{j=1}^{H} [m_{ij} \neq 0]}{W}}}
    \label{alpha}
\end{equation}

where $Z^{\prime}_0$ and $Z_{r}^{(k)}$ are the new noise latent and latent for regeneration respectively (as noted in \cref{alg:regen}), $\sigma$ is the acceptable range for the variance of the $Z_{r}^{(k)}$ (we used $\sigma$=0.25), $m$ is the corresponding mask, and $W$ is the width of $Z_{n_k}$ (W=512).

This formula is designed to ensure that any fixed value of the user-provided $\alpha^*$ value produces similar effects on the image even as the number of regeneration steps or the brush size/mask size are changed, thus making it more logically independent from the other parameters.

\item \textbf{Seed Number ($s'$)}: An integer number that will be used for generating the Gaussian noise pattern in the specified region. As with normal image generation, the UI provides buttons to randomize the seed or reuse the previous seed. Moving the box around (in box mode) or using the scroll wheel (in custom mask mode) will adjust the seed automatically.
\item \textbf{Brush Size $d$}: An integer value that dictates the radius of the box when utilized in box mode, or the size of the brush in custom mask mode (in pixels). 

\end{itemize}

\section{Additional Qualitative Examples}
\begin{figure*}[!ht]
  \centering
        \begin{minipage}[b]{0.99\textwidth}
      \centering
    \footnotesize Layer 1: \textit{``starry night - van gogh style'' }
  \end{minipage}
    \begin{minipage}[b]{0.19\textwidth}
    \includegraphics[width=\textwidth]{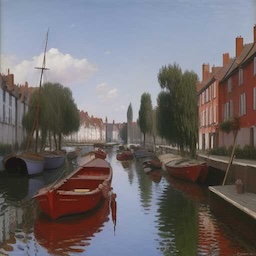}
  \end{minipage}
    \begin{minipage}[b]{0.19\textwidth}
    \includegraphics[width=\textwidth]{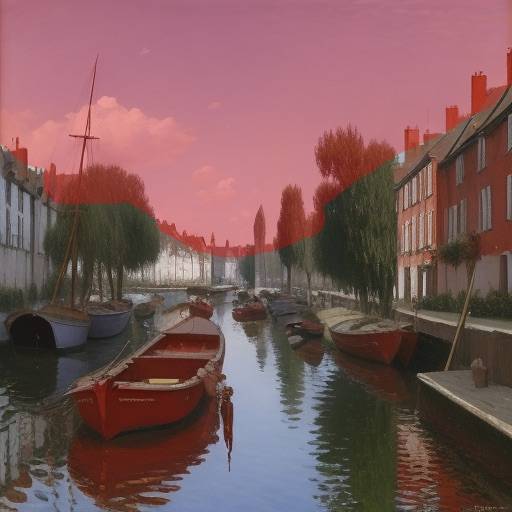}
  \end{minipage} \rulesep
          \begin{minipage}[b]{0.19\textwidth}
    \includegraphics[width=\textwidth]{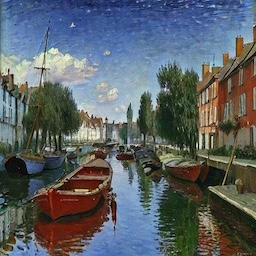}
  \end{minipage}
    \begin{minipage}[b]{0.19\textwidth}
    \includegraphics[width=\textwidth]{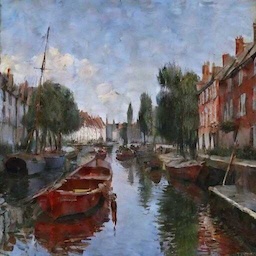}
  \end{minipage}
          \begin{minipage}[b]{0.19\textwidth}
    \includegraphics[width=\textwidth]{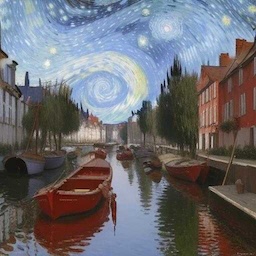}
  \end{minipage}
        \begin{minipage}[b]{0.99\textwidth}
      \centering
    \footnotesize Layer 1: \textit{``cat'' }$\mid$ Layer 2 \textit{``Jennifer Aniston''}
  \end{minipage}
   \begin{minipage}[b]{0.19\textwidth}
    \includegraphics[width=\textwidth]{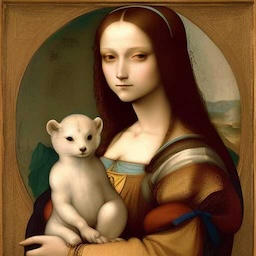}
  \end{minipage} 
    \begin{minipage}[b]{0.19\textwidth}
    \includegraphics[width=\textwidth]{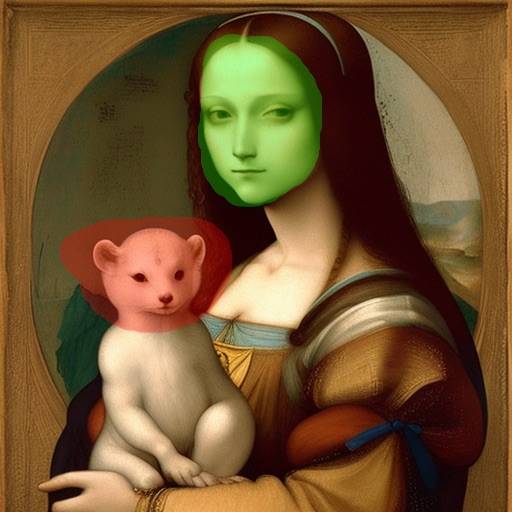}
  \end{minipage} \rulesep
          \begin{minipage}[b]{0.19\textwidth}
    \includegraphics[width=\textwidth]{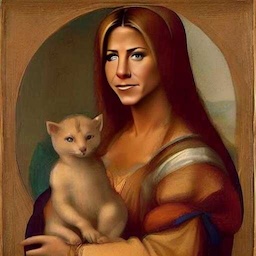}
  \end{minipage}
    \begin{minipage}[b]{0.19\textwidth}
    \includegraphics[width=\textwidth]{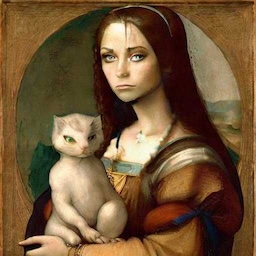}
  \end{minipage}
          \begin{minipage}[b]{0.19\textwidth}
    \includegraphics[width=\textwidth]{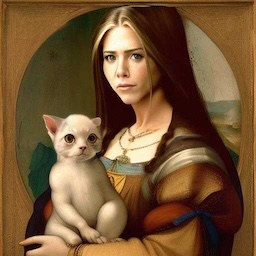}
  \end{minipage}

          \begin{minipage}[b]{0.99\textwidth}
      \centering
    \footnotesize Layer 1: \textit{``sunset, watercolor style'' }$\mid$ Layer 2 \textit{``boat''}
  \end{minipage}
    \begin{minipage}[b]{0.19\textwidth}
    \includegraphics[width=\textwidth]{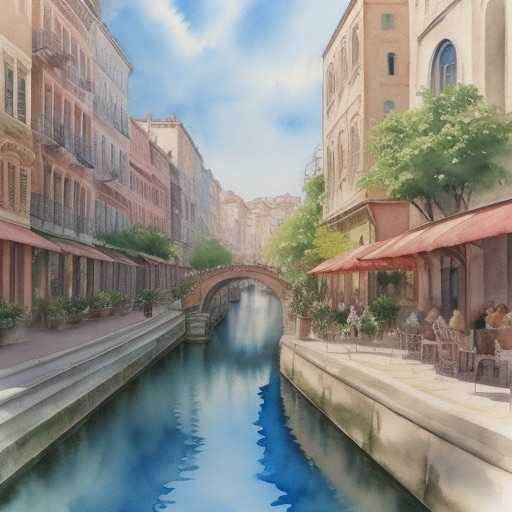}
  \end{minipage} 
    \begin{minipage}[b]{0.19\textwidth}
    \includegraphics[width=\textwidth]{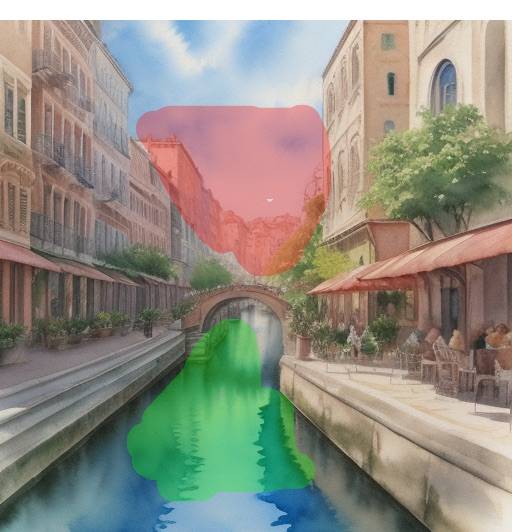}
  \end{minipage} \rulesep
          \begin{minipage}[b]{0.19\textwidth}
    \includegraphics[width=\textwidth]{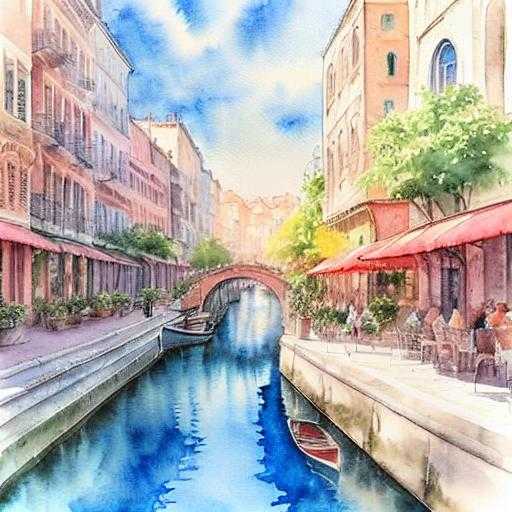}
  \end{minipage}
    \begin{minipage}[b]{0.19\textwidth}
    \includegraphics[width=\textwidth]{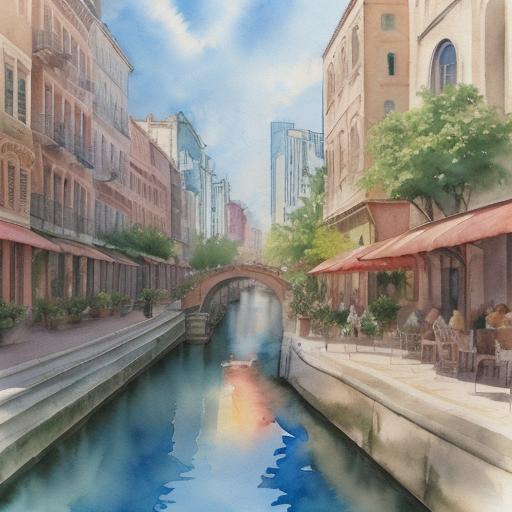}
  \end{minipage}
          \begin{minipage}[b]{0.19\textwidth}
    \includegraphics[width=\textwidth]{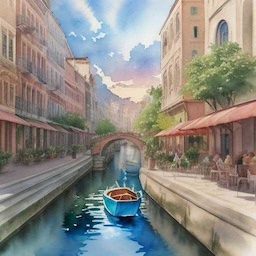}
  \end{minipage}

        \begin{minipage}[b]{0.99\textwidth}
      \centering
    \footnotesize Layer 1: \textit{``red pool ball'' }
  \end{minipage}
    \begin{minipage}[b]{0.19\textwidth}
    \includegraphics[width=\textwidth]{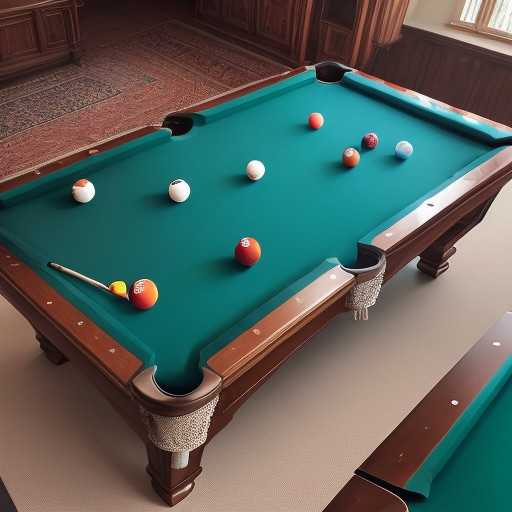}
    \caption*{input image}
  \end{minipage}
    \begin{minipage}[b]{0.19\textwidth}
    \includegraphics[width=\textwidth]{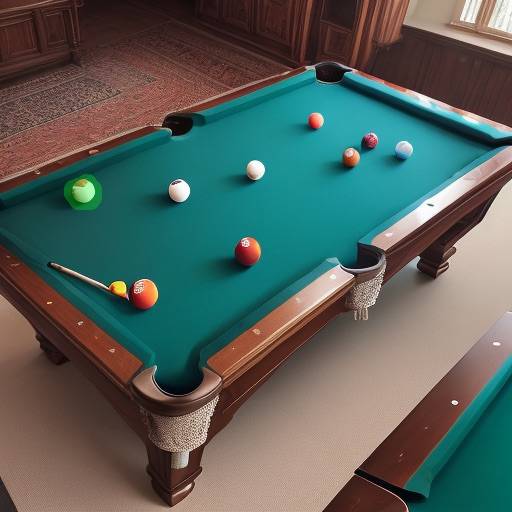}
    \caption*{editing mask}
  \end{minipage} \rulesep
          \begin{minipage}[b]{0.19\textwidth}
    \includegraphics[width=\textwidth]{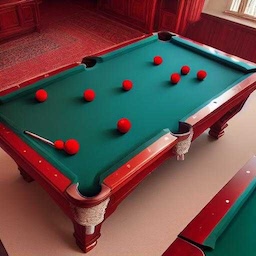}
    \caption*{IP2P}
  \end{minipage}
    \begin{minipage}[b]{0.19\textwidth}
    \includegraphics[width=\textwidth]{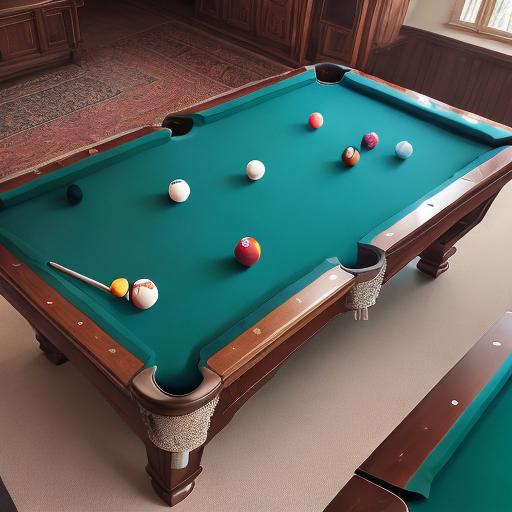}
    \caption*{SDI}
  \end{minipage}
          \begin{minipage}[b]{0.19\textwidth}
    \includegraphics[width=\textwidth]{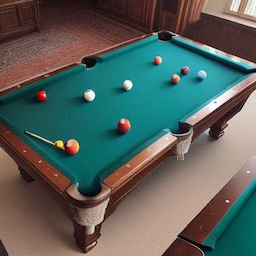}
    \caption*{\textbf{LDB (ours)}}
  \end{minipage}
  
\caption{
Qualitative results for the freeform part of the user study (Type 1 tasks)}

\label{freeform}
\end{figure*}

\cref{freeform} and \cref{fig:additionalmagic} present examples of Type 1 tasks (freeform) and Type 2 tasks (MagicBrush) respectively. All the images were edited by participants during the user study.

\begin{figure*}[!ht]
  \centering
        \begin{minipage}[b]{0.99\textwidth}
      \centering
     \textit{``barbie doll'' }
  \end{minipage}
          \begin{minipage}[b]{0.10\textwidth}
    \includegraphics[width=\textwidth]{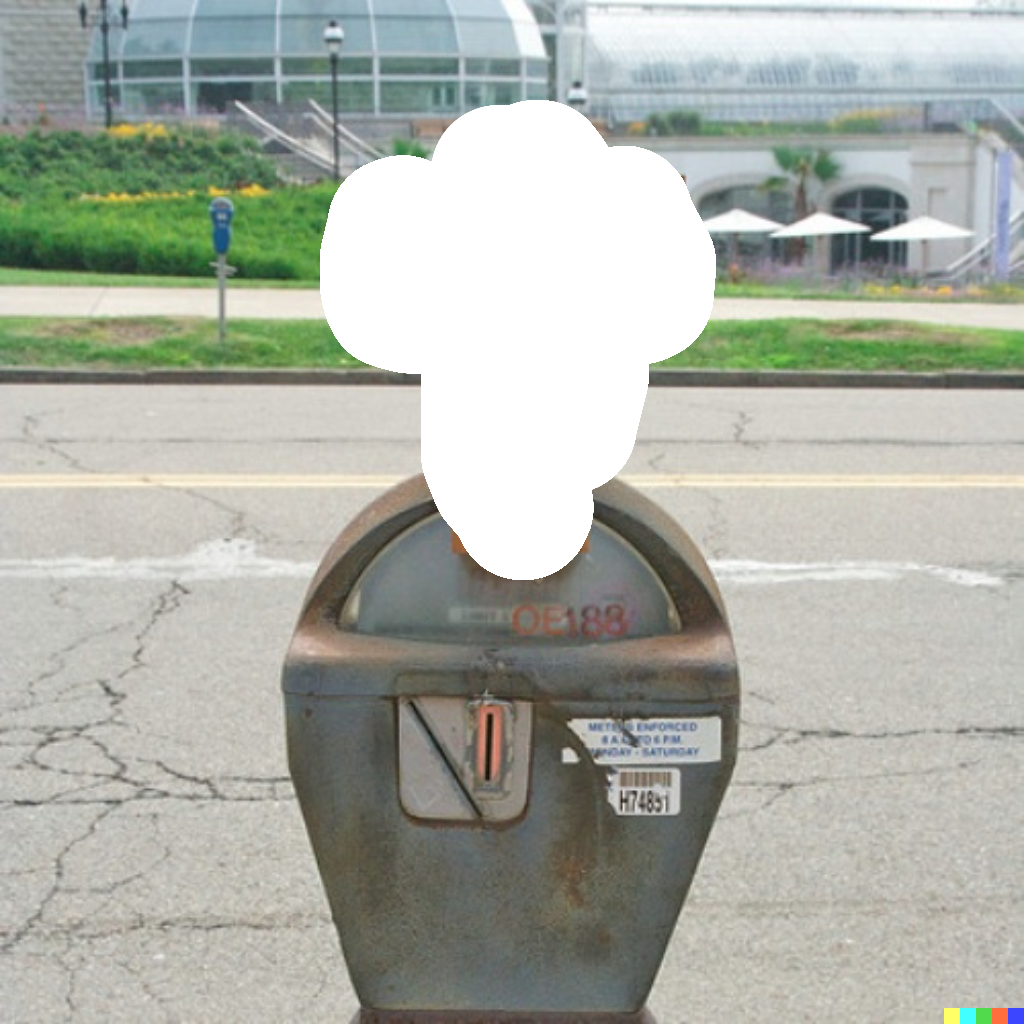}
      \end{minipage} 
          \begin{minipage}[b]{0.10\textwidth}
    \includegraphics[width=\textwidth]{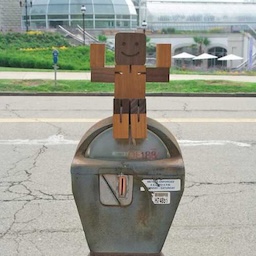}
      \end{minipage} 
          \begin{minipage}[b]{0.10\textwidth}
    \includegraphics[width=\textwidth]{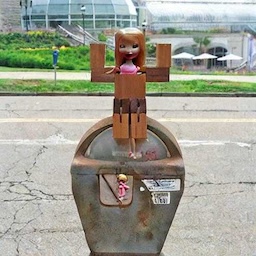}
  \end{minipage} 
    \begin{minipage}[b]{0.105\textwidth}
    \includegraphics[width=\textwidth]{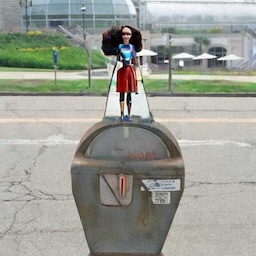}
  \end{minipage}
      \begin{minipage}[b]{0.105\textwidth}
    \includegraphics[width=\textwidth]{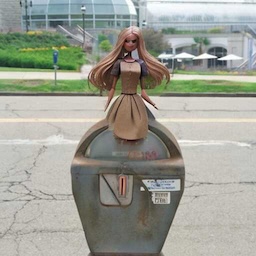}
  \end{minipage}
        \begin{minipage}[b]{0.105\textwidth}
    \includegraphics[width=\textwidth]{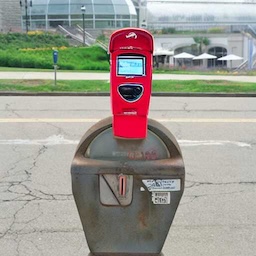}
  \end{minipage}
          \begin{minipage}[b]{0.105\textwidth}
    \includegraphics[width=\textwidth]{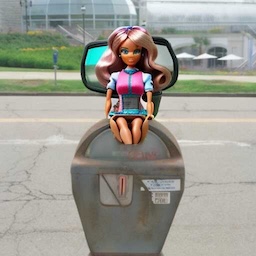}
  \end{minipage}
       \begin{minipage}[b]{0.105\textwidth}
      
    \includegraphics[width=\textwidth]{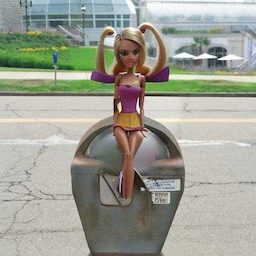}
  \end{minipage}
      \begin{minipage}[b]{0.105\textwidth}
    \includegraphics[width=\textwidth]{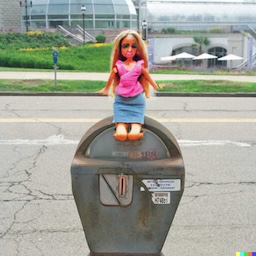}
  \end{minipage}

        \begin{minipage}[b]{0.99\textwidth}
      \centering
    \textit{``saddles''}
  \end{minipage}

      \begin{minipage}[b]{0.105\textwidth}
    \includegraphics[width=\textwidth]{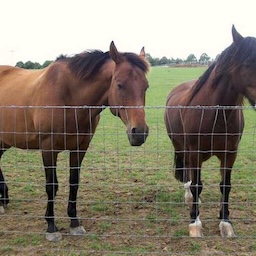}
  \end{minipage} 
          \begin{minipage}[b]{0.105\textwidth}
    \includegraphics[width=\textwidth]{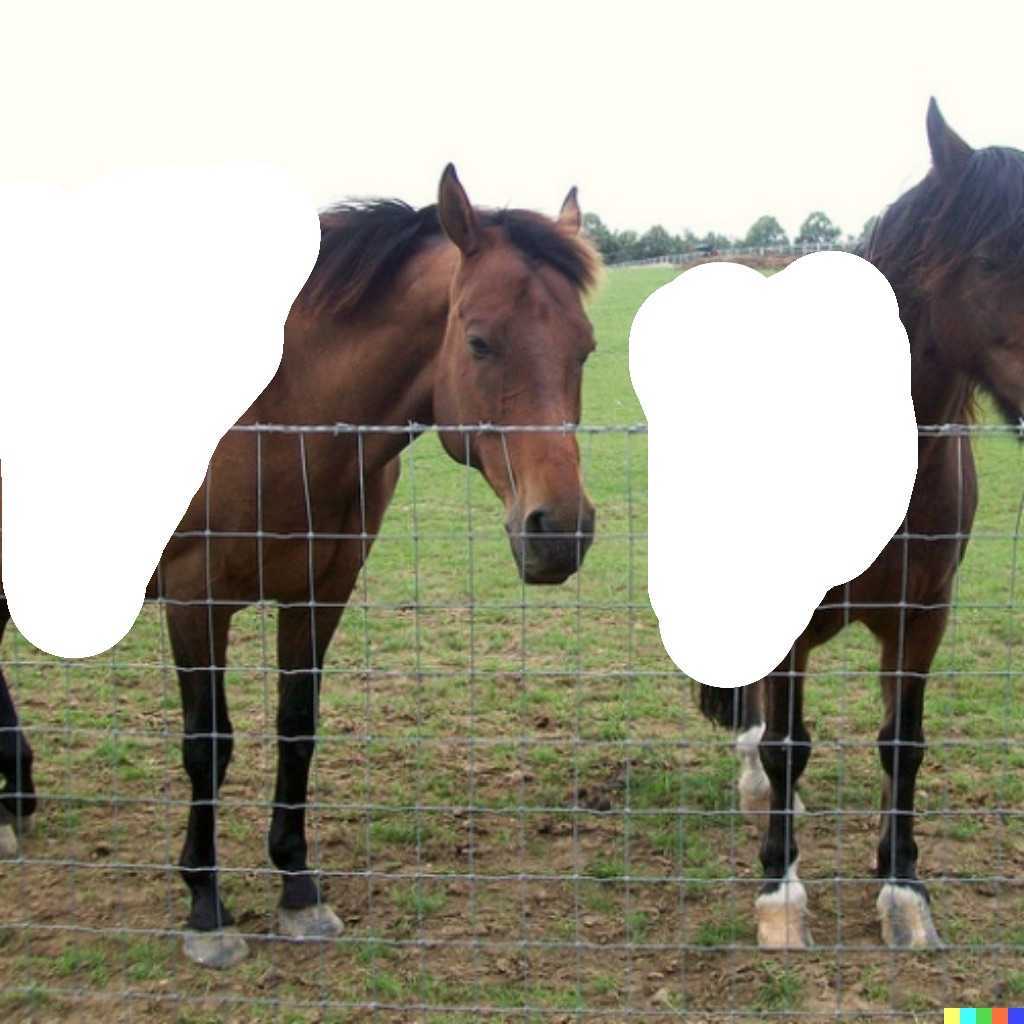}
  \end{minipage}\rulesep
    \begin{minipage}[b]{0.105\textwidth}
    \includegraphics[width=\textwidth]{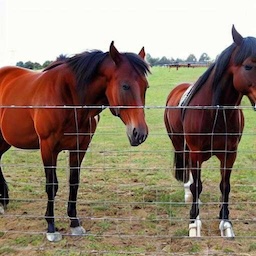}
  \end{minipage}
    \begin{minipage}[b]{0.105\textwidth}
    \includegraphics[width=\textwidth]{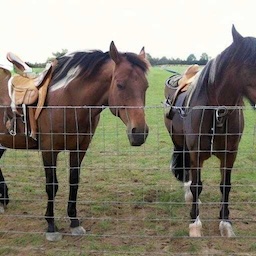}
  \end{minipage}
        \begin{minipage}[b]{0.105\textwidth}
    \includegraphics[width=\textwidth]{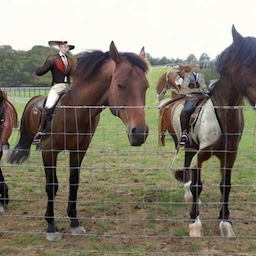}
  \end{minipage}
        \begin{minipage}[b]{0.105\textwidth}
    \includegraphics[width=\textwidth]{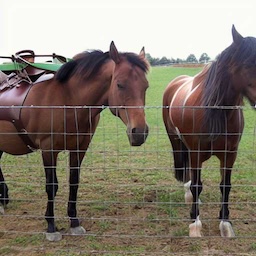}
  \end{minipage}
      \begin{minipage}[b]{0.105\textwidth}
    \includegraphics[width=\textwidth]{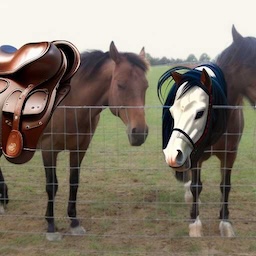}
  \end{minipage}
      \begin{minipage}[b]{0.105\textwidth}
    \includegraphics[width=\textwidth]{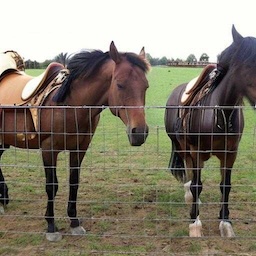}
  \end{minipage}
      \begin{minipage}[b]{0.105\textwidth}
    \includegraphics[width=\textwidth]{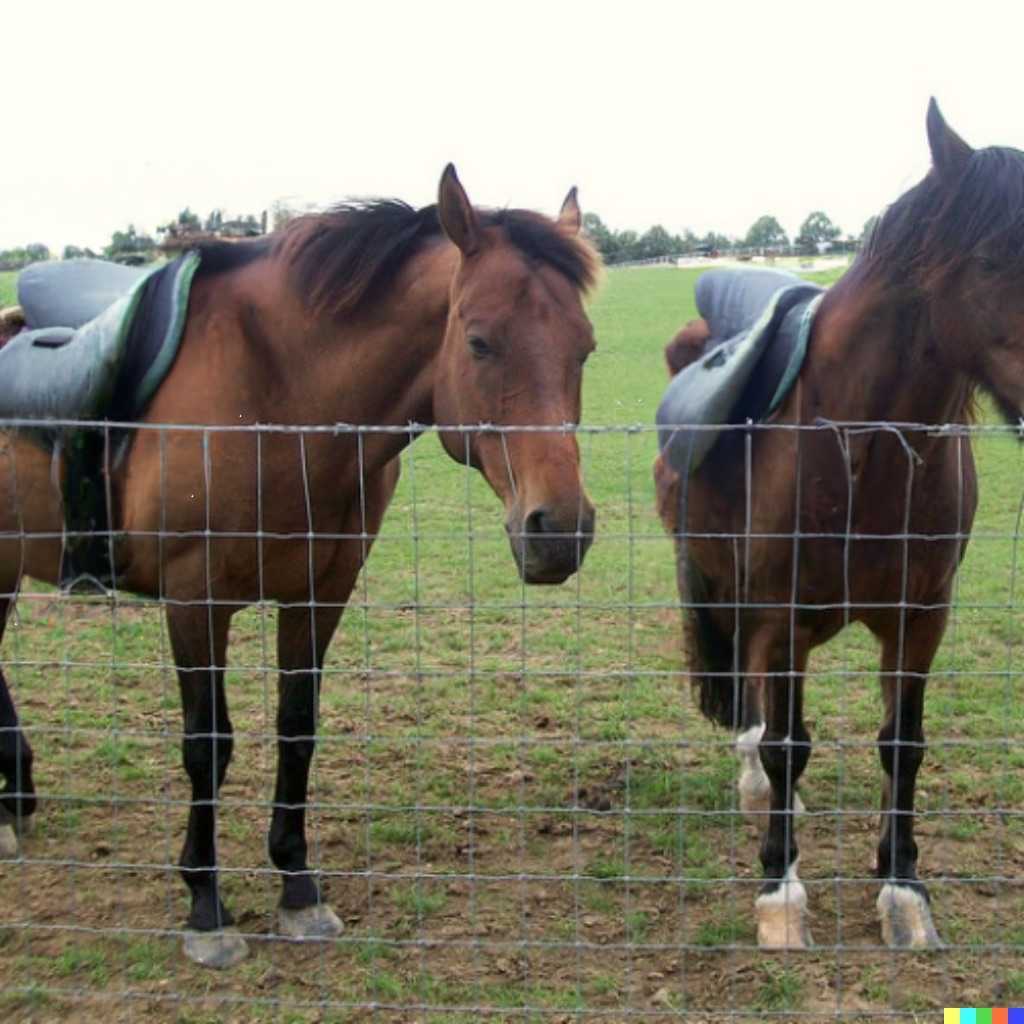}
  \end{minipage}

        \begin{minipage}[b]{0.99\textwidth}
      \centering
    \textit{``green bird perched on tree''}
  \end{minipage}

      \begin{minipage}[b]{0.105\textwidth}
    \includegraphics[width=\textwidth]{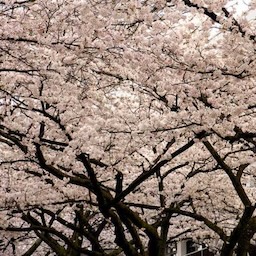}
  \end{minipage} 
          \begin{minipage}[b]{0.105\textwidth}
    \includegraphics[width=\textwidth]{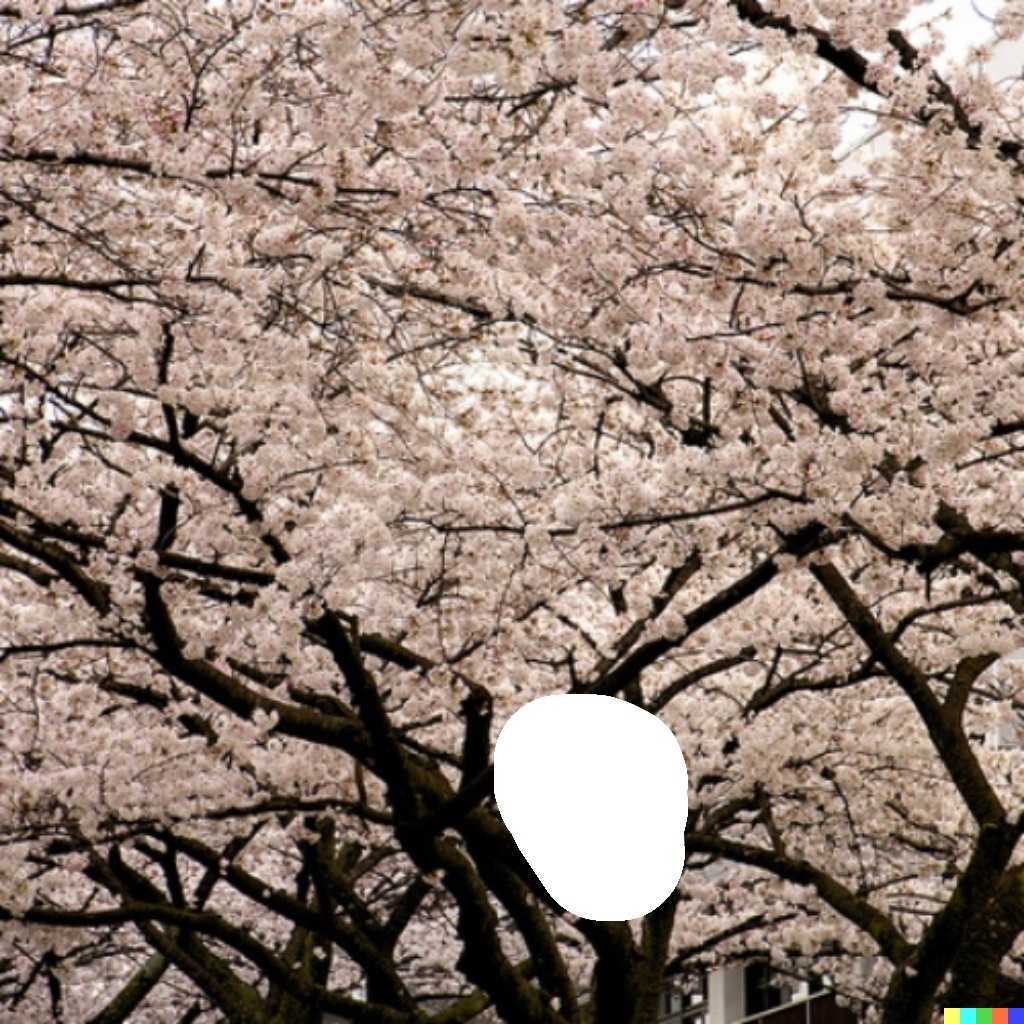}
  \end{minipage}\rulesep
    \begin{minipage}[b]{0.105\textwidth}
    \includegraphics[width=\textwidth]{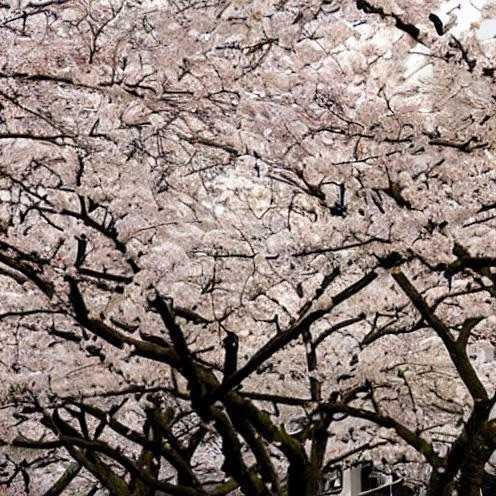}
  \end{minipage}
    \begin{minipage}[b]{0.105\textwidth}
    \includegraphics[width=\textwidth]{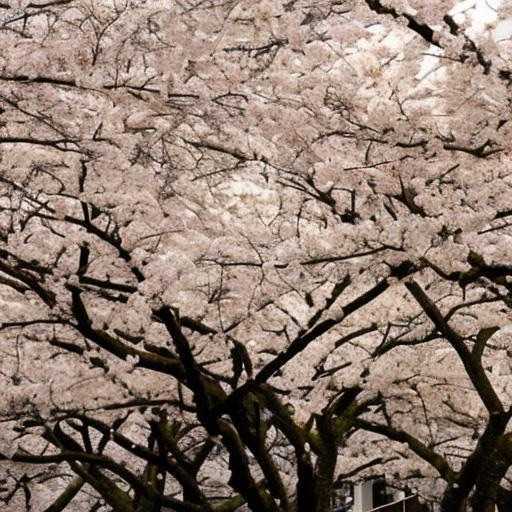}
  \end{minipage}
        \begin{minipage}[b]{0.105\textwidth}
    \includegraphics[width=\textwidth]{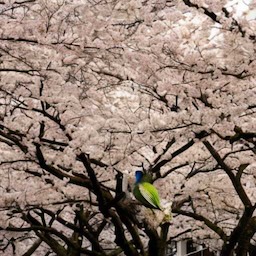}
  \end{minipage}
        \begin{minipage}[b]{0.105\textwidth}
    \includegraphics[width=\textwidth]{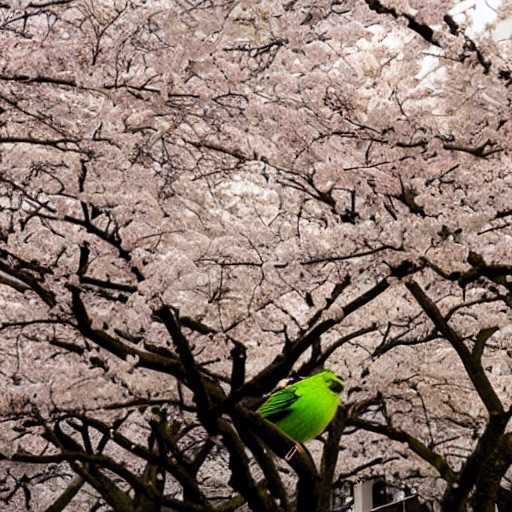}
  \end{minipage}
      \begin{minipage}[b]{0.105\textwidth}
    \includegraphics[width=\textwidth]{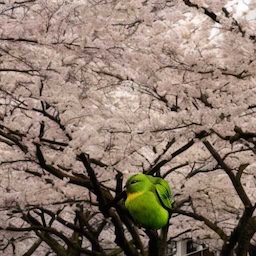}
  \end{minipage}
      \begin{minipage}[b]{0.105\textwidth}
    \includegraphics[width=\textwidth]{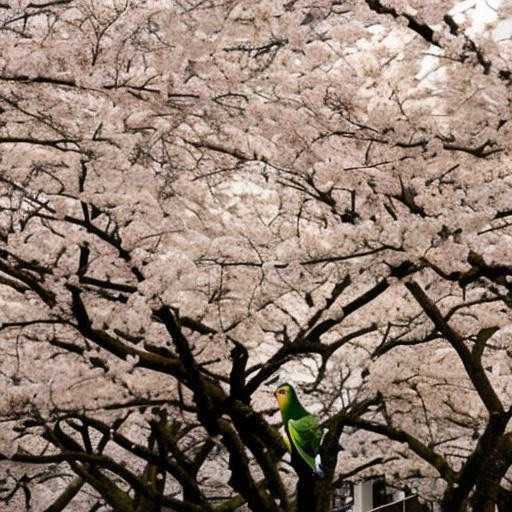}
  \end{minipage}
      \begin{minipage}[b]{0.105\textwidth}
    \includegraphics[width=\textwidth]{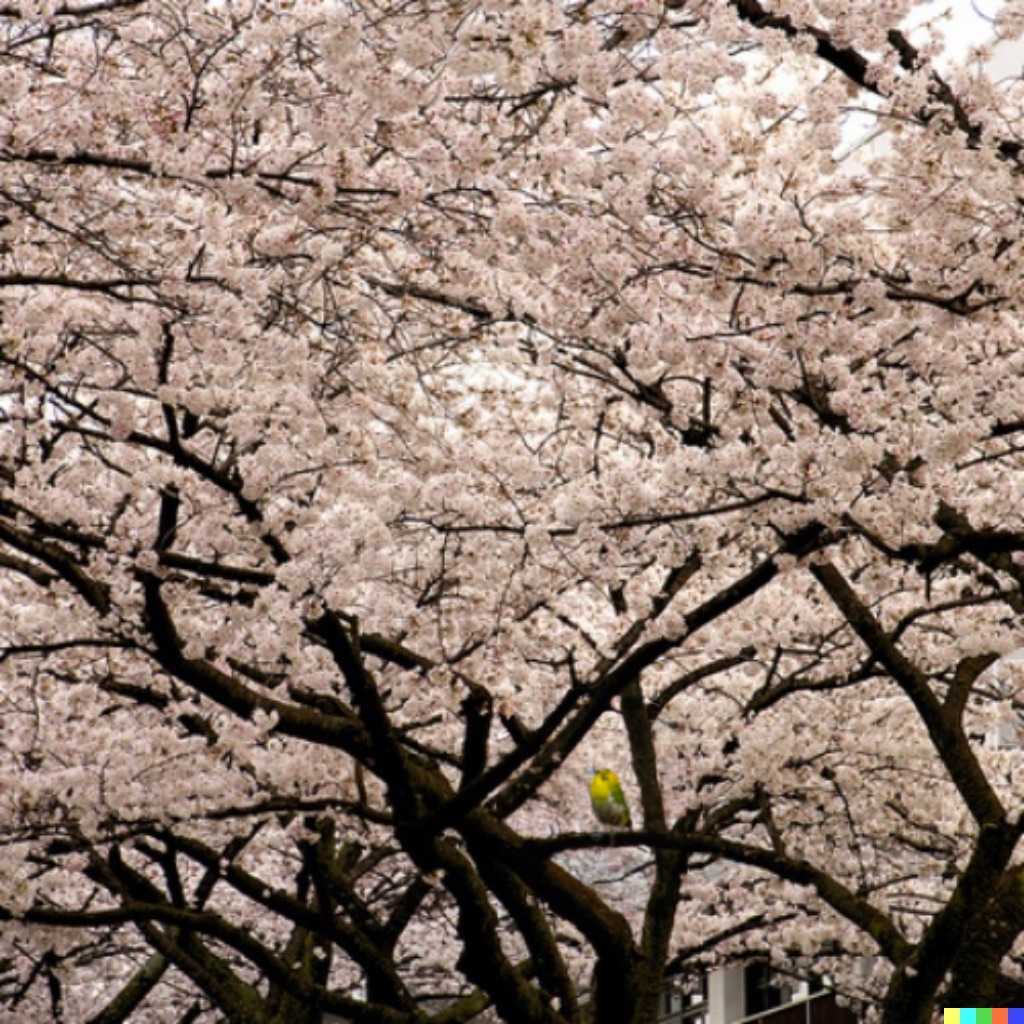}
  \end{minipage}

        \begin{minipage}[b]{0.99\textwidth}
      \centering
    \textit{``monkey''}
  \end{minipage}
      \begin{minipage}[b]{0.105\textwidth}
    \includegraphics[width=\textwidth]{figures/15272-input.jpg}
  \end{minipage} 
          \begin{minipage}[b]{0.105\textwidth}
    \includegraphics[width=\textwidth]{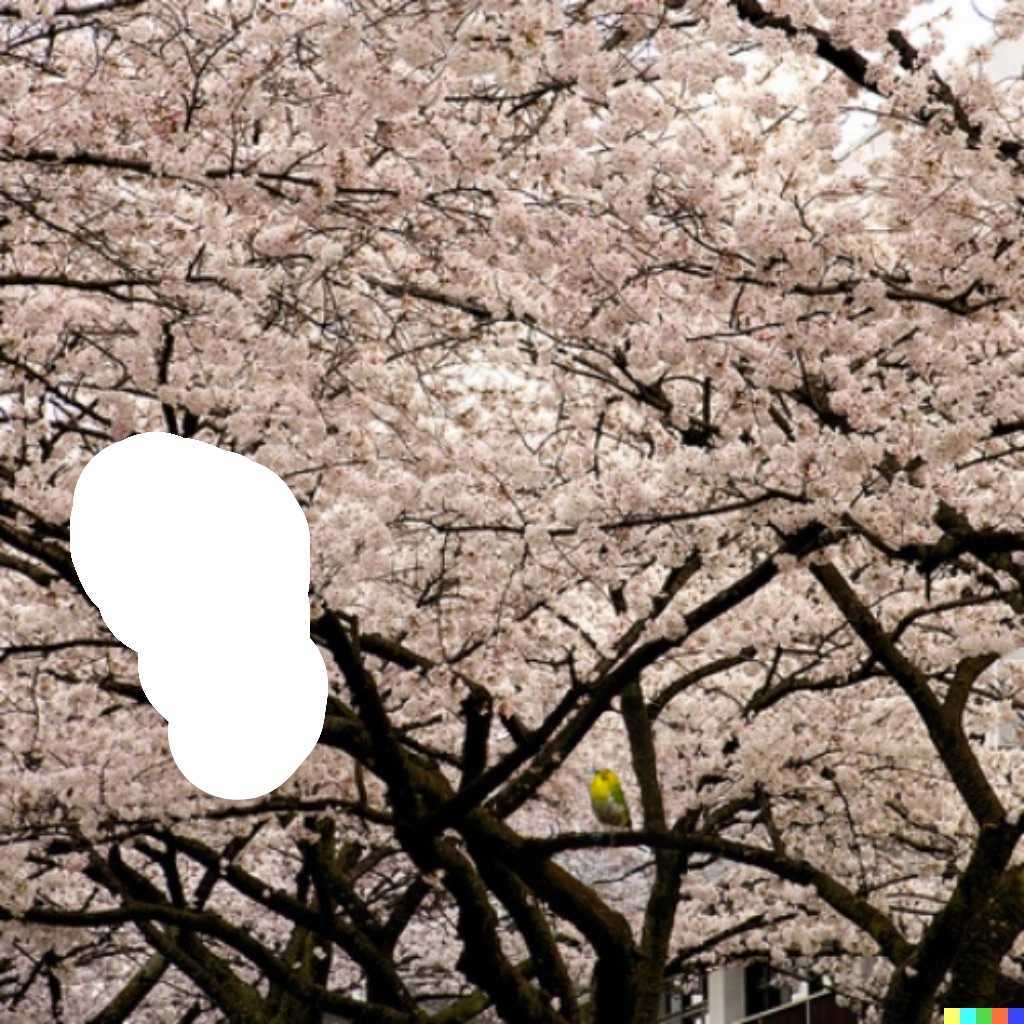}
  \end{minipage}\rulesep
    \begin{minipage}[b]{0.105\textwidth}
    \includegraphics[width=\textwidth]{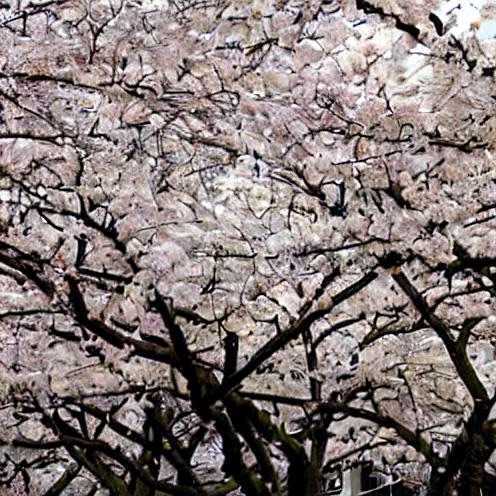}
  \end{minipage}
    \begin{minipage}[b]{0.105\textwidth}
    \includegraphics[width=\textwidth]{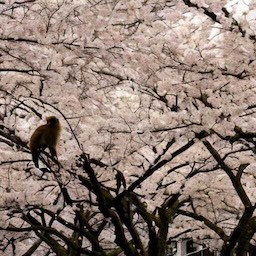}
  \end{minipage}
        \begin{minipage}[b]{0.105\textwidth}
    \includegraphics[width=\textwidth]{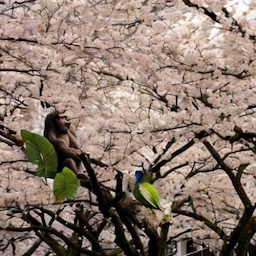}
  \end{minipage}
        \begin{minipage}[b]{0.105\textwidth}
    \includegraphics[width=\textwidth]{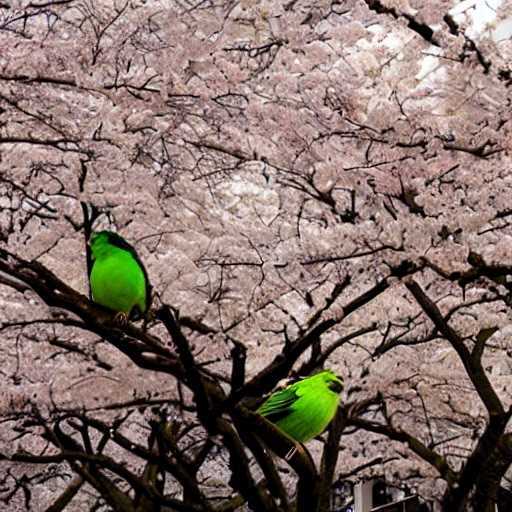}
  \end{minipage}
      \begin{minipage}[b]{0.105\textwidth}
    \includegraphics[width=\textwidth]{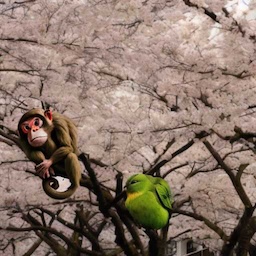}
  \end{minipage}
      \begin{minipage}[b]{0.105\textwidth}
    \includegraphics[width=\textwidth]{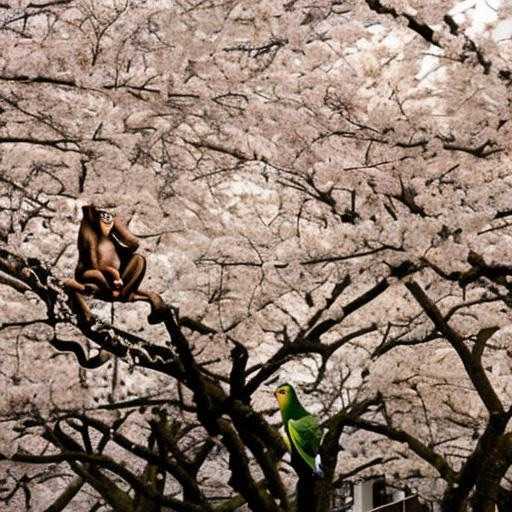}
  \end{minipage}
      \begin{minipage}[b]{0.105\textwidth}
    \includegraphics[width=\textwidth]{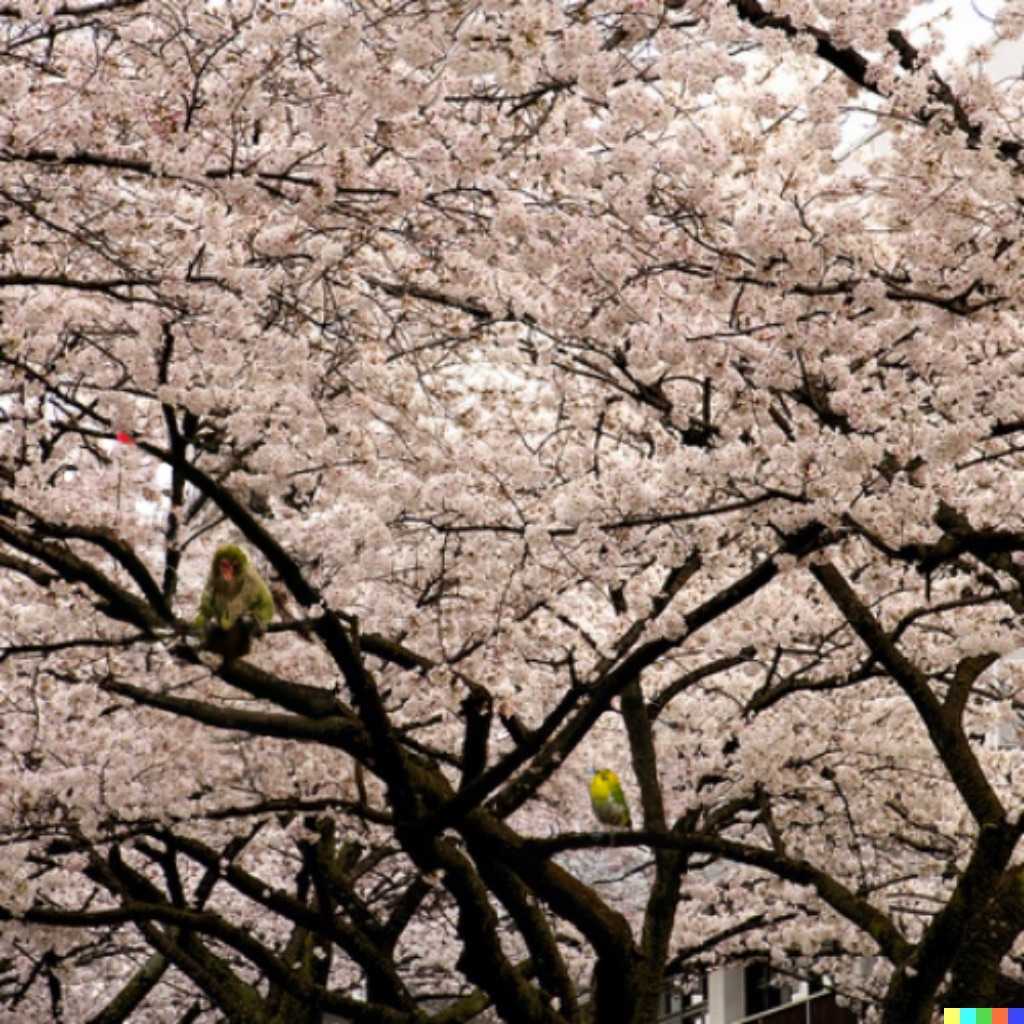}
  \end{minipage}

        \begin{minipage}[b]{0.99\textwidth}
      \centering
    \textit{``happy face emoticon''}
  \end{minipage}

      \begin{minipage}[b]{0.105\textwidth}
    \includegraphics[width=\textwidth]{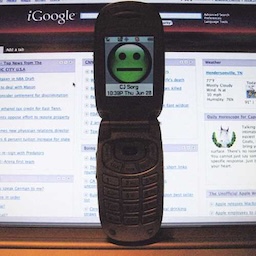}
  \end{minipage} 
          \begin{minipage}[b]{0.105\textwidth}
    \includegraphics[width=\textwidth]{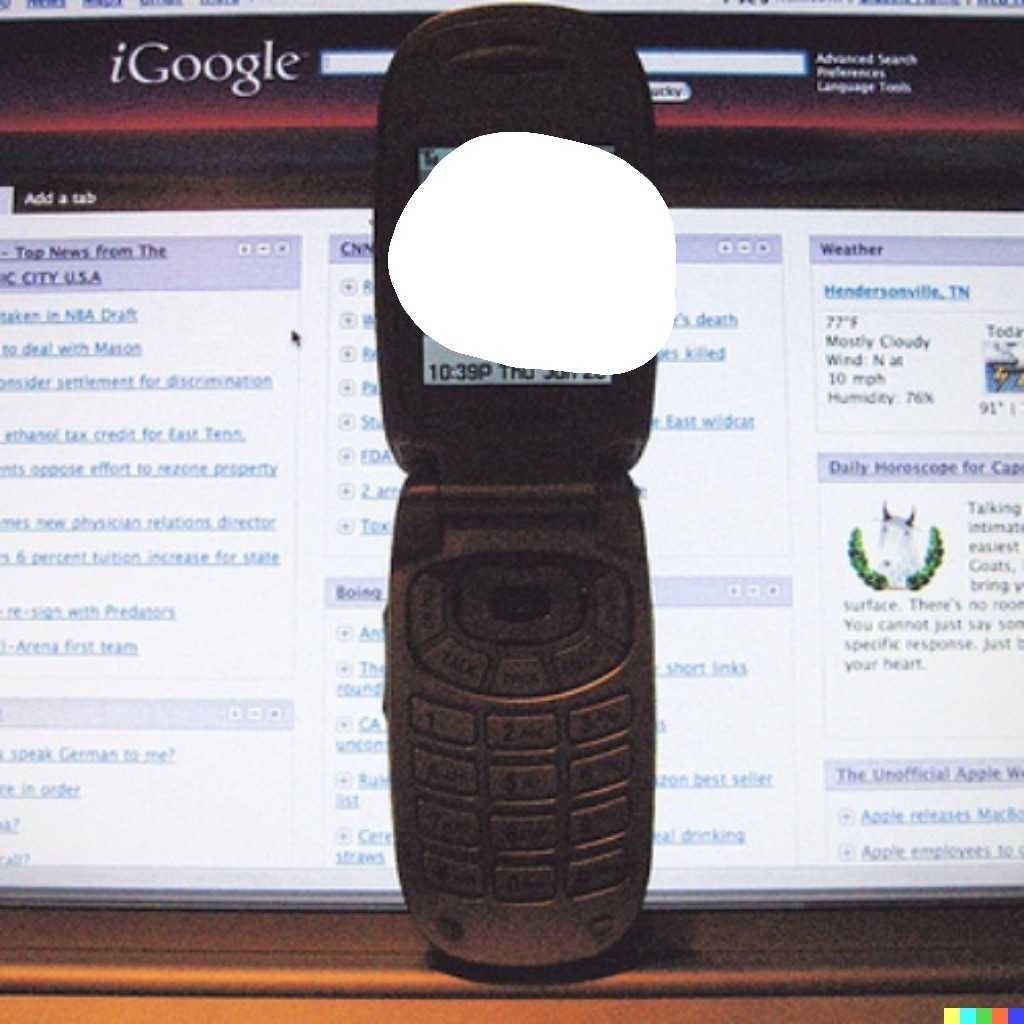}
  \end{minipage}\rulesep
    \begin{minipage}[b]{0.105\textwidth}
    \includegraphics[width=\textwidth]{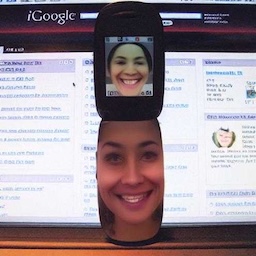}
  \end{minipage}
    \begin{minipage}[b]{0.105\textwidth}
    \includegraphics[width=\textwidth]{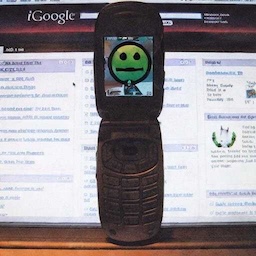}
  \end{minipage}
        \begin{minipage}[b]{0.105\textwidth}
    \includegraphics[width=\textwidth]{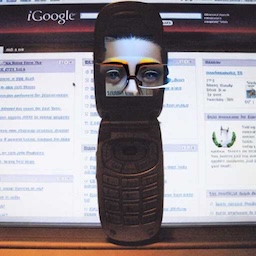}
  \end{minipage}
        \begin{minipage}[b]{0.105\textwidth}
    \includegraphics[width=\textwidth]{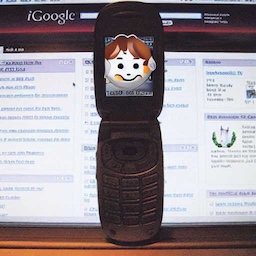}
  \end{minipage}
      \begin{minipage}[b]{0.105\textwidth}
    \includegraphics[width=\textwidth]{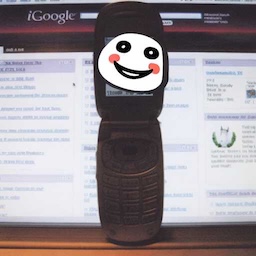}
  \end{minipage}
      \begin{minipage}[b]{0.105\textwidth}
    \includegraphics[width=\textwidth]{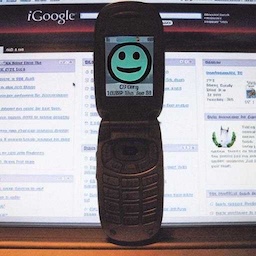}
  \end{minipage}
      \begin{minipage}[b]{0.105\textwidth}
    \includegraphics[width=\textwidth]{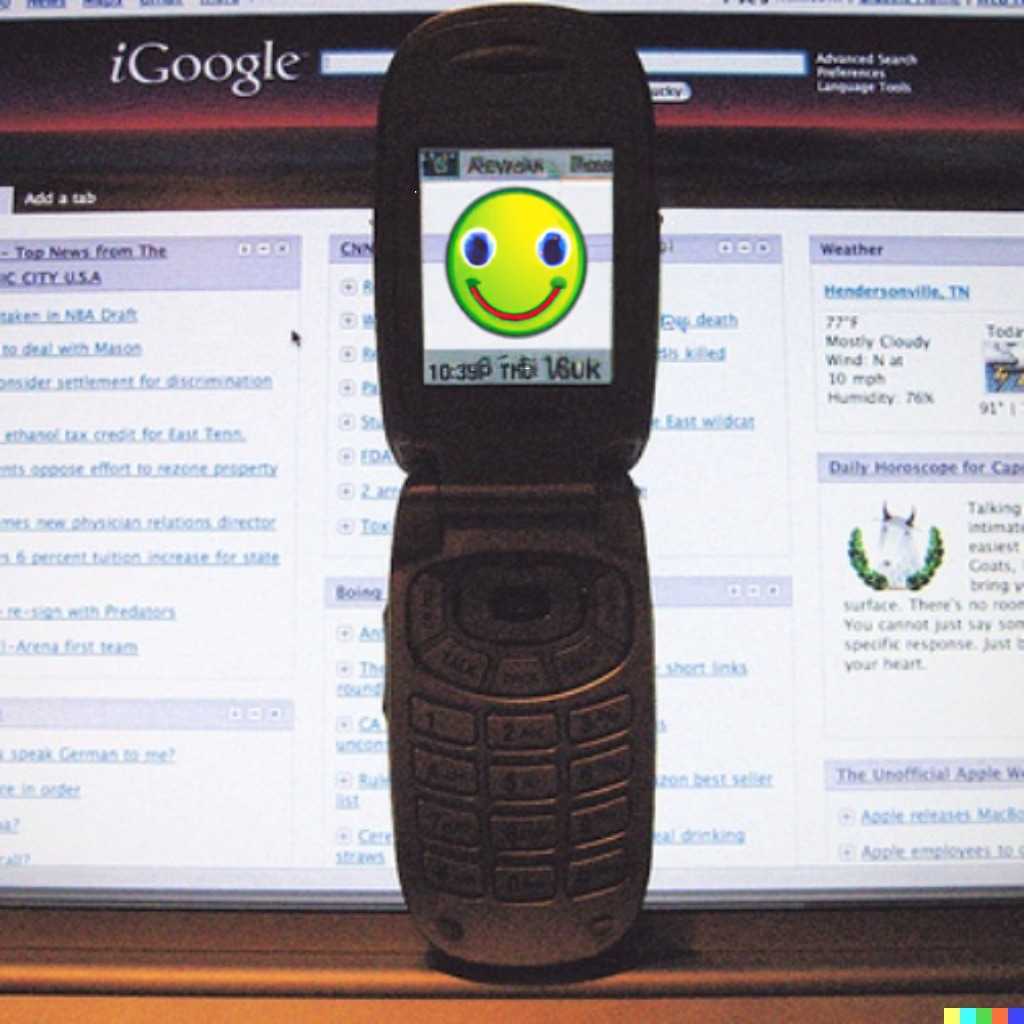}
  \end{minipage}

      \begin{minipage}[b]{0.99\textwidth}
      \centering
    \textit{``hat''}
  \end{minipage}

        \begin{minipage}[b]{0.105\textwidth}
    \includegraphics[width=\textwidth]{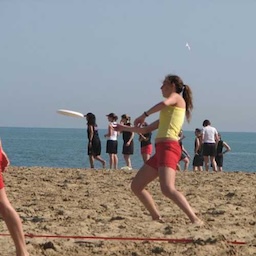}
        \caption*{\small Input image}
  \end{minipage}
    \begin{minipage}[b]{0.105\textwidth}
    \includegraphics[width=\textwidth]{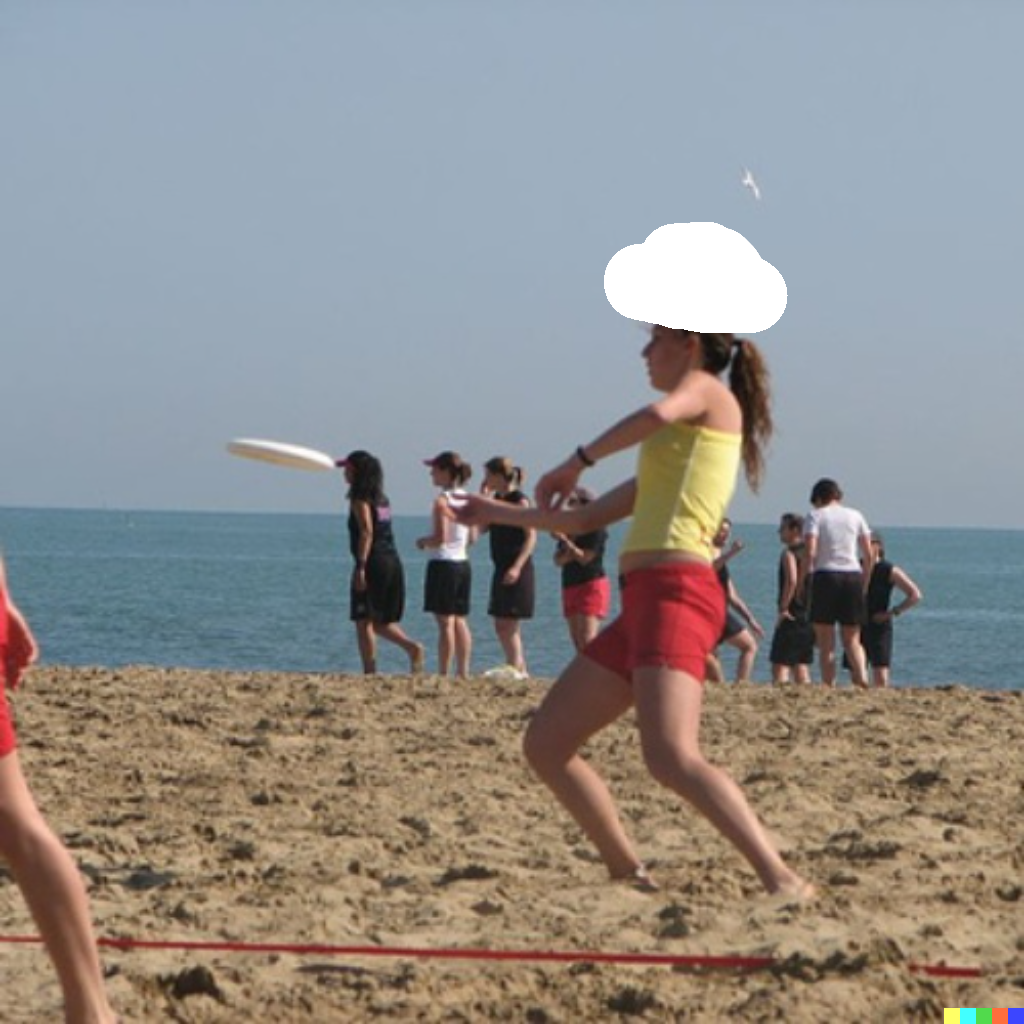}
        \caption*{editing mask}
  \end{minipage}\rulesep
    \begin{minipage}[b]{0.105\textwidth}
    \includegraphics[width=\textwidth]{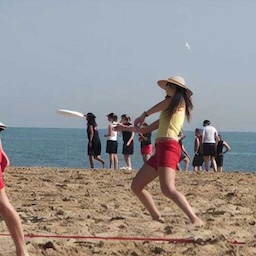}
        \caption*{IP2P}
  \end{minipage}
      \begin{minipage}[b]{0.105\textwidth}
    \includegraphics[width=\textwidth]{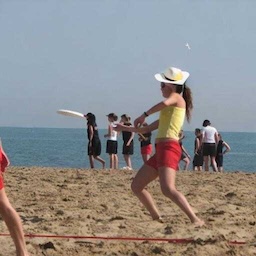}
        \caption*{SDI}
  \end{minipage}
        \begin{minipage}[b]{0.105\textwidth}
      \centering
        \includegraphics[width=\textwidth]{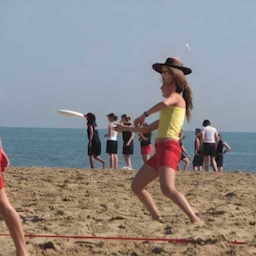}
        \caption*{BLD}
  \end{minipage}
        \begin{minipage}[b]{0.105\textwidth}
      \centering
        \includegraphics[width=\textwidth]{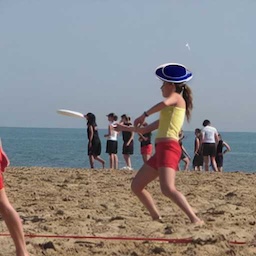}
        \caption*{HDP}
  \end{minipage}
          \begin{minipage}[b]{0.105\textwidth}
      \centering
        \includegraphics[width=\textwidth]{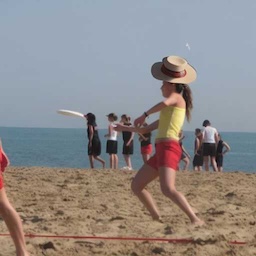}
        \caption*{BrushNet}
  \end{minipage}
      \begin{minipage}[b]{0.105\textwidth}
      \centering
        \includegraphics[width=\textwidth]{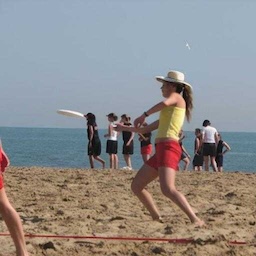}
        \caption*{\textbf{LDB (Ours)}}
  \end{minipage}
        \begin{minipage}[b]{0.105\textwidth}
    \includegraphics[width=\textwidth]{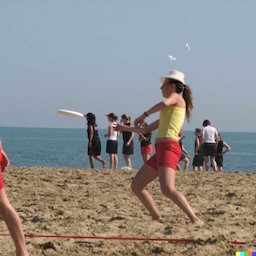}

        \caption*{GT}
  \end{minipage}

\caption{Additional qualitative examples on the MagicBrush dataset. Note that the images are not cherry picked and correspond to the user study (IP2P, SDI, LDB) and quantitative evaluation (BLD, HDP, BrushNet) with default settings.}
\label{fig:additionalmagic}
\end{figure*}

\begin{figure*}[!ht]
  \centering

        \begin{minipage}[b]{0.99\textwidth}
      \centering
    \textit{``looking at the camera''}
  \end{minipage}

      \begin{minipage}[b]{0.12\textwidth}
    \includegraphics[width=\textwidth]{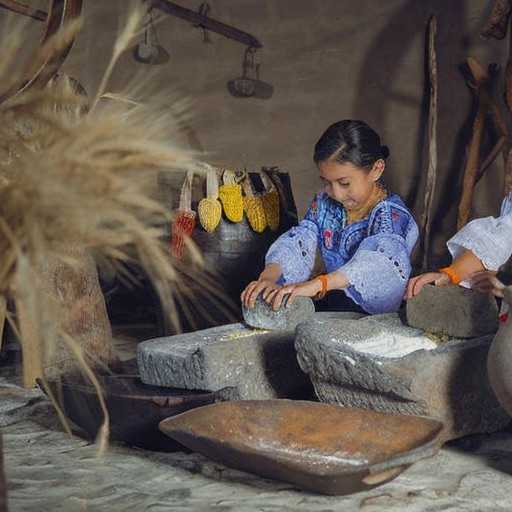}
  \end{minipage} 
          \begin{minipage}[b]{0.12\textwidth}
    \includegraphics[width=\textwidth]{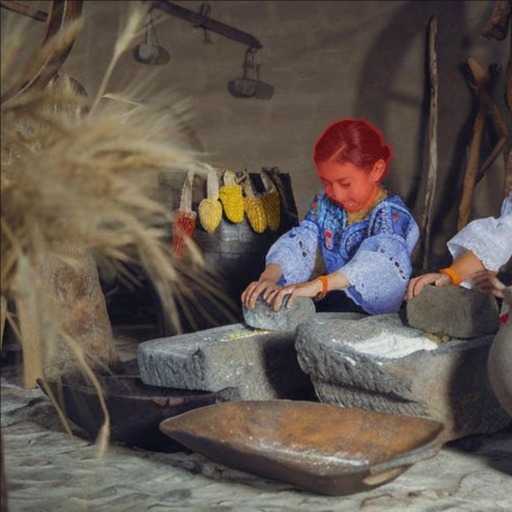}
  \end{minipage}\rulesep
    \begin{minipage}[b]{0.12\textwidth}
    \includegraphics[width=\textwidth]{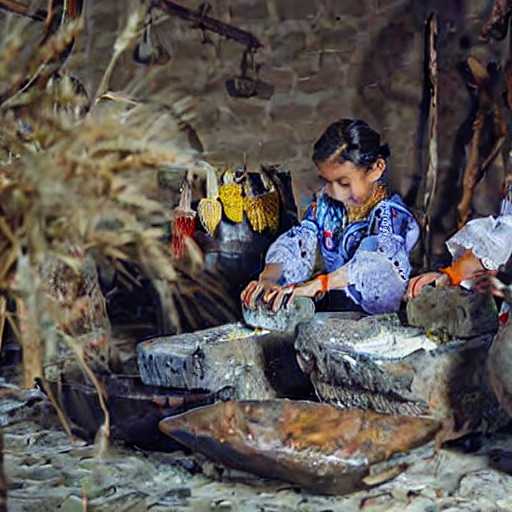}
  \end{minipage}
    \begin{minipage}[b]{0.12\textwidth}
    \includegraphics[width=\textwidth]{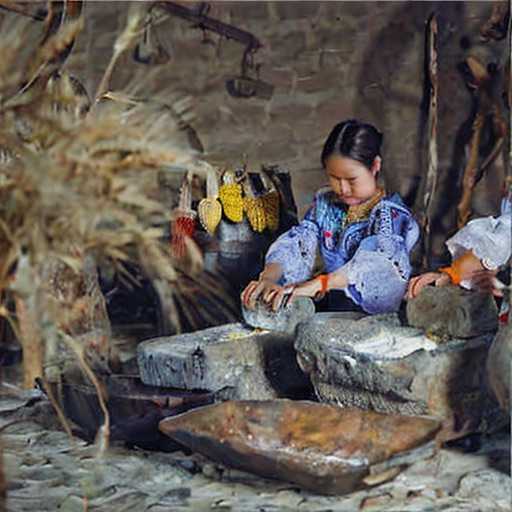}
  \end{minipage}
        \begin{minipage}[b]{0.12\textwidth}
    \includegraphics[width=\textwidth]{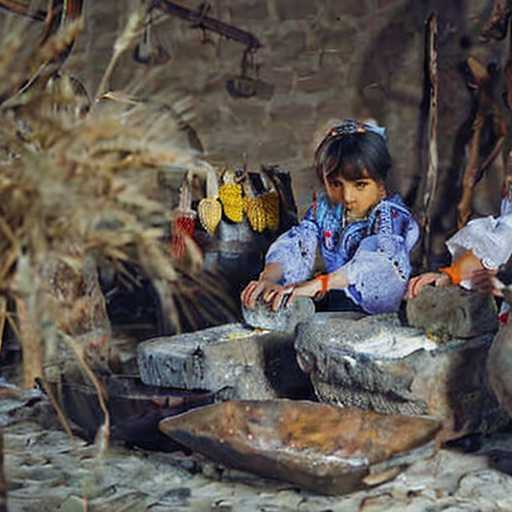}
  \end{minipage}
        \begin{minipage}[b]{0.12\textwidth}
    \includegraphics[width=\textwidth]{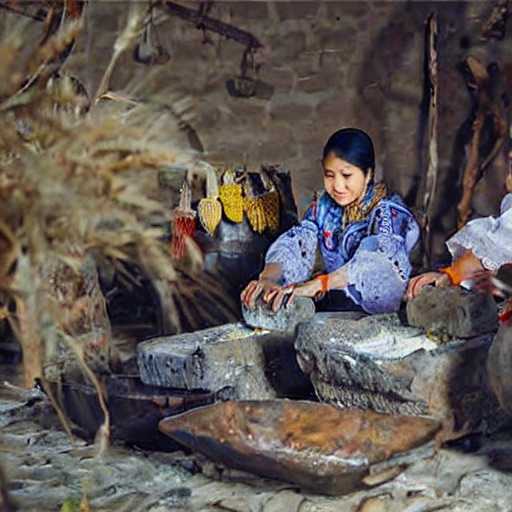}
  \end{minipage}
      \begin{minipage}[b]{0.12\textwidth}
    \includegraphics[width=\textwidth]{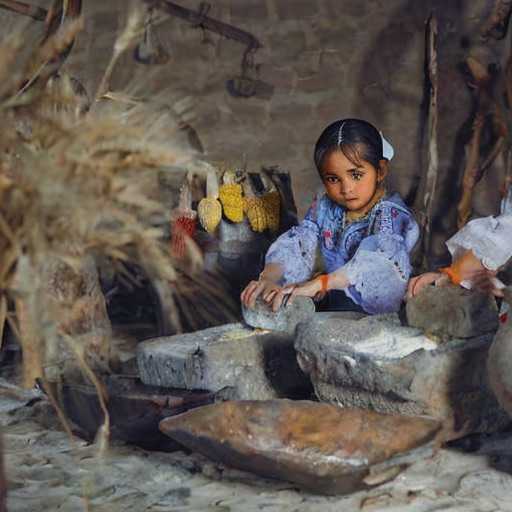}
  \end{minipage}
      \begin{minipage}[b]{0.12\textwidth}
    \includegraphics[width=\textwidth]{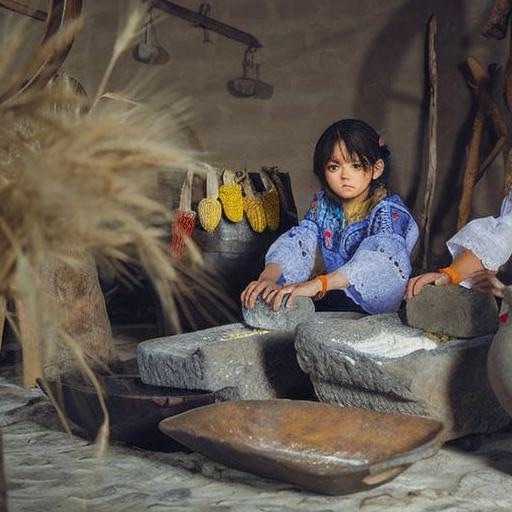}
  \end{minipage}

        \begin{minipage}[b]{0.99\textwidth}
      \centering
    \textit{``a field (remove dandelions)''}
  \end{minipage}

      \begin{minipage}[b]{0.12\textwidth}
    \includegraphics[width=\textwidth]{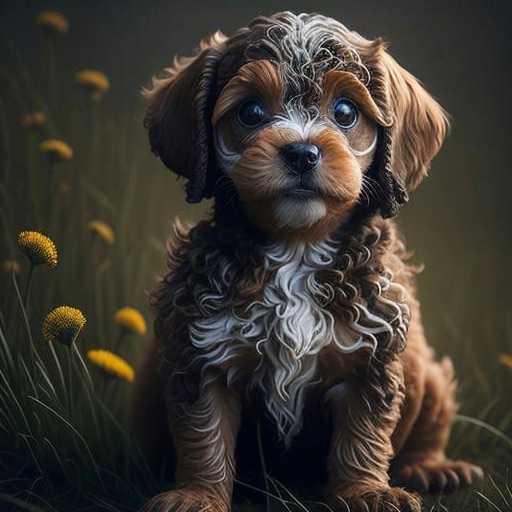}
  \end{minipage} 
          \begin{minipage}[b]{0.12\textwidth}
    \includegraphics[width=\textwidth]{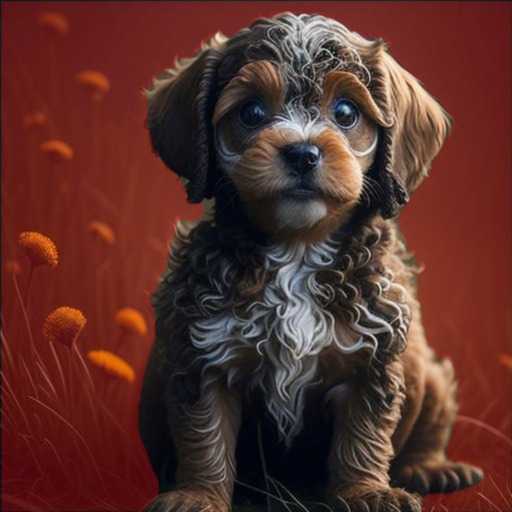}
  \end{minipage}\rulesep
    \begin{minipage}[b]{0.12\textwidth}
    \includegraphics[width=\textwidth]{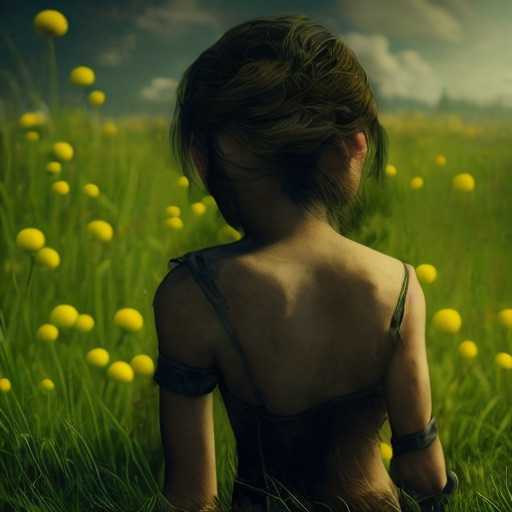}
  \end{minipage}
    \begin{minipage}[b]{0.12\textwidth}
    \includegraphics[width=\textwidth]{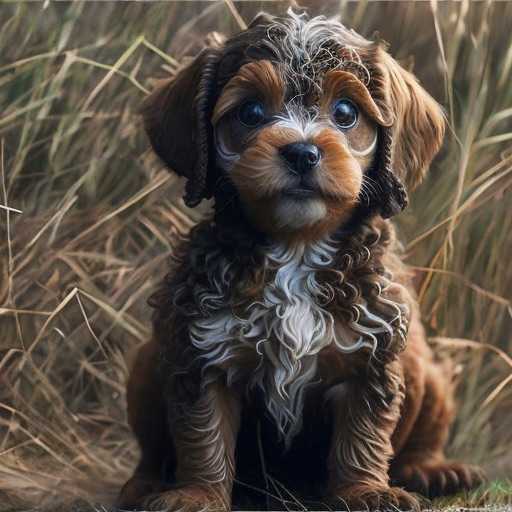}
  \end{minipage}
        \begin{minipage}[b]{0.12\textwidth}
    \includegraphics[width=\textwidth]{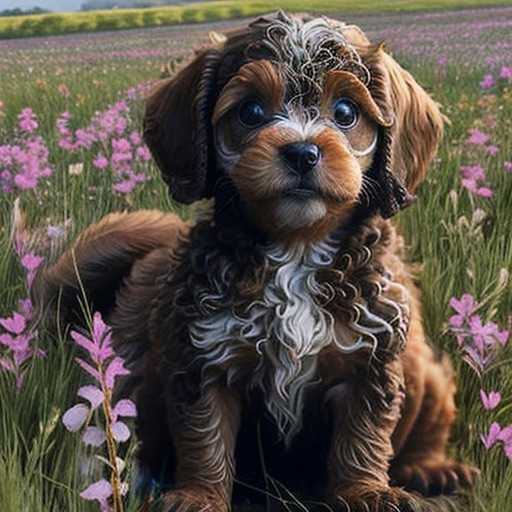}
  \end{minipage}
        \begin{minipage}[b]{0.12\textwidth}
    \includegraphics[width=\textwidth]{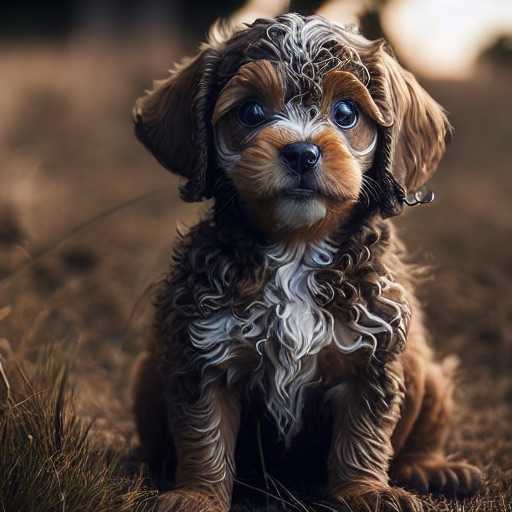}
  \end{minipage}
      \begin{minipage}[b]{0.12\textwidth}
    \includegraphics[width=\textwidth]{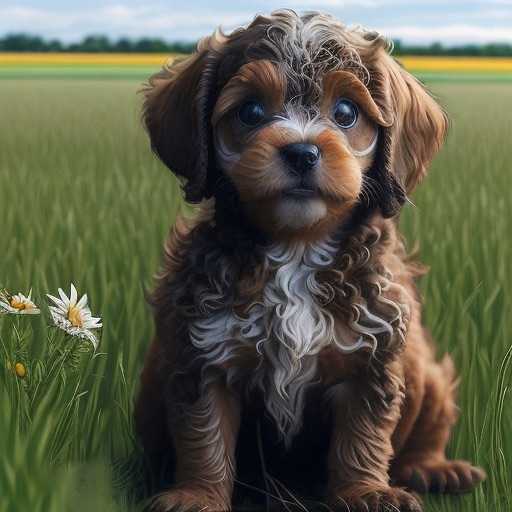}
  \end{minipage}
      \begin{minipage}[b]{0.12\textwidth}
    \includegraphics[width=\textwidth]{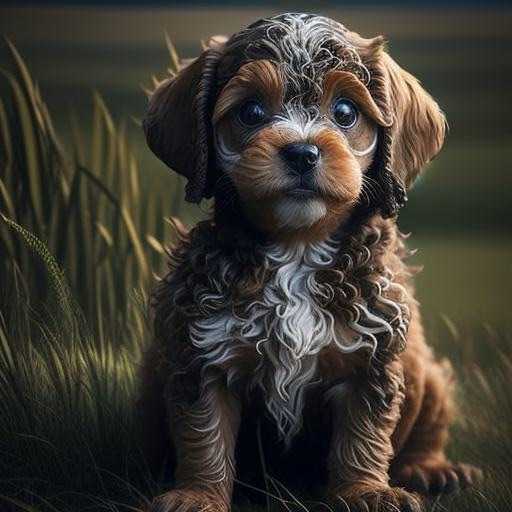}
  \end{minipage}

        \begin{minipage}[b]{0.99\textwidth}
      \centering
    \textit{``a foggy day''}
  \end{minipage}
      \begin{minipage}[b]{0.12\textwidth}
    \includegraphics[width=\textwidth]{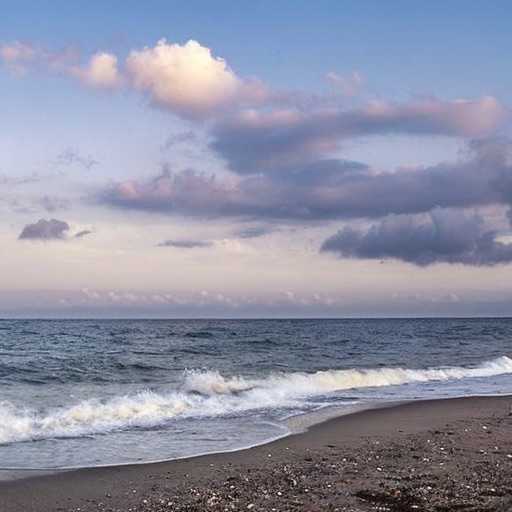}
  \end{minipage} 
          \begin{minipage}[b]{0.12\textwidth}
    \includegraphics[width=\textwidth]{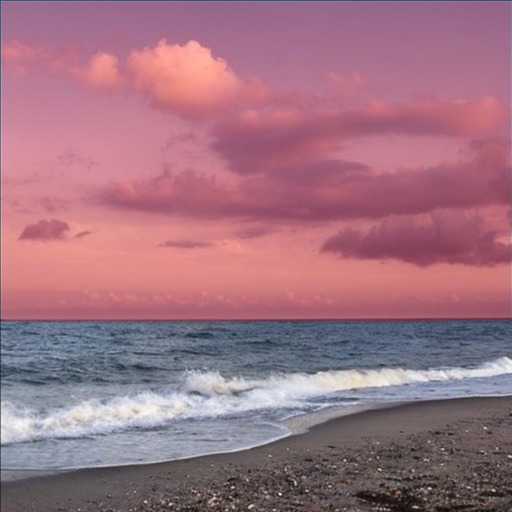}
  \end{minipage}\rulesep
    \begin{minipage}[b]{0.12\textwidth}
    \includegraphics[width=\textwidth]{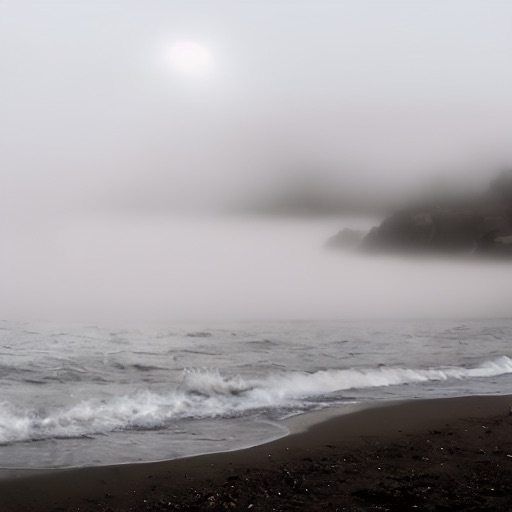}
  \end{minipage}
    \begin{minipage}[b]{0.12\textwidth}
    \includegraphics[width=\textwidth]{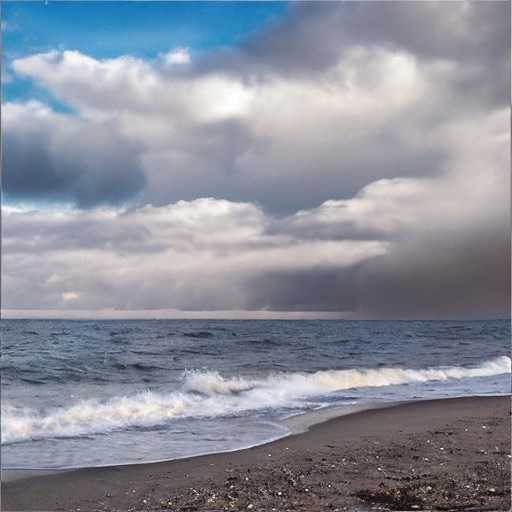}
  \end{minipage}
        \begin{minipage}[b]{0.12\textwidth}
    \includegraphics[width=\textwidth]{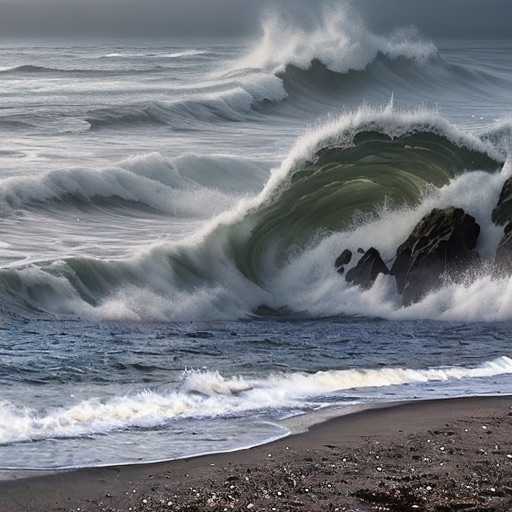}
  \end{minipage}
        \begin{minipage}[b]{0.12\textwidth}
    \includegraphics[width=\textwidth]{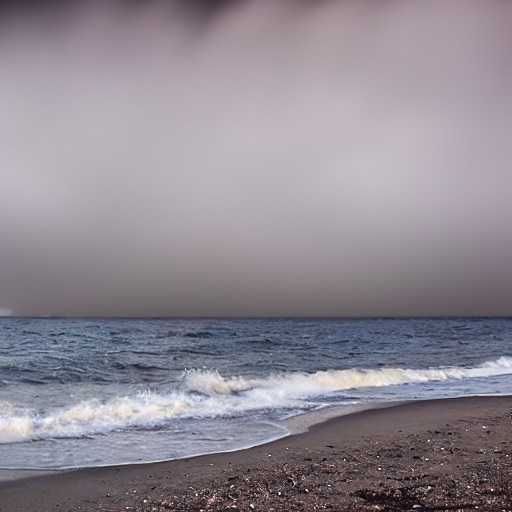}
  \end{minipage}
      \begin{minipage}[b]{0.12\textwidth}
    \includegraphics[width=\textwidth]{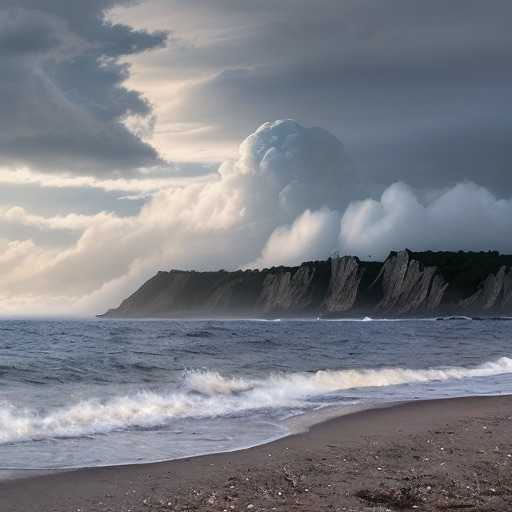}
  \end{minipage}
      \begin{minipage}[b]{0.12\textwidth}
    \includegraphics[width=\textwidth]{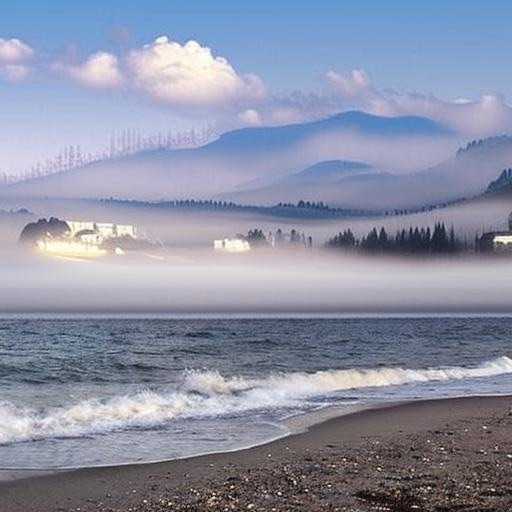}
  \end{minipage}

        \begin{minipage}[b]{0.99\textwidth}
      \centering
    \textit{``pig''}
  \end{minipage}
      \begin{minipage}[b]{0.12\textwidth}
    \includegraphics[width=\textwidth]{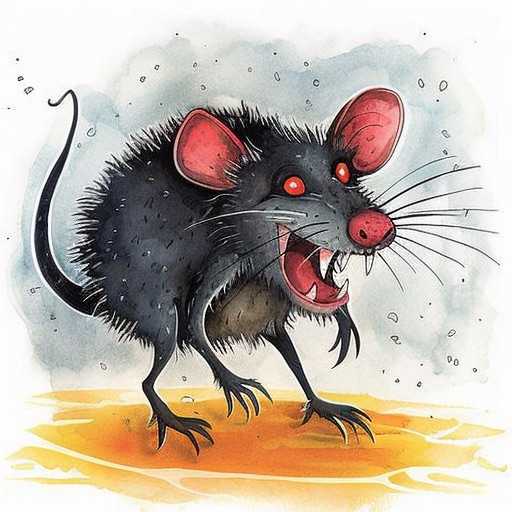}
  \end{minipage} 
          \begin{minipage}[b]{0.12\textwidth}
    \includegraphics[width=\textwidth]{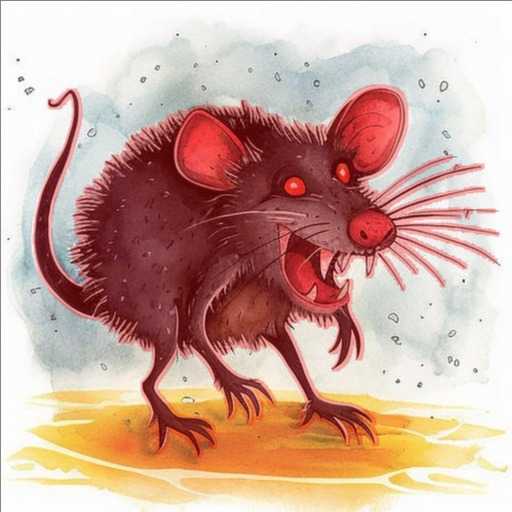}
  \end{minipage}\rulesep
    \begin{minipage}[b]{0.12\textwidth}
    \includegraphics[width=\textwidth]{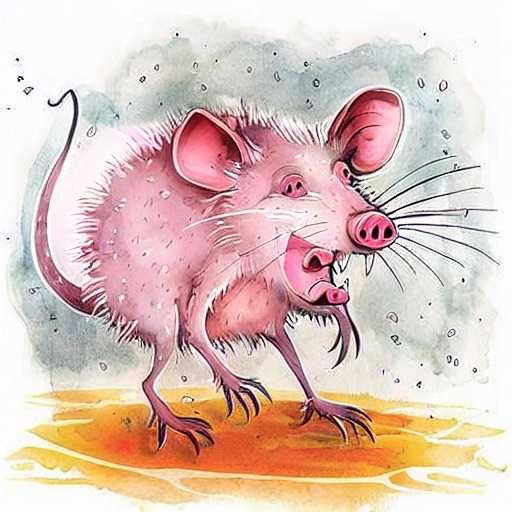}
  \end{minipage}
    \begin{minipage}[b]{0.12\textwidth}
    \includegraphics[width=\textwidth]{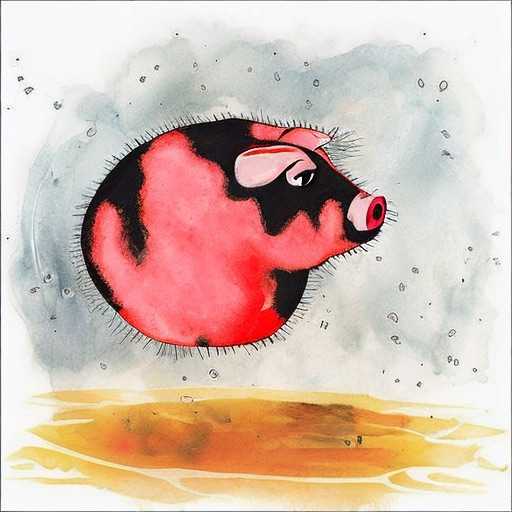}
  \end{minipage}
        \begin{minipage}[b]{0.12\textwidth}
    \includegraphics[width=\textwidth]{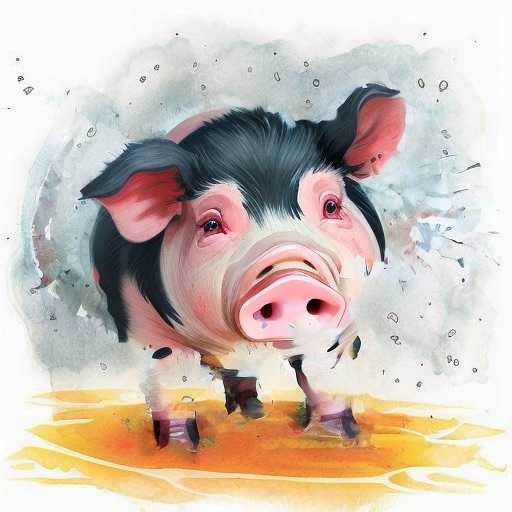}
  \end{minipage}
        \begin{minipage}[b]{0.12\textwidth}
    \includegraphics[width=\textwidth]{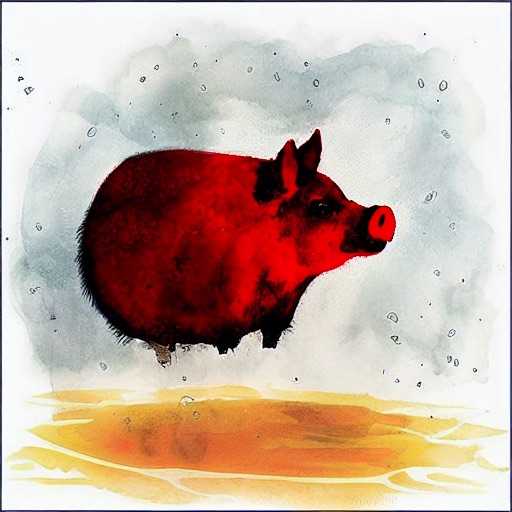}
  \end{minipage}
      \begin{minipage}[b]{0.12\textwidth}
    \includegraphics[width=\textwidth]{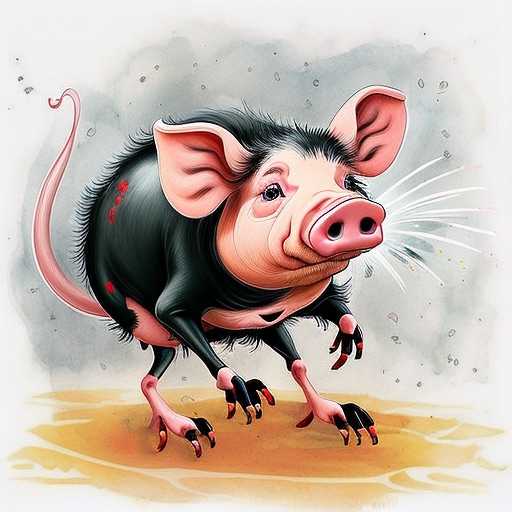}
  \end{minipage}
      \begin{minipage}[b]{0.12\textwidth}
    \includegraphics[width=\textwidth]{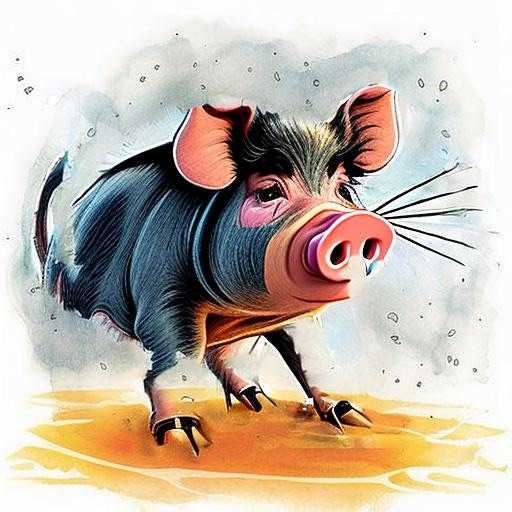}
  \end{minipage}

        \begin{minipage}[b]{0.99\textwidth}
      \centering
    \textit{``leopard''}
  \end{minipage}
      \begin{minipage}[b]{0.12\textwidth}
    \includegraphics[width=\textwidth]{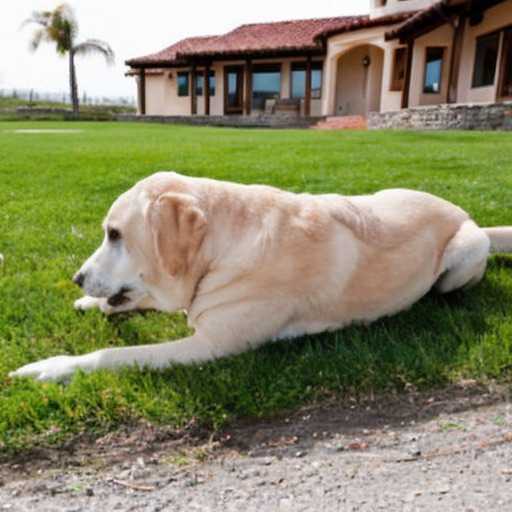}
  \end{minipage} 
          \begin{minipage}[b]{0.12\textwidth}
    \includegraphics[width=\textwidth]{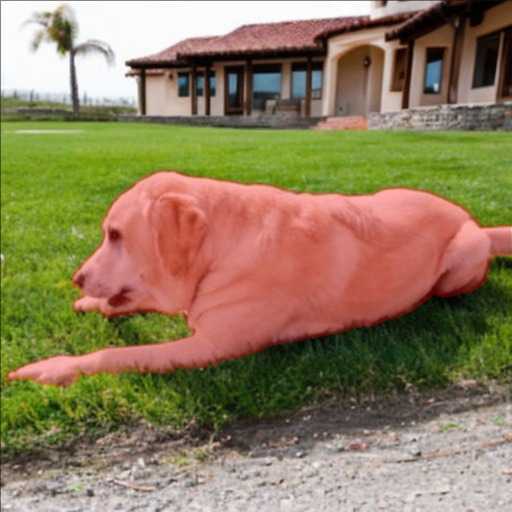}
  \end{minipage}\rulesep
    \begin{minipage}[b]{0.12\textwidth}
    \includegraphics[width=\textwidth]{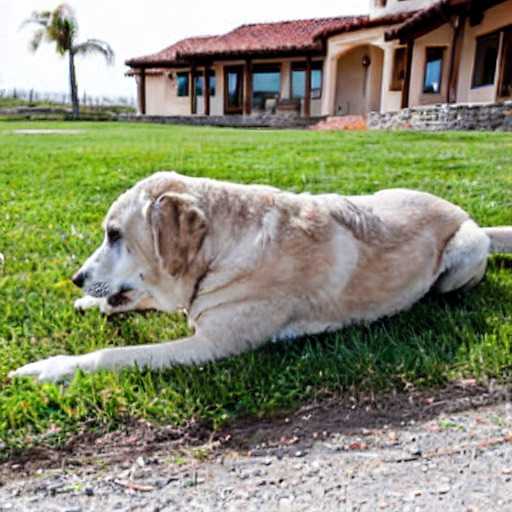}
  \end{minipage}
    \begin{minipage}[b]{0.12\textwidth}
    \includegraphics[width=\textwidth]{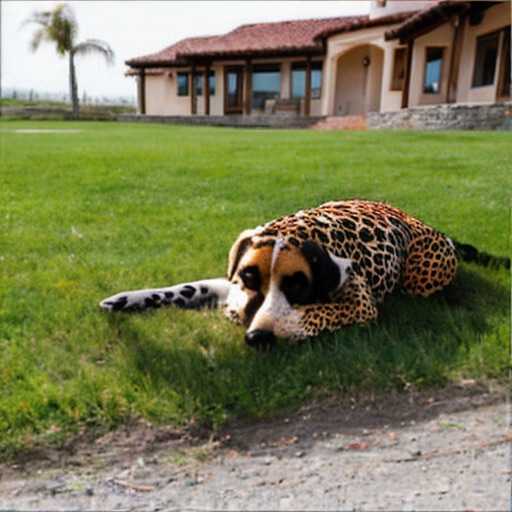}
  \end{minipage}
        \begin{minipage}[b]{0.12\textwidth}
    \includegraphics[width=\textwidth]{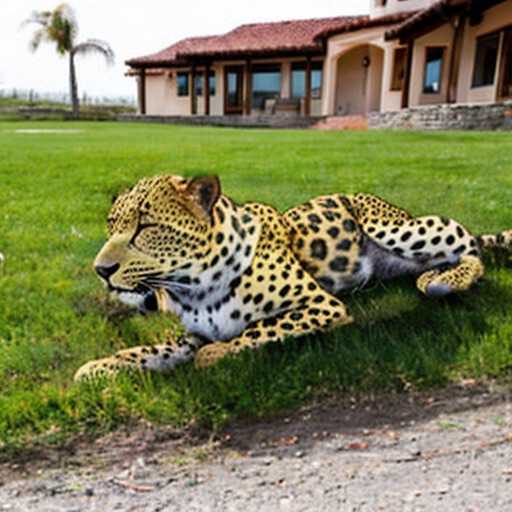}
  \end{minipage}
        \begin{minipage}[b]{0.12\textwidth}
    \includegraphics[width=\textwidth]{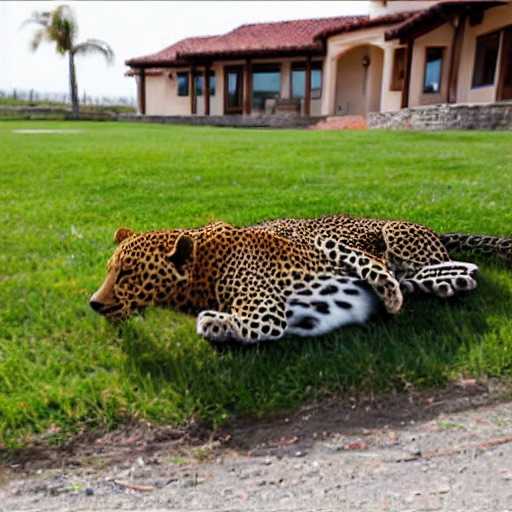}
  \end{minipage}
      \begin{minipage}[b]{0.12\textwidth}
    \includegraphics[width=\textwidth]{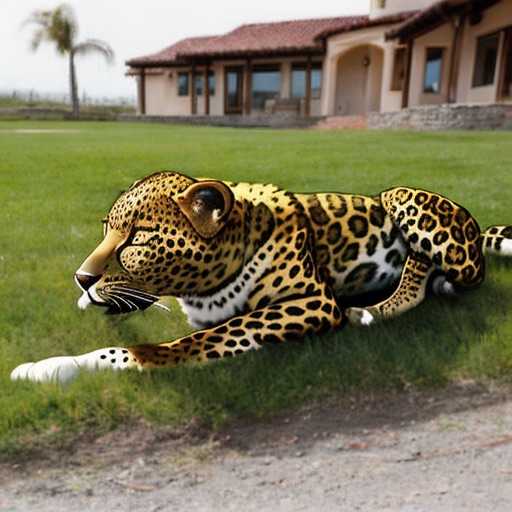}
  \end{minipage}
      \begin{minipage}[b]{0.12\textwidth}
    \includegraphics[width=\textwidth]{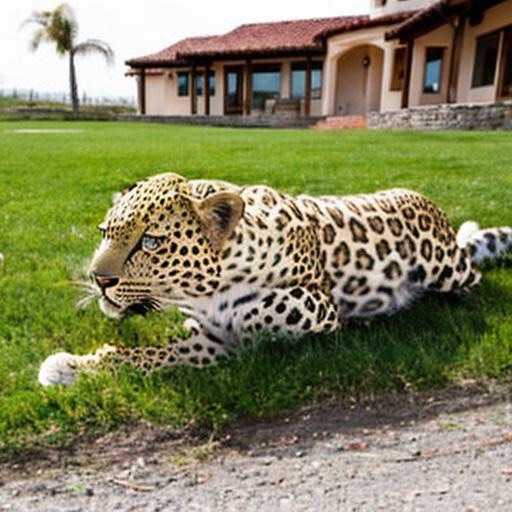}
  \end{minipage}

        \begin{minipage}[b]{0.99\textwidth}
      \centering
    \textit{``moon''}
  \end{minipage}
      \begin{minipage}[b]{0.12\textwidth}
    \includegraphics[width=\textwidth]{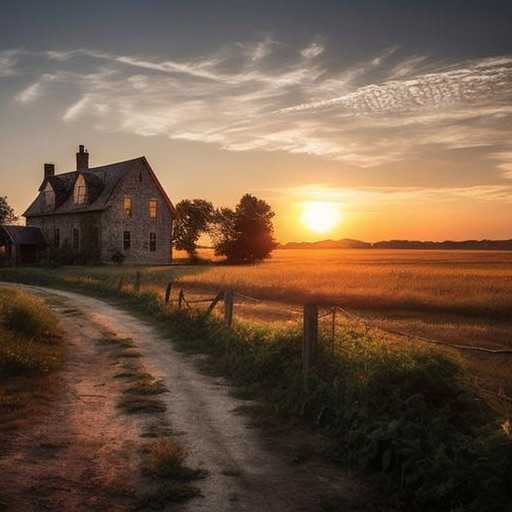}
  \end{minipage} 
          \begin{minipage}[b]{0.12\textwidth}
    \includegraphics[width=\textwidth]{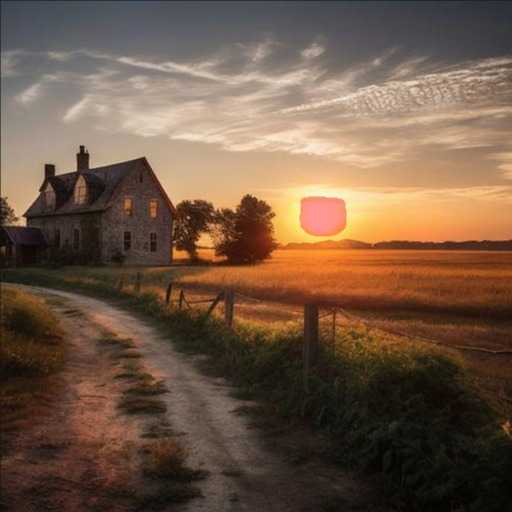}
  \end{minipage}\rulesep
    \begin{minipage}[b]{0.12\textwidth}
    \includegraphics[width=\textwidth]{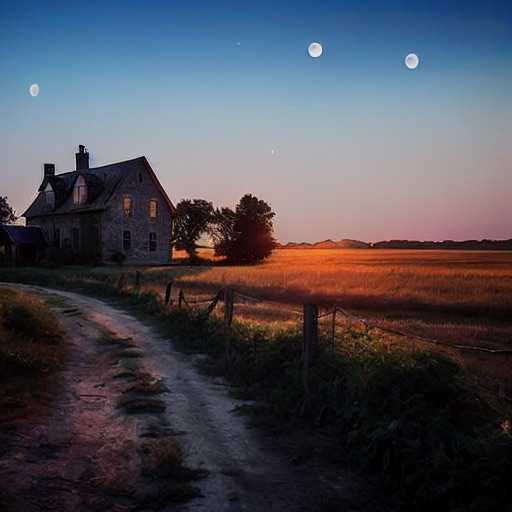}
  \end{minipage}
    \begin{minipage}[b]{0.12\textwidth}
    \includegraphics[width=\textwidth]{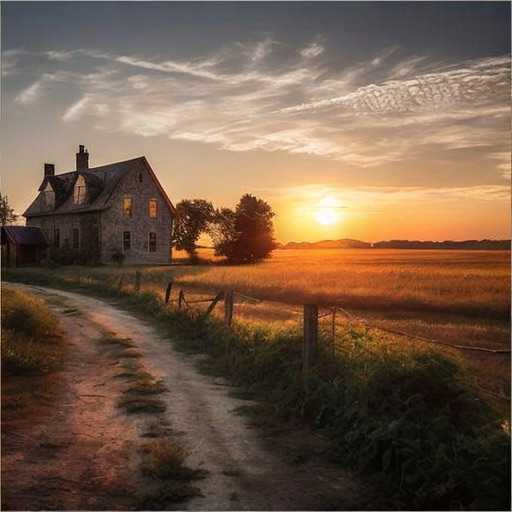}
  \end{minipage}
        \begin{minipage}[b]{0.12\textwidth}
    \includegraphics[width=\textwidth]{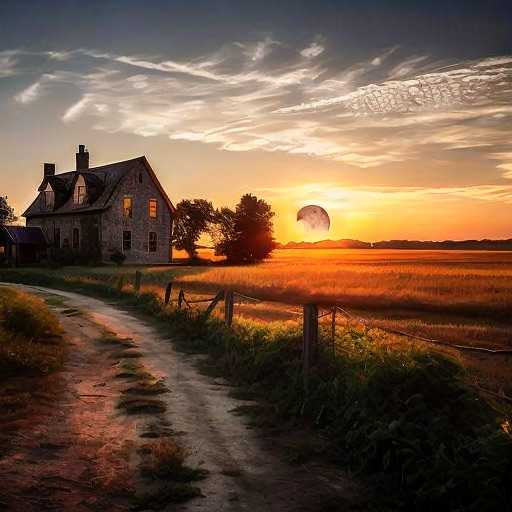}
  \end{minipage}
        \begin{minipage}[b]{0.12\textwidth}
    \includegraphics[width=\textwidth]{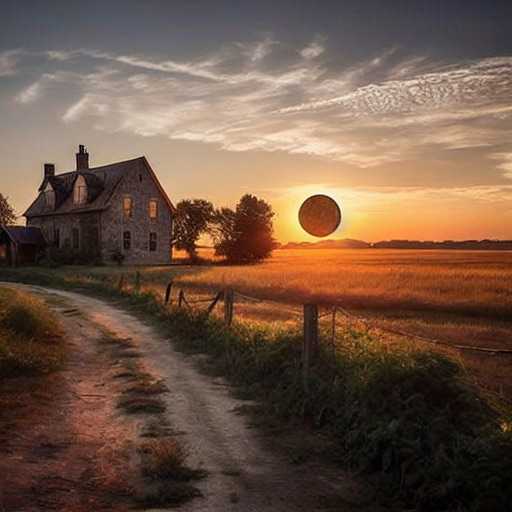}
  \end{minipage}
      \begin{minipage}[b]{0.12\textwidth}
    \includegraphics[width=\textwidth]{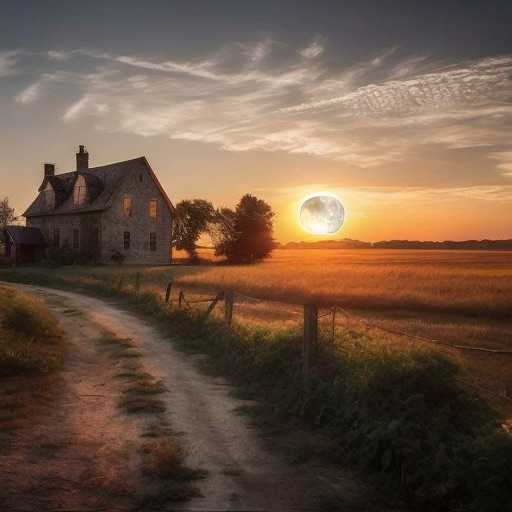}
  \end{minipage}
      \begin{minipage}[b]{0.12\textwidth}
    \includegraphics[width=\textwidth]{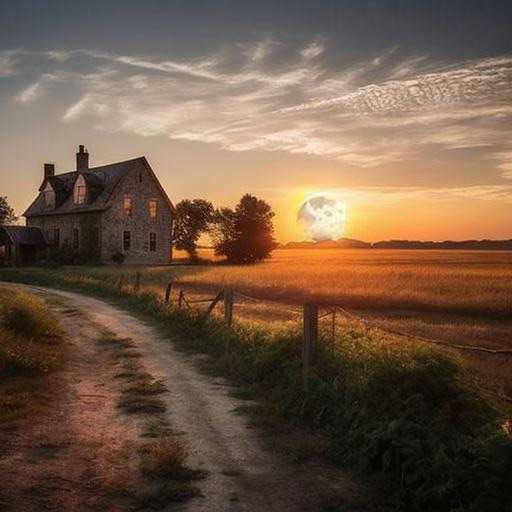}
  \end{minipage}

          \begin{minipage}[b]{0.99\textwidth}
      \centering
    \textit{``cartoon''}
  \end{minipage}
      \begin{minipage}[b]{0.12\textwidth}
    \includegraphics[width=\textwidth]{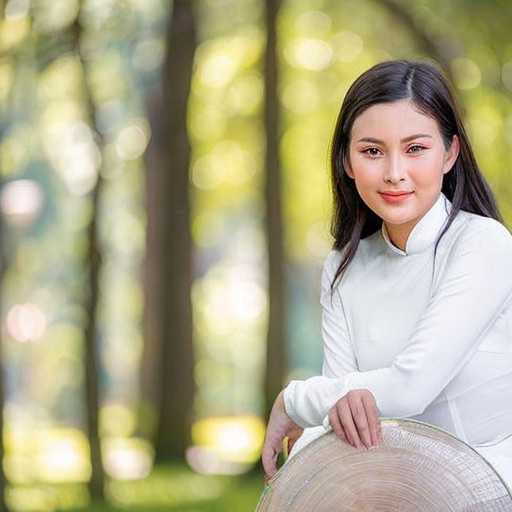}
  \end{minipage} 
          \begin{minipage}[b]{0.12\textwidth}
    \includegraphics[width=\textwidth]{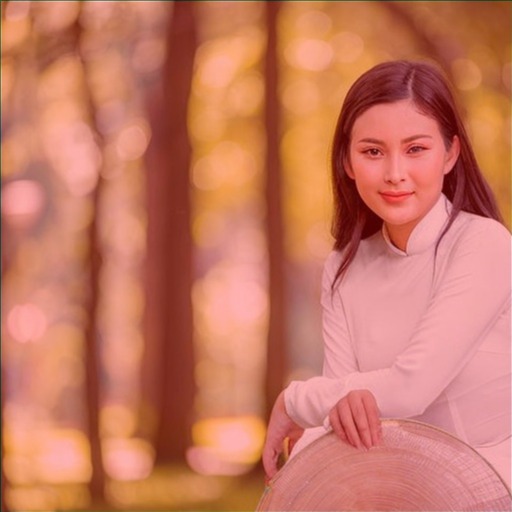}
  \end{minipage}\rulesep
    \begin{minipage}[b]{0.12\textwidth}
    \includegraphics[width=\textwidth]{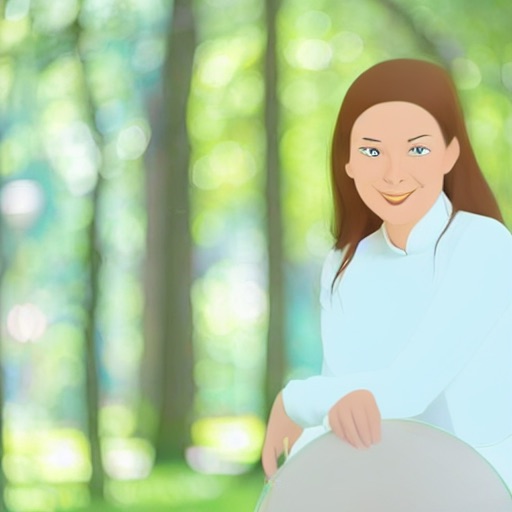}
  \end{minipage}
    \begin{minipage}[b]{0.12\textwidth}
    \includegraphics[width=\textwidth]{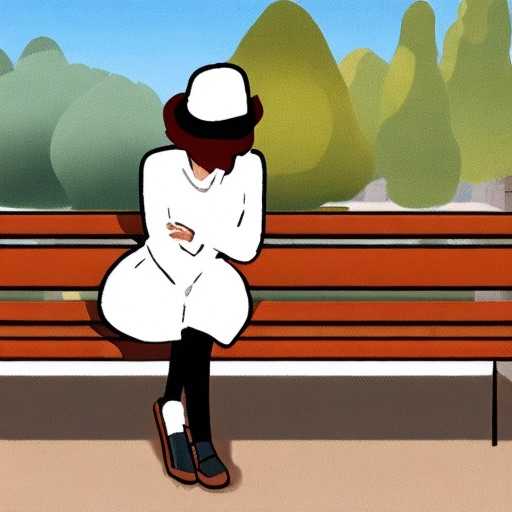}
  \end{minipage}
        \begin{minipage}[b]{0.12\textwidth}
    \includegraphics[width=\textwidth]{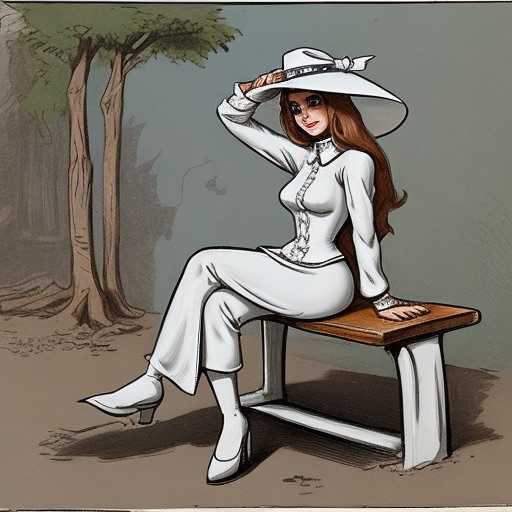}
  \end{minipage}
        \begin{minipage}[b]{0.12\textwidth}
    \includegraphics[width=\textwidth]{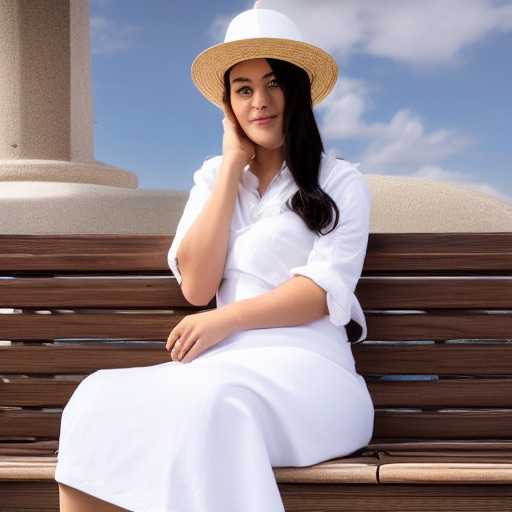}
  \end{minipage}
      \begin{minipage}[b]{0.12\textwidth}
    \includegraphics[width=\textwidth]{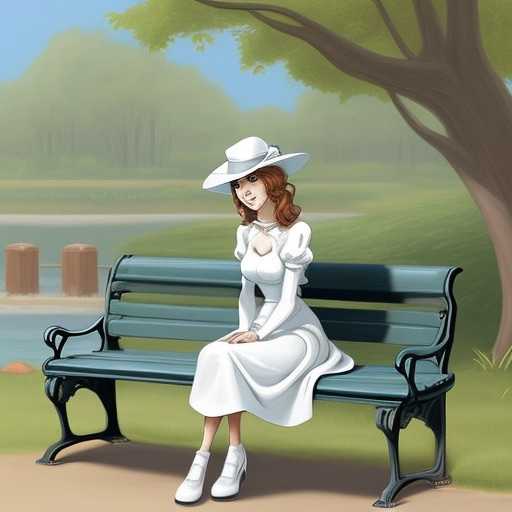}
  \end{minipage}
      \begin{minipage}[b]{0.12\textwidth}
    \includegraphics[width=\textwidth]{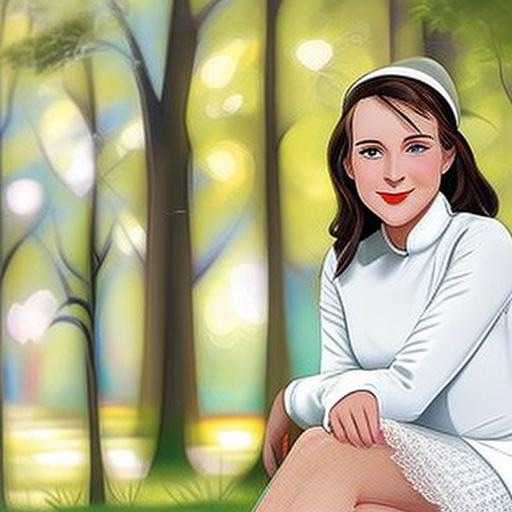}
  \end{minipage}
  
      \begin{minipage}[b]{0.99\textwidth}
      \centering
    \textit{``purse''}
  \end{minipage}

        \begin{minipage}[b]{0.12\textwidth}
    \includegraphics[width=\textwidth]{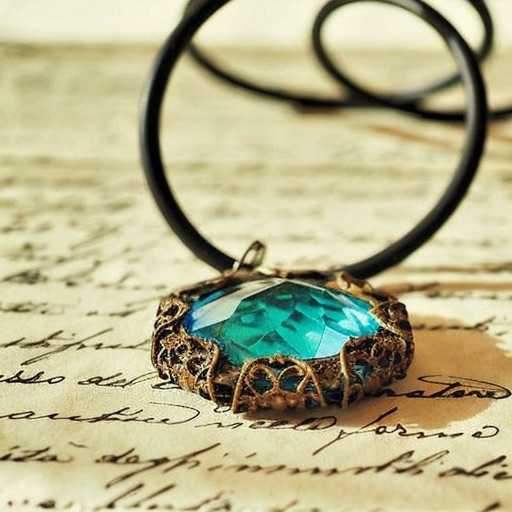}
        \caption*{\small Input image}
  \end{minipage}
    \begin{minipage}[b]{0.12\textwidth}
    \includegraphics[width=\textwidth]{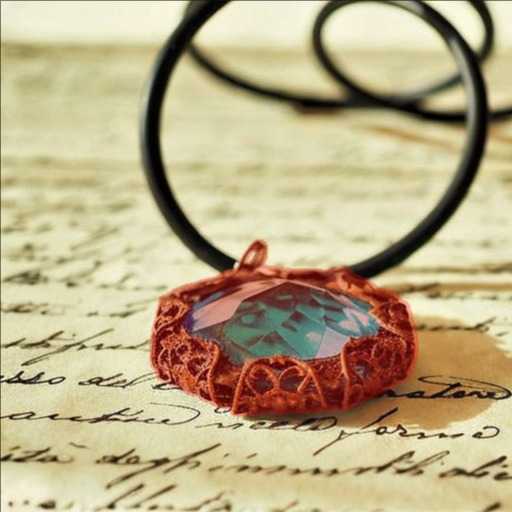}
        \caption*{editing mask}
  \end{minipage}\rulesep
    \begin{minipage}[b]{0.12\textwidth}
    \includegraphics[width=\textwidth]{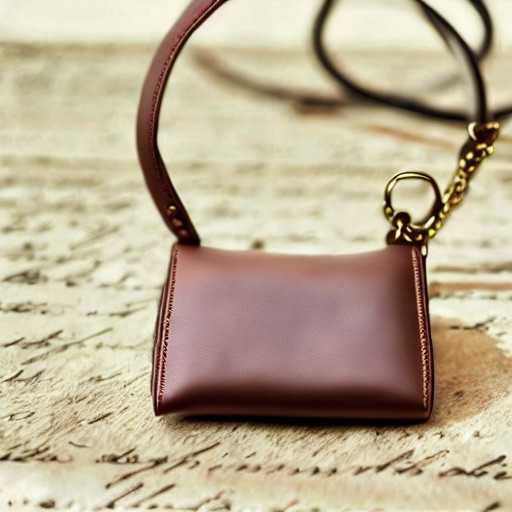}
        \caption*{IP2P}
  \end{minipage}
      \begin{minipage}[b]{0.12\textwidth}
    \includegraphics[width=\textwidth]{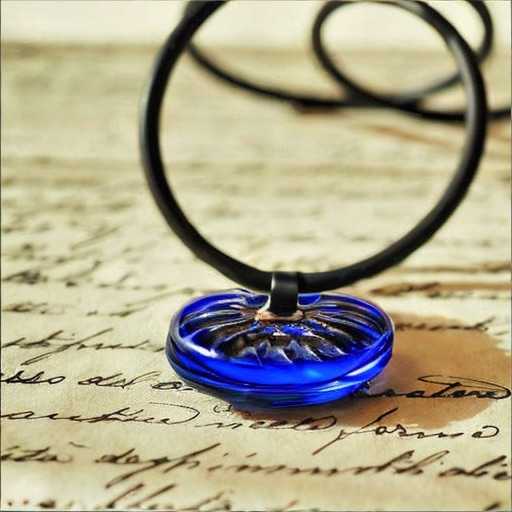}
        \caption*{SDI}
  \end{minipage}
        \begin{minipage}[b]{0.12\textwidth}
      \centering
        \includegraphics[width=\textwidth]{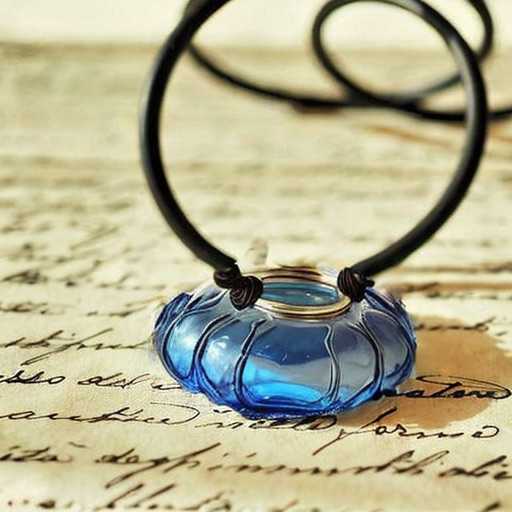}
        \caption*{BLD}
  \end{minipage}
        \begin{minipage}[b]{0.12\textwidth}
      \centering
        \includegraphics[width=\textwidth]{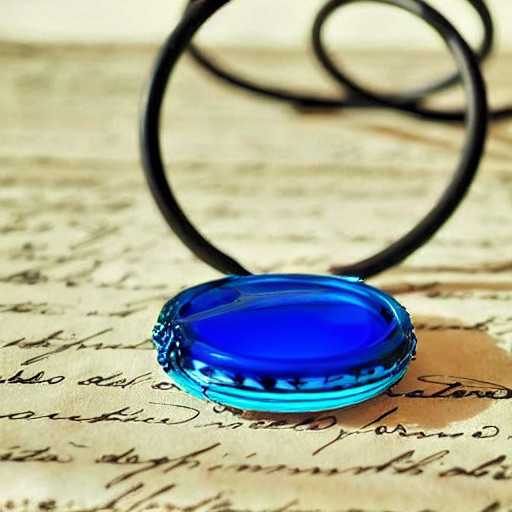}
        \caption*{HDP}
  \end{minipage}
          \begin{minipage}[b]{0.12\textwidth}
      \centering
        \includegraphics[width=\textwidth]{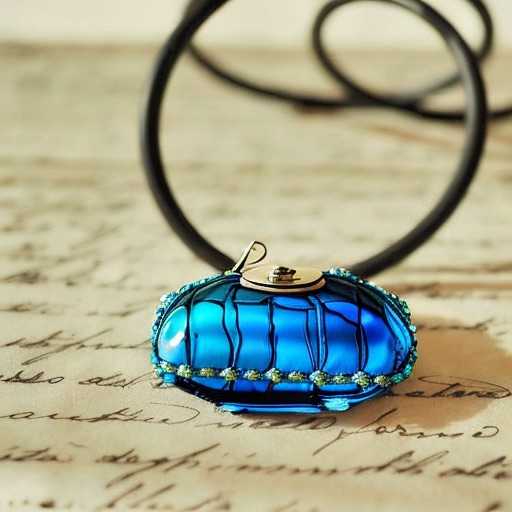}
        \caption*{BrushNet}
  \end{minipage}
      \begin{minipage}[b]{0.12\textwidth}
      \centering
        \includegraphics[width=\textwidth]{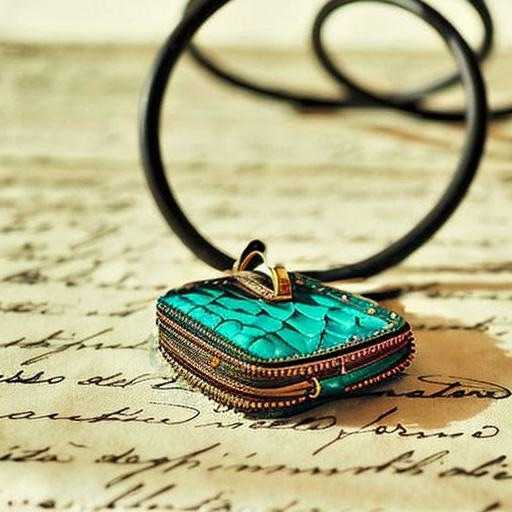}
        \caption*{\textbf{LDB (Ours)}}
  \end{minipage}

\caption{Additional qualitative examples on the PIE-Bench dataset. For all the methods, we used the default settings.}
\label{fig:additionalmagic}
\end{figure*}

\section{Ablation Study Details}
\subsection{Ablation on Mask Strength Control}
\label{sec:alpha}
\begin{figure*}
    \centering
    \includegraphics[width=0.8\linewidth]{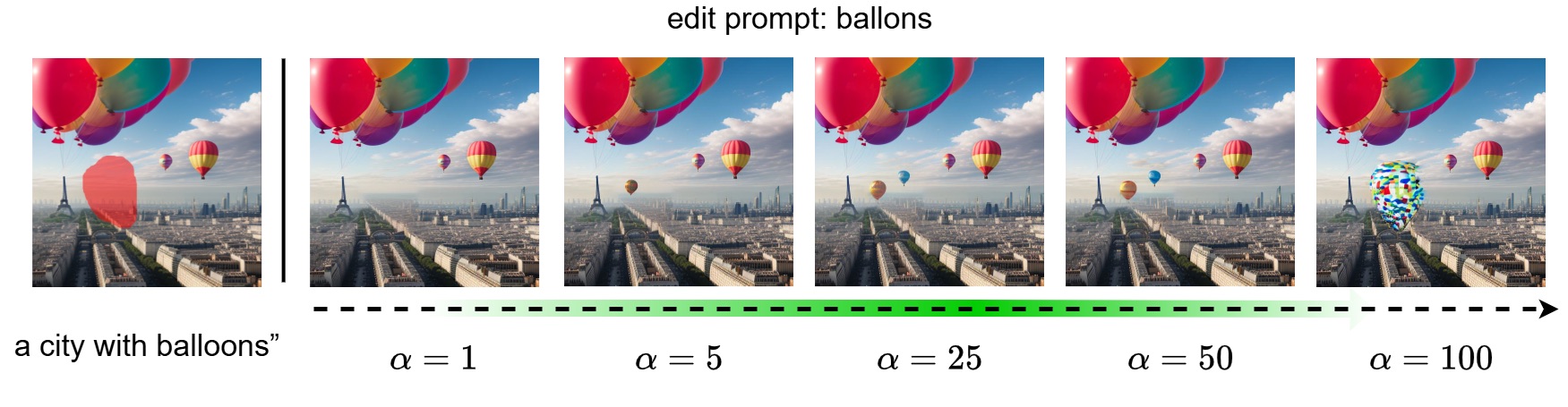}
  \caption{Ablation study on the effect of the strength parameter ($\alpha$) in LDB. We incrementally increase the mask strength ($\alpha$) while keeping the mask, seed, and intermediate denoising steps ($n$) fixed. A value of $\alpha$ that is too large introduces too much noise injection and may cause artifacts, while a value that is too small results in insufficient editing.}

\label{fig:alpha}
\end{figure*}

The magnitude of the edit applied by LDB is jointly governed by the number of edit steps ($n$) and the mask strength control ($\alpha$). These parameters control the amount of intermediate noise added to the latent image.  \cref{fig:alpha} illustrates the effect of varying $\alpha$.  As shown, excessively high $\alpha$ values (right), representing strong edits, prevent the LDM from effectively denoising, leading to artifacts. Conversely, insufficient $\alpha$ results in negligible edits. Furthermore, $n$ and $\alpha$ exhibit a coupled relationship.  When noise is introduced later in the diffusion process (higher $n$), the model has less denoising capacity, necessitating a higher $\alpha$ to achieve a noticeable edit. Conversely, with earlier noise injection (lower $n$), a sufficiently large $\alpha$ is required to prevent the additive noise from being entirely diffused away in the initial denoising steps.  Therefore, optimal editing requires careful consideration of both $n$ and $\alpha$, with $\alpha$ needing adjustment based on the chosen $n$ to balance edit strength and image quality. In our UI, we formulate the translation from $\alpha^*$ to $\alpha$ to decouple these two parameters by factoring in the variance and covariance of the intermediate latent, thus automatically adjusting $\alpha$ when $n$ changes.

\subsection{Caching Latents Ablation Metrics}
\begin{figure*}
  \centering
  
        \includegraphics[width=0.99\textwidth]{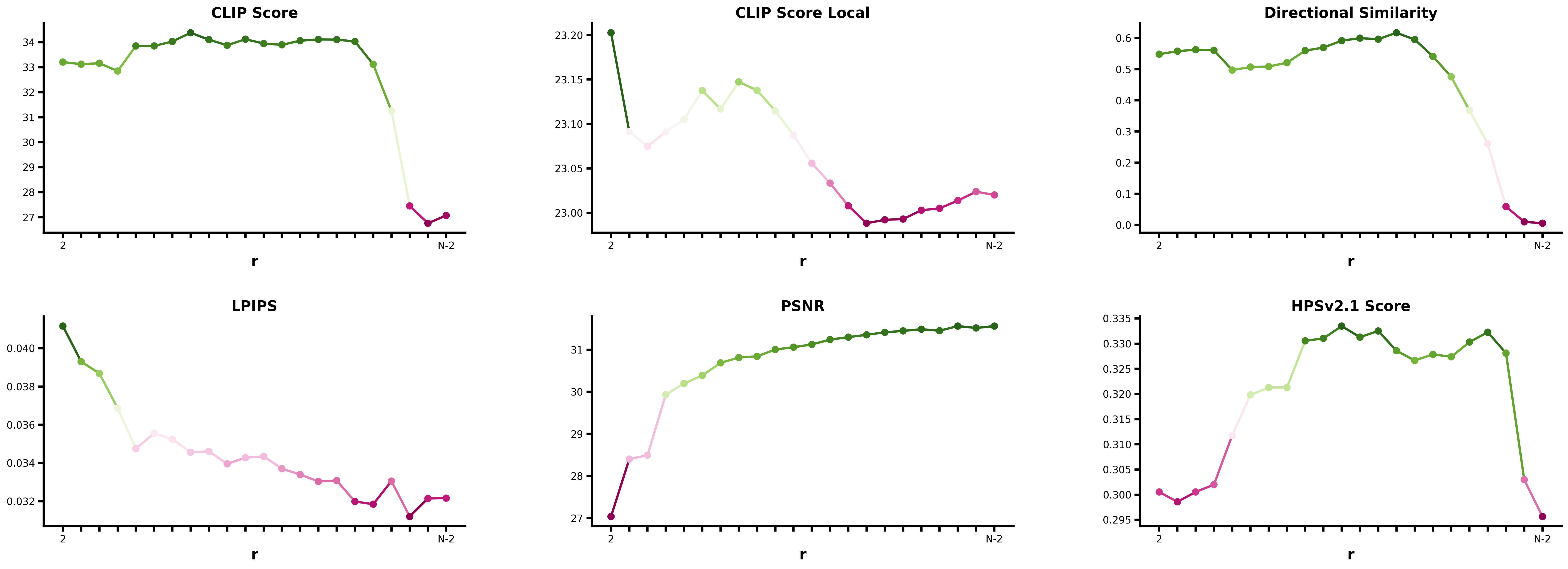}
\caption{
Quantitative evaluation of metrics across different regeneration step values ($r$). The x-axis represents the regeneration step $r$, increasing from left to right from 2 to $N-2$, while the y-axis shows the corresponding score values for each metric.}

\label{r_abl_metrics}
\end{figure*}

\begin{figure*}
  \centering
  
        \includegraphics[width=0.99\textwidth]{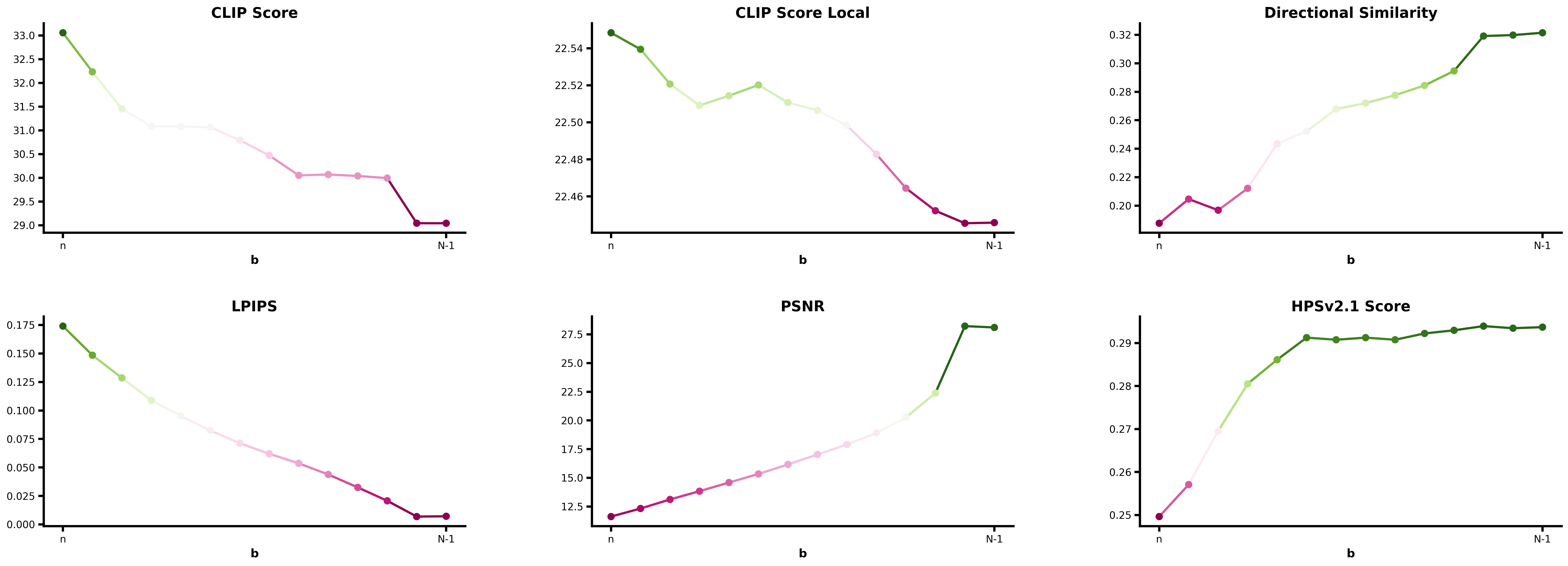}
\caption{
Quantitative metrics for different blending steps ($b$): The x-axis represents the blending step $b$, increasing from left to right from $b=n$ to $N$, while the y-axis shows the corresponding score values for each metric. Smaller $b$ steps lead to poor background protection, while larger $b$ values preserve background integrity and improve edit effectiveness.
}
\label{b_abl_metrics}
\end{figure*}

\cref{r_abl_metrics} and \cref{b_abl_metrics} present the graphs for quantitative metrics on the ablation studies as discussed in \cref{abl}.

\section{Video Editing Examples}
\begin{figure*}
    \centering
    \includegraphics[width=\linewidth]{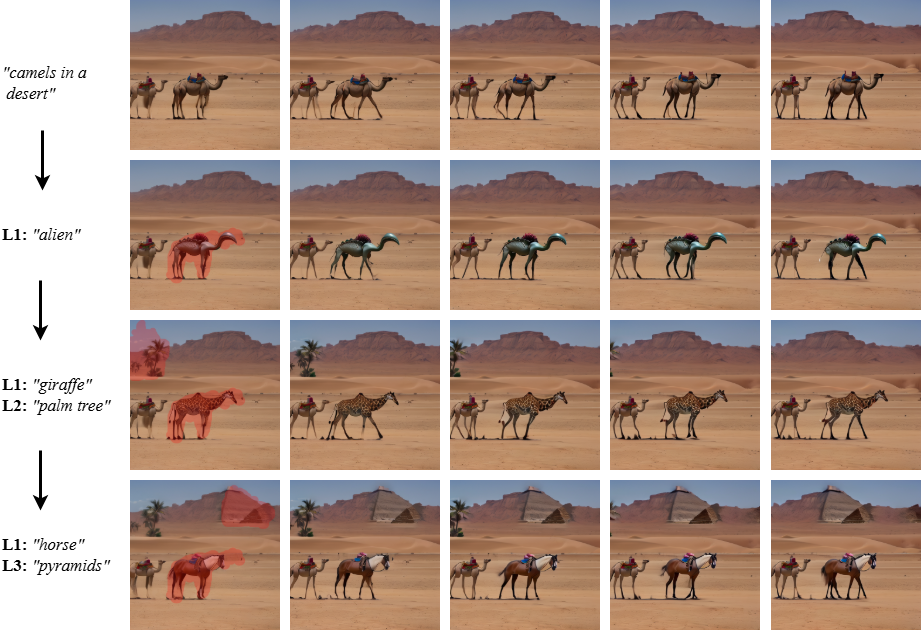}
    \caption{Video editing examples using LDB and Stable Video Diffusion (SVD). The top row displays frames from an input video generated by SVD. For localized editing, we define a mask on the first frame and apply LDB edits to this initial frame.  LDB's caching mechanism is then extended to the temporal dimension within SVD, enabling efficient propagation of edits across subsequent frames. This allows for the creation of multiple editing layers, and even non-sequential fast modifications to different parts of the video by revisiting and adjusting previous layers, while maintaining temporal coherence.}
    \label{fig:video_editing}
\end{figure*}

We integrated LDB with several diffusion image transformers (DiT) and spatio-temporal video generation models. In \cref{fig:video_editing}, we demonstrate examples of video editing by integrating LDB into SVD \cite{blattmann2023stablevideo}. 

\section{User Study Details}
\subsection{Procedure and Task Description}
The user cohort comprised four females and three males, with an average age of 30.4 years. Two participants were proficient in image generative models and Stable Diffusion, while the remaining five were graphic design students who used Adobe Photoshop and Illustrator on a daily basis.
The study was conducted remotely; participants were provided a link to access the tool. 

The study started with a brief introduction to each of the methods. Following this, participants received a short tutorial on how to navigate the user interface (UI). Subsequently, they were provided with a 5-minute window to explore the various options and sections of the tool, becoming familiar with the use of each section.

A dedicated task section was incorporated into the user interface (UI) specifically for the user study. Each type of task comprised three rounds of edits using the three methods: LDB, IP2P, and SDI.

\begin{figure*}
  \centering
        \includegraphics[width=0.99\textwidth]{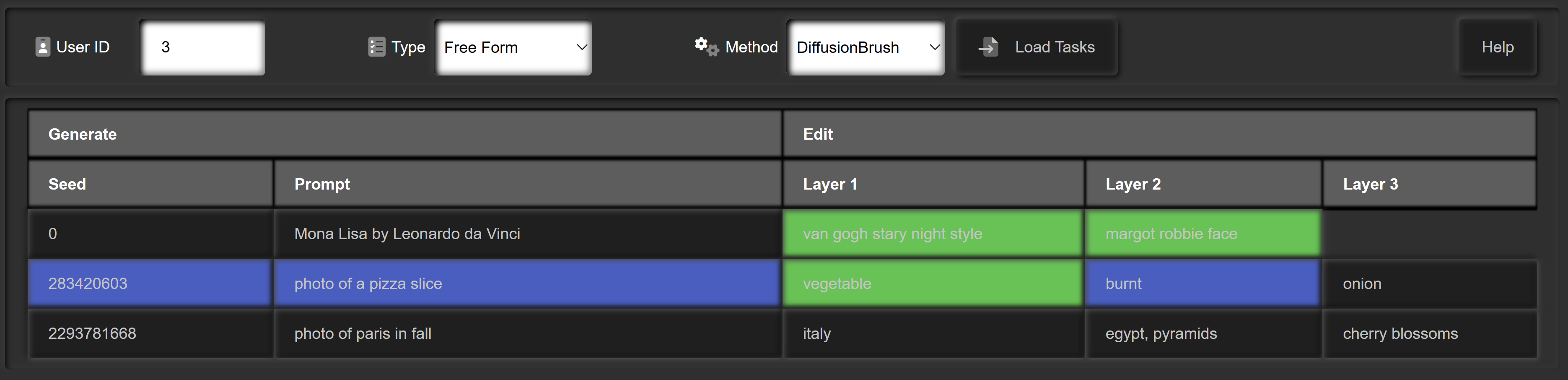}
\caption{Overview of the tasks section, where users can interact to load, select, and save each task. Tasks that are selected are highlighted in blue, while those completed and saved are highlighted in green.}
\label{tasks}
\end{figure*}

Each user was assigned a unique user ID, and tasks were randomly selected and pre-assigned to users. Throughout the study, users interacted with the task table to load, select, and save each task. An example of the task section is illustrated in  \cref{tasks}.

As mentioned in \Cref{study_proc}, the user study consisted of two types of tasks: free-form (type 1) and pre-determined (type 2) tasks. For the type 1 tasks, we selected specific types of edits that showcase various functionalities and capabilities of the system. Here are the description of edit types along with an example used during the user study:
\begin{enumerate}
    \item Stack layers and create sequential edits (draw with LDB):
    \begin{itemize}
        \item Input image: photo of a beautiful beach.
        \item Layer 1: boat (Introduce a boat in the sea)
        \item Layer 2: rocks (Scatter weathered rocks along the shoreline)
        \item Layer 3: birds (Populate the sky above the boat with a flock of birds.
    \end{itemize}
    \item Modify attributes and features of objects:
    \begin{itemize}
    \item Input image: portrait of a young man 
        \item Layer 1: blond (Transform a person's hair color to blond).
        \item Layer 2: joker (Perform facial manipulation by swapping one person's face with another's, reshaping identities.)
    \end{itemize}
    \item Correct image imperfections and errors:
    \begin{itemize}
        \item Input image: portrait of a man holding an umbrella
        \item Layer 1: remove the rod that is mistakenly placed
        \item Layer 2: fix the extra part on the side of the coat
        
    \end{itemize}
    \item Enhance discernibility of similar objects through modification:
    \begin{itemize}
        \item Input image: aerial photo of a pool table with balls
        \item Layer 1: change the colour of a specific ball (third ball from the left) to red
    \end{itemize}
    \item Target specific regions for style transfer, refining aesthetics:
    \begin{itemize}
        \item Input image: Mona Lisa by Leonardo Da Vinci
        \item Layer 1: make the left part of the background similar to Van Gogh starry night style.
    \end{itemize}
\end{enumerate}

In our study design, we strategically chose the combination of seeds and prompts to encompass and evaluate these functionalities. Each user was given three seed-prompt items and tasked with creating and editing up to three layers of edits. For the majority of the tasks, $N$, i.e. the total number of steps for editing was set to $n=5$. All the images were generated using Dreamshaper-7 \cite{dreamshaper} and the DDIM scheduler.

For the LDB method, users started by selecting a layer with an existing edit instruction from the task table, then created the corresponding layer in the UI. They had the option of choosing either the box option or the custom mask option. The task was followed by drawing the mask, tweaking the controls or edit prompt if needed, and completing the edit. Once the task was complete, the user saved the edit and moves on to the next task.

Users followed a similar procedure for the IP2P and SDI methods, with the exception of creating layers, as these methods do not incorporate layering capabilities. After completing each layer edit task, users saved the edits, and the user interface (UI) stacked subsequent edits onto the edited image.
For IP2P method, users were required to write the instruction prompt and then adjust the image and text guidance scales and regeneration steps to finalize the edit. On the other hand, for the SDI method, users drew a mask and controlled the edit using the strength control. Completion times for each task were recorded for both methods.

Type 2 tasks, corresponding to the MagicBrush dataset \cite{zhang2024magicbrush}, were more structured, with the mask, edit prompt, and input images provided by the dataset. MagicBrush utilized crowd workers to collect manual edits using DALL-E 2 \cite{ramesh2022hierarchical}. This process involved 5,313 editing sessions and 10,388 editing iterations, resulting in a robust benchmark for instructional image editing. Additionally, the dataset provides manually annotated masks and instructions for each edit and contains up to three layers of edits. Users selected each image, started with the provided mask, could modify the mask if necessary, adjusted the control parameters and prompt, and saved and completed the task for each method.

\subsection{Evaluation Survey}
After completing the image editing tasks, the participants were asked to complete a three stage evaluation survey. The first part included a System Usability Scale (SUS) form to rate the usability, ease of use, design, and performance of each method. SUS is a standard usability evaluation survey which is widely used in user-experience literature \cite{brooke_susa_1996}. 
The participants were presented with 10 questions about each of the methods and were asked to rate each system on a scale of 1 to 5 for each question. A rating of 1 indicated strong disagreement, while a rating of 5 indicated strong agreement. The questions were designed to assess the participants' perceptions of the effectiveness, ease of use, and overall user experience of each tool. 
Below is the list of the questions:

\begin{enumerate}[label=\textbf{Q}\textsubscript{\textbf{\arabic*}}, itemsep=0em]
    \item I think that I would like to use this tool frequently.
    \item I found the tool unnecessarily complex.
    \item I thought the tool was easy to use.
    \item I think that I would need the support of a technical person to be able to use this tool.
    \item I found the various functions in this tool were well integrated.
    \item I thought there was too much inconsistency in this tool.
    \item I would imagine that most people would learn to use this tool very quickly.
    \item I found the tool very cumbersome to use.
    \item I felt very confident using the tool.
    \item I needed to learn a lot of things before I could get going with this tool.
\end{enumerate}

SUS consists of positive and negative phrasing questions. Q2, 4, 6, 8, and 10 are negatively framed, therefore on the chart, red colours means better SUS score and Q1, 3, 5, 7, and 9 are considered positively framed and hence, more green colours demonstrate better score.


\begin{figure*}[!ht]

          \begin{minipage}[t]{0.32\textwidth}
  
        \includegraphics[width=\textwidth]{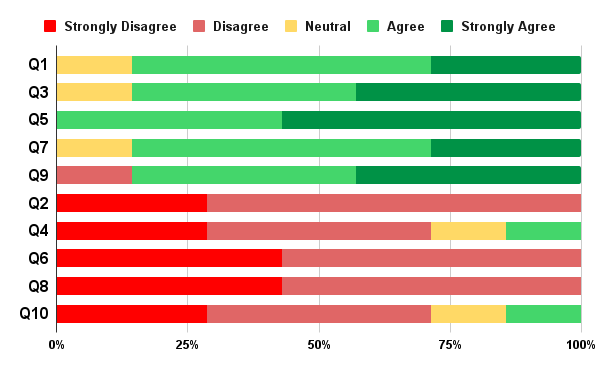}
                  \vspace{-0.7cm}\caption{LDB usability}
  \end{minipage}
        \begin{minipage}[t]{0.32\textwidth}
    \includegraphics[width=\textwidth]{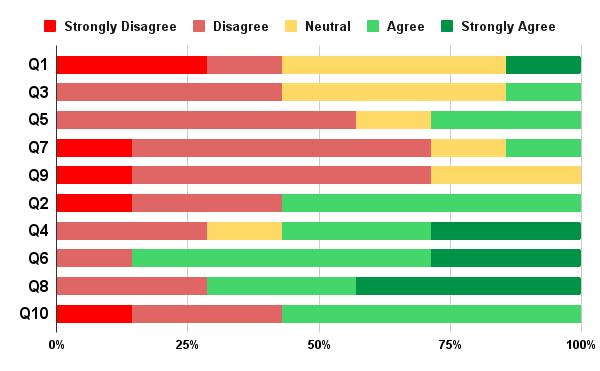}
\caption{SDI usability}
  \end{minipage} 
          \begin{minipage}[t]{0.32\textwidth}

\includegraphics[width=\textwidth]{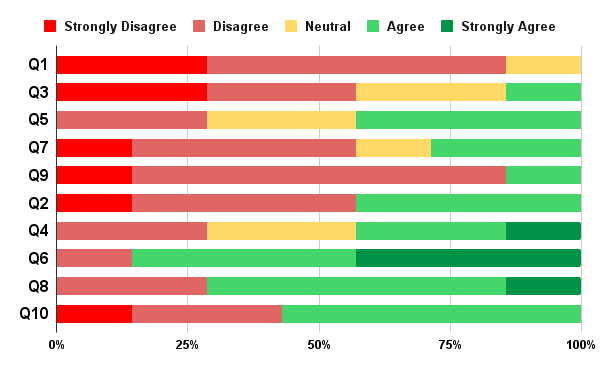}
\caption{IP2P usability}
\label{ps-res}
  \end{minipage}

\caption{Results of Q1 - Q10 for the usability of each system among different participants. For odd questions, green colors show more desirable feedback. Even questions are designed with negative wording and more red colors show more favorable feedback.}

\label{fig:usability}
\end{figure*}
The survey was followed by an interview with each participant to gather specific feedback and insights based on their artistic background and experience using the different tools. These processes provided valuable information on the strengths and weaknesses of each tool, as well as how it can be improved to better serve users. 

The following multiple-choice questions were also asked for evaluating the performance of each method:
\begin{itemize}
\item How much time did it take you to complete the image editing task using the tool you used in this study? [Much less time/About the same/Much more time]

\item How did you find each of the tools in terms of effectiveness in achieving the desired edits? [Very effective/Somewhat effective/Neutral/Somewhat ineffective/Very ineffective]

\item How does each of the tools you used perform in terms of time to complete the editing task? [Much faster/Somewhat faster/Acceptable/Somewhat slower/Much slower]

\item How likely are you to use each of these tools as an AI image editing tool in the future? [Very likely/Somewhat likely/Neutral/Somewhat unlikely/Very unlikely]

\end{itemize}
The entire study, including filling out the evaluation surveys, took not more than 90 minutes.

\begin{figure*}[!ht]
  \centering
      \begin{minipage}[b]{0.19\textwidth}
    \includegraphics[width=\textwidth]{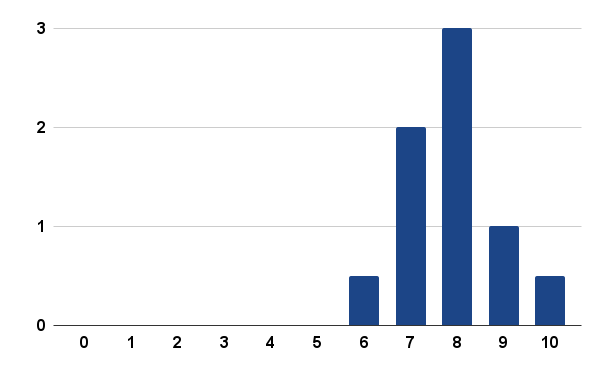}
    \caption*{\small Enjoyment}
    
  \end{minipage} 
          \begin{minipage}[b]{0.19\textwidth}
    \includegraphics[width=\textwidth]{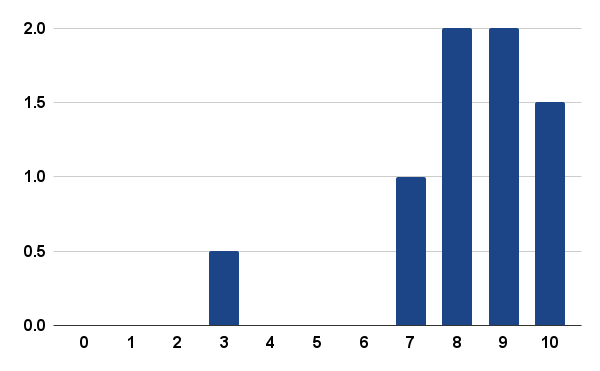}
        \caption*{\small Expressiveness}
  \end{minipage}
        \begin{minipage}[b]{0.19\textwidth}
    \includegraphics[width=\textwidth]{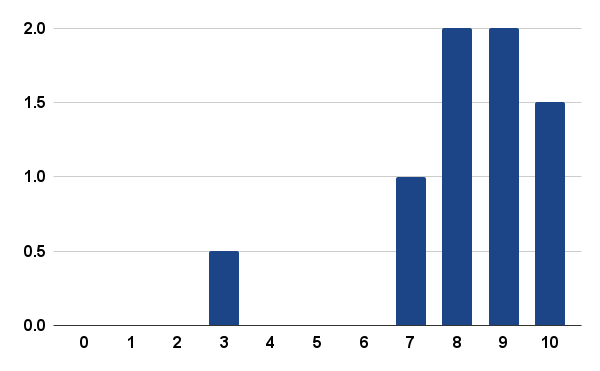}
    \caption*{\small Exploration}
    
  \end{minipage} 
          \begin{minipage}[b]{0.19\textwidth}
    \includegraphics[width=\textwidth]{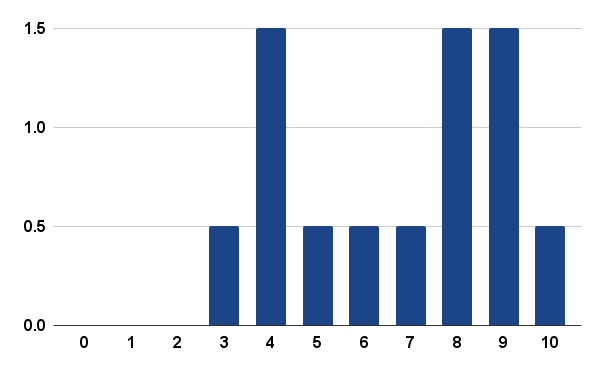}
        \caption*{\small Immersion}
  \end{minipage}
            \begin{minipage}[b]{0.19\textwidth}
    \includegraphics[width=\textwidth]{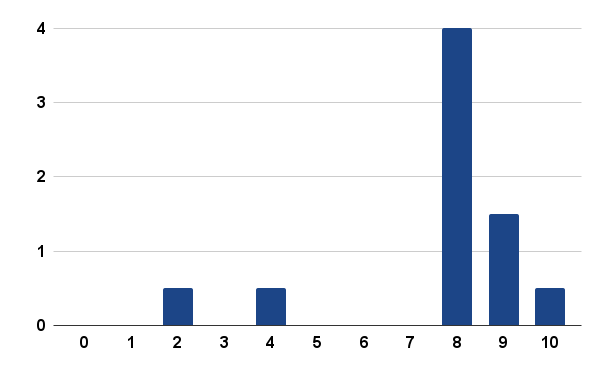}
        \caption*{\small Result Worth Effort}

  \end{minipage}
\caption{Histogram of the Creativity Support Index from the user study survey.}
        \label{fig:csi}
\end{figure*}
\cref{fig:csi} illustrates the outcomes of the post-study CSI survey. Overall, participants expressed positivity towards LDB, indicating that it enhanced their enjoyment, exploration, expressiveness, and immersion, while also deeming the results worth their effort.
The CSI score results also show that one participant responded neutrally or negatively to certain aspects, likely due to their being accustomed to the Photoshop tool. Furthermore, there was notable variability in immersion scores, with several participants giving lower ratings. This variability suggests that while some users felt deeply engaged with the tool, others may have encountered challenges or distractions affecting their immersive experience. Analyzing specific factors such as interface design, task complexity, and user preferences could offer insights into enhancing immersion in future iterations of LDB. Despite this variability, the majority of participants found the tool effective and engaging, highlighting its potential usefulness in creative workflows.

One of the most common comments regarding the usability of different methods was that participants found it challenging to find the optimal settings for IP2P and SDI. For example, one user mentioned, ``In InstructPix2Pix, increasing the image guidance scale often distorts the edited image too much, and if the text guidance scale is too high, the edited image looks completely different. After many trials and errors, when I find a good combination, the next image behaves differently. Also, SD-inpainting half the times fails to produce a satisfactory result.''

Another user, who is an expert in graphic design, suggested, ``Layers are very helpful. I would like to see the control numbers on top of them as I change them, not beside them. Also, having an undo button is crucial and would be very helpful. Additionally, I would suggest adding a blend option to each layer, similar to Photoshop''. These suggestions will be taken into consideration for future improvements.

\subsection{System Usability Scale (SUS)}
 \cref{fig:usability} presents the results of the SUS survey among participants after using LDB, SDI, and IP2P. Based on the bar charts, participants indicated that they are more likely to use LDB compared to IP2P and SD-Inpainting, and that they find it the easiest tool to use. In addition, participants in Q4 expressed that they would not require technical assistance to use the system in the future, indicating its overall good design. These findings were further supported by the interview feedback. For example, when asked about their understanding of the different parameters in the tool, one participant stated: ``I believe that I understand the functionality of each parameter. I need to increase the mask strength value if I want to make bigger changes. The tool is quite intuitive and easy to use, and I think I can easily use it without needing any technical support.'' This feedback highlights that the tool has a user-friendly design and can be easily understood and used by a wide range of users. Based on the survey results, the SUS score for LDB is calculated as 80.35\%, while IP2P and SDI achieve a score of 38.21\% and 37.5\% respectively. 

For CSI \cite{cherry2014quantifying} questionnaire we used all questions, excluding questions about collaboration as it is  not relevant for our tool. The CSI measures dimensions of Exploration, Expressiveness, Immersion, Enjoyment, and Results Worth Effort in a tool. CSI helps in understanding how well LDB support creative work overall, as well as pointing out which aspects of creativity support may need attention.

Figure \ref{fig:csi} illustrates the outcomes of the post-study CSI survey. Overall, participants expressed positivity towards LDB, indicating that it enhanced their enjoyment, exploration, expressiveness, and immersion, while also deeming the results worth their effort.

\section{Initial User Study Insights}
We initially developed an earlier version of LDB, called \textit{Diffusion Brush}, with the objective of re-randomizing targeted regions for fine-tuning (\eg fixing small details that were generated incorrectly) and without layering functionalities. Subsequently, we conducted a user study to assess its usability and features and based on the feedback received from this first study, we made significant improvements and revamped the tool. In the first user study, we compared the early version of LDB with SDI and manual editing in Adobe Photoshop \cite{adobephotoshop}, involving five expert users.

While the majority of participants acknowledged that Diffusion Brush was faster than manual editing, some participants suggested that even faster editing would be significantly beneficial, aiding in random idea generation for artists. To address this feedback, we incorporated a caching mechanism, as explained in \Cref{caching}, designed an efficient front-end to communicate with the machine learning backbone, and highly optimized the overall pipeline, achieving as little as 140 ms of inference time for a single edit on a high-end consumer GPU.

Furthermore, a few users struggled with finding the optimal brush strength control, a similar challenge observed in SD-inpainting as well. To address this, we devised a more generalized approach. While the earlier version of our system also supported multiple masks, these masks were not fully independent, and deleting or hiding them was not possible without performing operations in a specific order. This observation prompted the creation of a more streamlined and flexible mask management system.

Additionally, insights gathered from the first round of interviews indicated the need for further improvement in various aspects of the tool's functionality and user experience. These inputs guided us in refining the tool and enhancing its usability for a wider range of users.
Lastly, in the first user study, three participants specifically mentioned this feedback. One participant stated, ``I really like the tool as it is right now; it certainly provides value for me in my editing tasks and makes my life easier. But one feature that I would love to see is to be able to tell the system how to make these changes. I still want to use the masking editing, but if I can tell it what to do it would be great.''. 
Based on the findings of the new user study, it is evident that this feature has been well-implemented into the system. All users participating in the current study affirmed the effectiveness of this feature.